\documentclass{article}

    \PassOptionsToPackage{numbers, compress}{natbib}

\usepackage[square,numbers]{natbib}
\usepackage[preprint]{neurips_2022}

\usepackage[utf8]{inputenc} 
\usepackage[T1]{fontenc}    
\usepackage{hyperref}       
\usepackage{url}            
\usepackage{booktabs}       
\usepackage{amsfonts}       
\usepackage{nicefrac}       
\usepackage{microtype}      
\usepackage{xcolor}         

\usepackage[ruled]{algorithm2e} 
\usepackage{amsmath}
\usepackage{amsthm}
\usepackage{amssymb}
\usepackage{subfigure} 
\usepackage{svg}
\usepackage{wrapfig}
\usepackage{graphicx}                                                           
\usepackage{float} 

\usepackage{multirow}
\usepackage{hyperref}

\theoremstyle{plain}

\title{AsyncFedED: Asynchronous Federated Learning with Euclidean Distance based Adaptive Weight Aggregation}

\author{
    Qiyuan Wang, Qianqian Yang, Shibo He, Zhiguo Shi, Jiming Chen \\
    The State Key Laboratory of Industrial Control Technology \\
    Zhejiang University \\
    \texttt{\{qy\_wang, qianqianyang20, s18he, shizg, cjm\}@zju.edu.cn} \\
}

\begin{document}

\maketitle

\begin{abstract}
In an asynchronous federated learning framework, the server updates the global model once it receives an update from a client instead of waiting for all the updates to arrive as in the synchronous setting. This allows heterogeneous devices with varied computing power to train the local models without pausing, thereby speeding up the training process. However, it introduces the stale model problem, where the newly arrived update was calculated based on a set of stale weights that are older than the current global model, which may hurt the convergence of the model. In this paper, we present an asynchronous federated learning framework with a proposed adaptive weight aggregation algorithm, referred to as AsyncFedED. To the best of our knowledge this aggregation method is the first to take the staleness of the arrived gradients, measured by the Euclidean distance between the stale model and the current global model, and the number of local epochs that have been performed, into account. Assuming general non-convex loss functions, we prove the convergence of the proposed method theoretically. Numerical results validate the effectiveness of the proposed AsyncFedED in terms of the convergence rate and model accuracy compared to the existing methods for three considered tasks.

\end{abstract}

\section{Introduction}

Federated learning enables a number of clients to train a global model cooperatively without raw data sharing which preserves data privacy\cite{FedAvg, openproblem, briggs2020federated,FedAvgConvergence,fedthroretical}. However the classic \textit{Synchronous Federated Learning}(SFL)\cite{FedProx,FedOPT,AdaptiveFL,FedSA} faces the issue of straggler effect, caused by device heterogeneity, where the server has to wait all the updates from the clients, which can be significantly delayed by the straggler in the system \cite{cipar2013solving,harlap2016addressing,chai2020fedat}. \textit{Asynchronous Federated Learning}(AFL)\cite{chen2020asynchronous} has been proposed to tackle this issue, where each client updates the global model independently instead of waiting for other updates to arrive, which shows more flexibility and scalability.

Consider a AFL setting where client $i$ joined the training process at global model iteration $t-\tau$ and downloaded the latest version global model weight denoted as $x_{t-\tau}$. Generally, its local update results should be aggregated with $x_{t-\tau+1}$. However, in AFL setting there are other clients updating the global model independently. Hence, when client $i$ upload its local results, the global model iteration has reached $t$. Directly aggregating the model update by client $i$ and the current model weight $x_t$ may hurt the model convergence, referred to as the stale model problem\cite{asynchronoussurvey,FedSA}.

This stale model problem has been extensively studied in the \textit{Asynchronous Parallel Stochastic Gradient Descent}(A-PSGD) setting \cite{Assumption3,hogwild,psgd,agarwal2011distributed,feyzmahdavian2016asynchronous,paine2013gpu}, but few in the  AFL setting\cite{asynchronoussurvey}, due to its unique challenges such as multiple local epochs and statistical heterogeneity caused by Non-IID data distribution\cite{noniidoptimizing,briggs2020federated,FedAvgConvergence}. The existing strategies to alleviate the stale model problem or simply bound staleness measure the staleness of a client in two ways: by local training time cost\cite{AsyncFed,Assumption3} or by the number of iteration lag $\tau$\cite{aviv2021learning,asgd,psgd,FedBuff}.

\begin{figure}
    \centering
    \subfigure[Griewank function]{\includegraphics[width=0.37\textwidth]{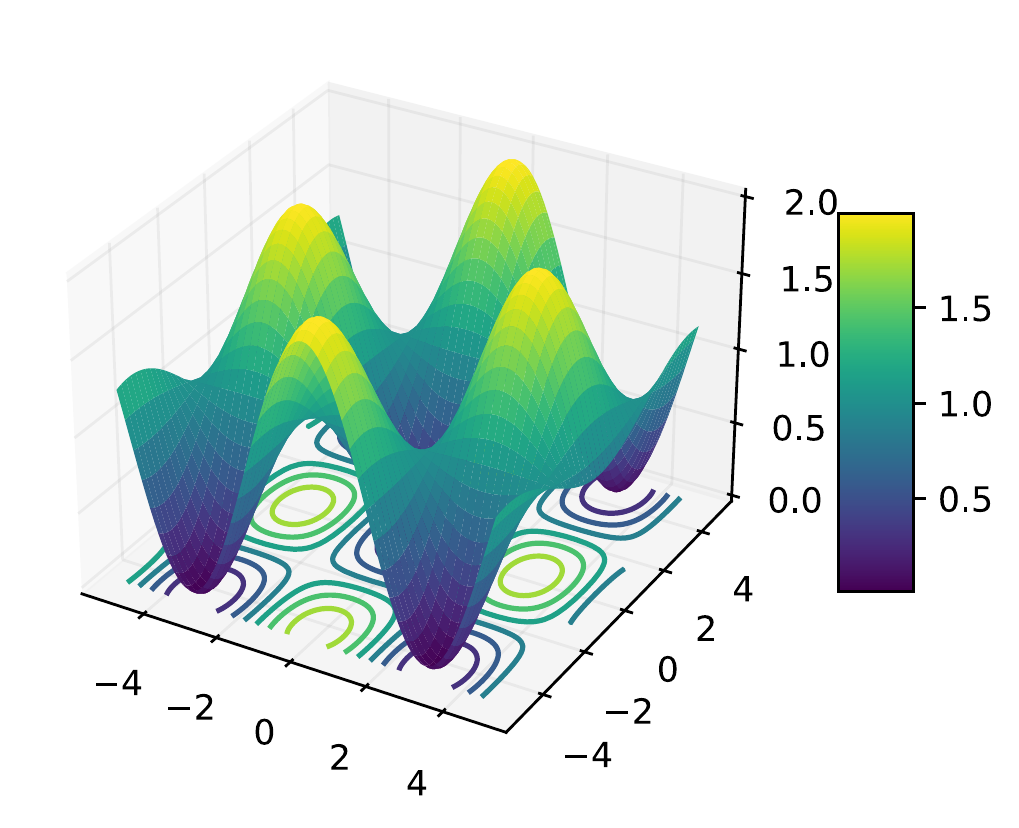}}
    \subfigure[Contour view]{\includegraphics[width=0.59\textwidth]{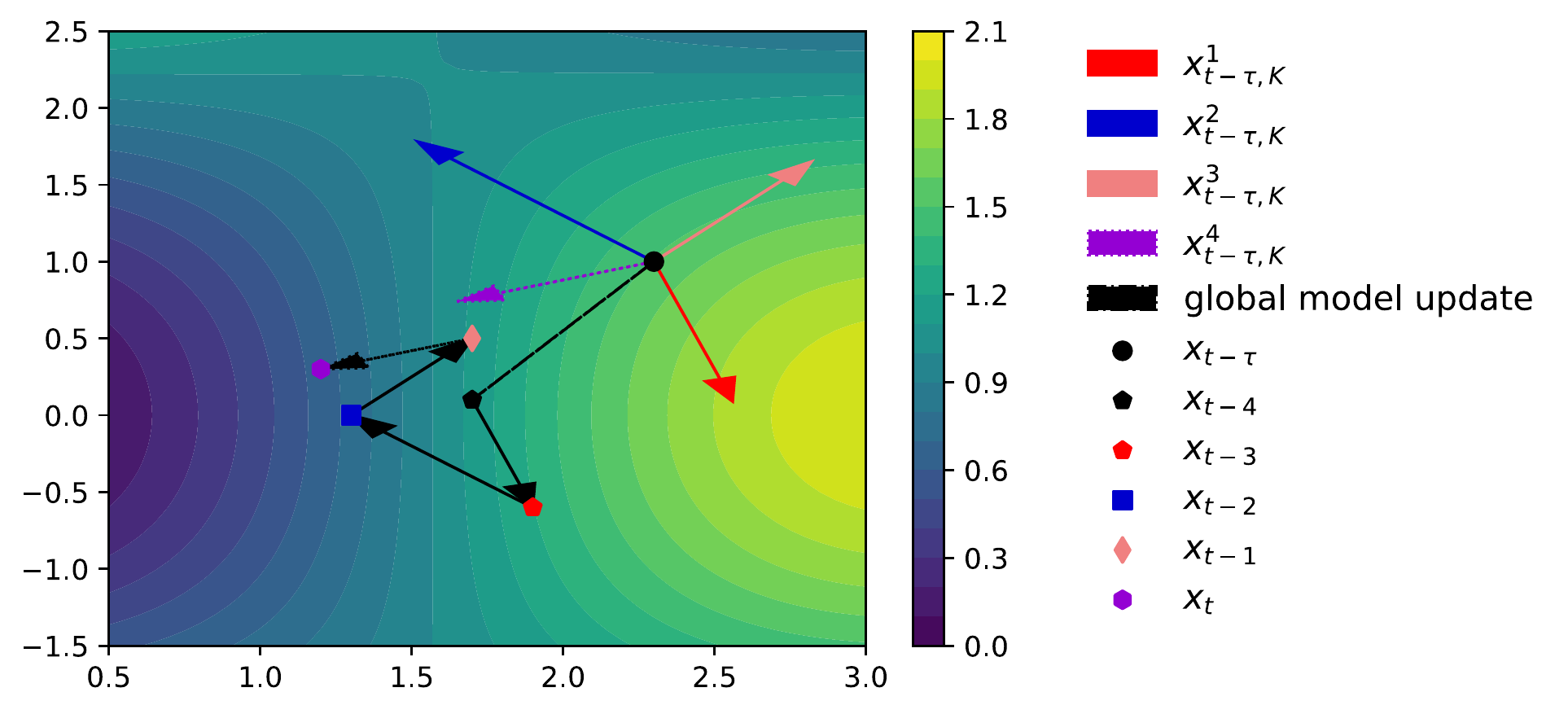}}\label{contour} 
    \caption{A toy example with the Griewank function\cite{griewankfunction}. Assuming four AFL clients indexed by $1,2,3,4$ in this example, they independently downloads the global model parameters $x_{t-\tau}$ at  iteration $t-\tau$, and each performs $K$ local epochs updates independently. Due to devices heterogeneity, the time each client spends to complete $K$ local epochs is different. Assume the arrived order is $x_{t-\tau,K}^1, x_{t-\tau,K}^2, x_{t-\tau,K}^3, x_{t-\tau,K}^4$, represented by coloured arrows in (b), e.g., $x_{t-\tau,K}^4$ comes from the slowest client. Assume that when the first among these four updates  $x_{t-\tau,K}^1$ arrived at the server, the global model has been updated to $x_{t-4}$. Then these updates aggregate  with the global model one by one as shown by the black arrows in (b). Note that the directions(direction of coloured arrows in (b)) of local updates are diverse due to the Non-IID data distribution, which cause the unstable declining for loss function as observed. We note that, when evaluate the staleness based on iteration lag or local training time, $x_{t-\tau,K}^4$ will be set with a very small aggregate weight(the length of black arrow) or even being discarded. 
    } 
    \label{toy example}
\end{figure}

\paragraph{Limitations}\label{limitions} Although the experimental results have shown the effectiveness of these strategies on solving the stale model problem to some extents, the limitations in practical scenario are obvious, which can be illustrated by a toy example in Figure \ref{toy example} where useful update from a slow client is mistakenly discarded. The limitation of existing AFL algorithms that quantifies the staleness of the local model update using the local training time cost or the iteration lag can be summarized as follows:

\begin{itemize}
    \item \textbf{Lack of adaptability to changing resources.} When the resources in the training environment , for instance, the communication resources and the computing capacity, changes, the weight aggregation scheme should be adjusted according, which has been overlooked in the existing work. Additionally, by simply bounding the staleness of the local update, the existing scheme limit the number of local epochs, which may miss an optimal $K$ that speeds up the model convergence, which will be explained later in this paper. 
    
    \item \textbf{Discard useful update.}
    As illustrated by the toy example showed in Figure \ref{toy example}, an useful update may arrive late at the server, which will be wrongly discarded as the staleness of this update, evaluated by the iteration lag or local training time, is above certain threshold. This obviously can hurt the convergence of the training model and slow down the training.

\end{itemize}

\paragraph{Main works and contributions} Taking into account the above concerns, this paper proposed a novel AFL framework, referred to as AsyncFedED, with adaptive weight aggregation algorithm, and the performance has been validated both theoretically and empirically. Our main works and contributions can be summarized as follows: 

\begin{itemize}
    \item \textbf{An evaluation method for staleness.} We proposed a novel method to evaluate staleness taking into account both the Euclidean distance between $x_t$ and $x_{t-\tau}$ and the local model drift, which can make full use of local update results to speed up the convergence. To the best of our knowledge, it is the first work to evaluate staleness based Euclidean distance.
    
    \item \textbf{An adaptive model aggregation scheme with regards to the staleness.} 
    We set the global learning rate adaptively for each received model update from clients with regards to the staleness of this update. We also propose an update rule of the number of local epochs each client performs at every iteration, which balances the staleness of the received model updates from all the clients.  
    
    \item \textbf{Theoretical analysis.} Assuming general non-convex loss functions, we theoretically prove the convergence of the proposed  AsyncFedED algorithm.

    \item \textbf{Experimental verification.} We compare the proposed AsyncFedED with four baselines for three different tasks. The numerical results validate the effectiveness of AsyncFedED in terms of convergence rate, model accuracy, and robustness against clients suspension.

\end{itemize}

The rest of this paper is organized as follows. Section \ref{relatedworks} introduces some related works, Section \ref{problemFormulation} gives the problem formulation and Section \ref{adaptivealgotithms} describes AsyncFedED. Main theoretical analyses and discussions are presented in Section \ref{theoreticalresults}. Section \ref{experimentalresults} describes the experimental settings and gives experimental results. Section \ref{conclusions} is conclusions. Due to the space constraints, more proof details and experimental details can be found in Appendix \ref{proof details} and Appendix \ref{experimentaldetails}, respectively.

\section{Related Works}\label{relatedworks}

\paragraph{Stale model problem in A-PSGD} As pioneer works of A-PSGD, \cite{dean2012large,gupta2016model,mania2015perturbed,feyzmahdavian2016asynchronous,mitliagkas2016asynchrony} allow server to update global model using stale model parameters. Theoretically, \citet{agarwal2011distributed} give the convergence analysis with convex loss functions and \citet{asgd} provides a more accurate description for A-PSGD with non-convex loss functions. \citet{Assumption3} present the first analysis of the trade-off between error and the actual runtime instead of iterations. In order to alleviate the negative impact brought by stale 
models, \citet{delaycompensation} studied the Taylor expansion of the gradient function and propose a cheap yet effective approximator of the Hessian matrix, which can compensate for the error brought by staleness.

\paragraph{Asynchronous federated learning}

To mitigate the impact of straggler and improve the efficiency of SFL, some AFL schemes are introduced\cite{chen2020asynchronous,gu2021privacy,AsyncFed,chen2018lag}. Specially, \citet{wu2020safa} classified participated clients into three classes and only the sustainable clients are asynchronous. \citet{FedBuff} set a receive buffer, when the number of local updates server received reaches the buffer size, the local updates in the receive buffer will be weighted averaged and perform a global model aggregation. \citet{FedSA} divide federated learning training process into two stages, during the convergence stage, each client $i$ is assigned a different $K_i$ based on the overall status, ensuring that the local update arrives at server at the same time.

\paragraph{Adaptive algorithms in federated learning}

Most hyperparameter optimization methods\cite{zaheer2018adaptive,wu2019global,ward2019adagrad} under the framework of machine learning can directly implement in federated learning. For instance, \citet{FedOPT} constructed a optimization framework and introduced the SGD optimization methods such as Adam\cite{Adam}, Momentum\cite{li2019gradient} into federated learning. Considering communication and computing efficacy for mobile devices, all local loss functions in \cite{FedProx} have been added with a regularization term to improve the performance on Non-IID data. To save communication bandwidth and improve convergence rate, based on the overall status of the previous aggregation round, \citet{AdaptiveFL} give a principle to adaptability set local update epoch $K$ for next round.

\section{Problem Formulation}\label{problemFormulation}

Consider a total of $m$ clients that jointly train a learning model in a federated learning setting to find an optimal set of model parameters $x^*\in\mathbb{R}^{d}$ that minimize a global loss function:
\begin{equation}
    x^*=\underset{x\in\mathbb{R}^{d}}{\arg \max } F(x):=\frac{1}{m}\sum_{i=1}^{m} f_i(x, \xi_i)
\end{equation} 
where $f_i(\cdot)$ and $i\in [m]$ represents the local loss function and the local dataset at client $i$, respectively, and $F(\cdot)$ denotes the global loss function. We consider the Non-IID case, where the distribution of $\xi_i$ is not identical on different devices. 

The considered asynchrounous federated learning framwork operates as follows. The central server initializes the weights of the global learning model at the beginning, denoted by $x_1$, where we denote by $x_t$ the global weights at iteration $t$. Clients can join the training process at any iteration by downlowning the current global weights, then performing local stochastic gradient descent with local dataset and sending back the gradients back to the server. For example, assume that client $i$ joins at iteration $t-\tau$. It then downloads the latest version of model weights, that is $x_{t-\tau}$, and then runs local SGD for $K$ epochs. Let $\xi_{i,k}\in \xi_i$ denote the mini-batch in $k-th$ SGD step and $\eta_{i,k}$ the local learning rate for this epoch. We have the update of the local model in this step given as 
\begin{equation}
    x_{t-\tau,k}^i = x_{t-\tau,k-1}^{i} - \eta _{i,k} \nabla f_{i} (x_{t-\tau,k-1}^i; \xi _{i,k}). \label{local update}
\end{equation}
After $K$ epochs of local training, the local model at client $i$ is then updated to 
\begin{equation}
    x_{t-\tau,K}^i = x_{t-\tau} - \sum_{k=1}^{K}\eta _{i,k} \nabla f_{i} (x_{t-\tau,k-1}^i; \xi _{i,k})
\end{equation}
Note that the server has access to the global model at iteration $t-\tau$, $x_{t-\tau}$. Client $i$ then only needs to send the update to the server, which is referred to as pseudo gradient in the literature \cite{FedBuff},\cite{FedOPT}, which is
\begin{equation}
\begin{aligned}
    \Delta_i (x_{t-\tau,K}) &= x_{t-\tau,K}^i - x_{t-\tau} = - \sum_{k=1} ^{K}\eta _{i,k} \nabla f_i (x_{t-\tau,k-1}^i; \xi _{i,k}). \label{pseudo gradient}
\end{aligned}
\end{equation}
The server then aggregate $\Delta_i (x_{t-\tau,K})$ with the current global model $x_t$ by:

\begin{equation}
    x_{t+1} = x_{t} + \eta_{g,i} \Delta_{i}(x_{t-\tau,K})
    \label{global update}
\end{equation}
where $\eta_{g,i}$ represents the learning rate set for the update from client $i$, which takes into account the staleness of the locally available global weights $x_{t-\tau_{i_l}}$ at client $i_l$ and the current global model $x_t$ as explained in detail in the next section.

\section{Euclidean Distance based Adaptive Federated Aggregation}\label{adaptivealgotithms}

In this section, we introduce the proposed AsyncFedED with an Euclidean distance based adaptive federated aggregation scheme that adapts the global learning rate for each received update according to the staleness of this update, and adaptively adjust the number of local epochs $K_i$ such that the staleness of each update is within a given value. 

\paragraph{Staleness}
We measure the staleness of each update by
\begin{equation}
    \gamma(i,\tau) = \frac{\left\|x_t - x_{t-\tau} \right\|}{ \left\| \Delta_i (x_{t-\tau,K}) \right\|} \label{determine gamma}
\end{equation}
where the numerator in the right-hand side term is the Euclidean distance between the  current global model $x_t$ and the stale global model $x_{t-\tau}$ used by client $i$ to obtain this update. The denominator is the $\ell^2$-norm of the update $\Delta_i (x_{t-\tau,K})$ based on the intuition that the larger $\Delta_i (x_{t-\tau,K})$ is,  the more this update is able to move from the old global model $x_{t-\tau,K}$, and hence, is less stale.

\paragraph{Adaptive learning rate } We set the learning rate for each update according to its staleness as follows
\begin{equation}
    \eta_{g, i} = \frac{\lambda}{\gamma(i,\tau)+\varepsilon} \label{adaptive eta_g}
\end{equation}

where $\lambda$ and $\varepsilon$ are two hyperparameters which can be carefully tuned with regards to the specific training tasks.  We note that $\varepsilon$ also serves as an offset, which makes sure that the denominator is not zero when $\left\|x_t - x_{t-\tau}\right\|\to 0$ as the model converges. We have that the learning rate $\eta_{g, i}$ decreases with the staleness of the update, which meets the intuition that the more significant the staleness is, the less the corresponding gradients should be updated to the global model.

\paragraph{Adaptive no. local epochs}
Different from the existing works on the AFL problem where every client runs the same number of local epochs for each update despite the device heterogeneity, which may hurt the convergence of the learning model, the proposed AsyncFedED sets the number of local epochs for each client adaptively by 

\begin{equation}
    K_{i,n+1} = K_{i,n} + \mit{E}[(\Bar{\gamma} - \gamma(i,\tau_n)) \cdot \kappa ]\label{adaptiveK}
\end{equation}
\begin{wrapfigure}{r}{0.5\textwidth}
\vspace{-0.2cm}
  \begin{center}
        \begin{algorithm}[H]\label{algorithm1}
        \caption{Training procedure at the server}
        \KwIn{Initial model weights $x_1$; hyperparameters  $\lambda$, $\varepsilon$, $\Bar{\gamma}$, $K_{initial}$, $\kappa$; $t=1$.}
        \KwOut{Global model parameters $x_T$.}

        \If{receive a connection request and $t < T$} 
        {  Send ($x_{t}$, $K_{initial}$) to the newly connected client.
        } 
        \If{receive a model update from client $i$}{
            Receive local update results ($\Delta_i(x_{t-\tau,K_{i,n}})$, $t-\tau$, $K_{i,n}$); \\
            Read the stale global model parameters from GMIS by iteration index $t-\tau$; \\
            Evaluate staleness $\gamma(i,\tau)$ by Eq.(\ref{determine gamma}); \\
            Calculate the global learning rate $\eta_g$ for this update by Eq.(\ref{adaptive eta_g}); \\
            Update global model by Eq.(\ref{global update}) and append it to GMIS; \\
            Calculate $K_{i,n+1}$ by Eq.(\ref{adaptiveK}); \\
            Send ($x_{t}$, $K_{i,n+1}$) to client $i$ to start a new iteration of local training.
        }
    \end{algorithm}
    \begin{algorithm}[H]\label{algorithm2}
        \caption{Local training procedure at client $i$, $i\in [m]$}
        \KwIn{local data set $\xi_i$.}
        \While{Connected with server}{
            Download parameters ($x_{t}$, $K_{i,n}$); \\
            \For{$k=1 \to K_{i,n}$}
            {
                Sample mini-batch $\xi_{i,k}$ randomly from $\xi_i$;\\
                Obtain local stochastic gradient descent by Eq.(\ref{local update}) with the optimizer of choice;
            }
            Compute the pseudo gradient $\Delta_i(x_{t,K_{i,n}})$ by Eq.(\ref {pseudo gradient}); \\
            Upload local results ($\Delta_i(x_{t,K_{i,n}})$,$t$,$K_{i,n}$) to server.
        }
    \end{algorithm}
  \end{center}
\vspace{-1.5cm}
\end{wrapfigure}
where $K_{i,n}$ denotes the number of local epochs at client $i$ for the $n$th update, $\gamma(i,\tau_n)$ is the staleness of this update, $\kappa$ is the step length that governs how much the number of local epochs changes each time and $\mit{E}[\cdot]$ denotes the floor function that output the largest integer smaller than the input. We note that this update rule pushes the number of local epochs to increase if the staleness of the current update is smaller than a given threshold $\Bar{\gamma}$, which will then leads to the increase of the staleness in the next update according to \eqref{determine gamma}. Eventually, the number of local epochs $K_{i,n}$ will converge to a value such that the staleness $\gamma(i,\tau_n)$ of the updates by any client $i$ converges to an identical given value $\Bar{\gamma}$ despite the capacity heterogeneity among different clients. We note that the choice of $\Bar{\gamma}$ controls a tradeoff between the update frequency and the staleness of updates in the way that a smaller $\Bar{\gamma}$ results in less stale updates and smaller $K_{i,n}$, and vice versa.

We summarize both the training procedures at the server and at the clients by the proposed AsyncFedED in Algorithm \ref{algorithm1} and Algorithm \ref{algorithm2}, respectively. We note that the server stores a sequence of all the versions of the global models from the beginning of the training until to the current time slot,  referred to as \textit{Global Model Iteration Sequence} (GMIS), where one can find the stale model weights by the iteration index and calculate the staleness of the arrive updates.
As the other federated learning framework, the local training at each client is independent from the server and other clients given the initial global weights. Therefore, each client can use any optimizer and local learning rate of its choice as long as it upload the pseudo gradients to the server after each iteration of local training.

\section{Theoretical Analysis}\label{theoreticalresults}
\subsection{Preliminaries}\label{assumptions}
We first present the necessary assumptions in order to prove the convergence of the proposed AFL approach, listed in the following. 

\paragraph{Assumption 1}\label{assumption1}(L-Lipschitz gradient). \textit{For any $i\in[m]$, $f_i$ is non-convex and L-smooth, i.e., $\|\nabla f(x_1)- \nabla f(x_2)\| \leq L\|x_1-x_2\|$, $\forall x_1, x_2 \in \mathbb{R}^d$,.}

\paragraph{Assumption 2}\label{assumption2}(Unbiased local gradient and bounded local variance). \textit{For each client, the stochastic gradient $\nabla f_{i}(x;\xi_{i,k})$ is unbiased, i.e., $\mathbb{E}_{\xi_{i,k}}\left[{\nabla f_{i}(x;\xi_{i,k}) | x}\right] = \nabla f_i(x)$, $\forall i\in[m]$, and the variance of stochastic gradients are bounded by $\sigma_l$: $\left\|\nabla f_{i}\left(x; \xi_{i,k}\right)-\nabla f_{i}(x)\right\|^{2} \leq \sigma _{l}^{2}$, $\forall x \in \mathbb{R}^{d}, \xi _{i,k} \in \xi_i, i\in[m], k \in \{1,2,\cdots,K\}$.}

\paragraph{Assumption 3}\label{assumption3}(Global Variance). The discrepancy between local loss function gradient $\nabla f_i$ and global loss function gradient $\nabla F$ is bounded by $\sigma _g$: $\left\|\nabla f_{i}\left(x\right)-\nabla F(x)\right\|^{2} \leq \sigma _{g}^{2}$, for $\forall x\in\mathbb{R}^{d}$, $i\in [m]$.


\paragraph{Assumption 4}\label{assumption4}(Bounded Staleness). \textit{ Assume that $\gamma(i,\tau) \leq \Gamma, i\in[m],\tau\leq t$, where $\Gamma$ is the upper limit of $\gamma(i,\tau)$.}

Assumption 1 is a common assumption in optimization problems\cite{openproblem,AdaptiveFL,fedthroretical,goldstein1977optimization}. Assumption 2 and 3 are widely used in analysing federated learning\cite{FedBuff,FedProx,FedOPT,FedAvgConvergence} to bound the variance caused by stochastic gradients and statistical heterogeneity. We note that $\sigma_g$ quantifies the difference between the local loss function $f_i(\cdot)$ and the global loss function $F(\cdot)$. In Assumption 4, we bound the maximum of staleness $\gamma(i,\tau)$ of the gradient update by any client, which is easily achieved by simply discarding any update that is older than the given threshold.
Also note that we have also relaxed the assumption that $\left\|\nabla f_i (x) \right\|^2 \leq G $, usually made in existing works\cite{FedBuff,FedAvgConvergence}, which may not be realistic in some settings with strong convex loss functions\cite{personalized,khaled2019first,nguyen2019new}.

\subsection{Theoretical Results}
In this section, we present the theoretical analysis on the convergence of the proposed ASL approach. We further define $M:=\frac{1}{K}\sum_{k=1}^{K}\eta_{i,k}^2$, which is the average of the squared local learning rate over K epoches, $N :=  
\frac{1}{K}\sum_{k=1}^{K}\eta_{i,k}$, which is the average of the learning rate, $\Theta(\Gamma) = 1 - \left(\frac{\Gamma^2K\ln{(K+1)}}{3}+4K^2L^2M\Gamma^2 \right)$, where we recall the $\Gamma$ is the upper bound of the staleness. When Assumption 1 to 4 mentioned above hold, we have the following theorems and corollary. The detailed proof of which are presented in Appendix \ref{proof details} due to the limitation of space.

\paragraph{Theorem 1} (The upper bound of the local model drift\footnote{Similar conclusions can be found in \cite{FedOPT,personalized} where they consider a fixed local learning rate. Our result relaxes the assumption on the local learning rate and also achieve a tighter upper bound where the client local drift grows at a linear rate with local epoch $k$ instead of $k^2$ as in \cite{FedOPT}.}). 
\textit{If $\eta_{i,k}^2 \leq \frac{1}{6(2k+1)kL^2}$, after client $i$ performing $k$ epochs SGD with the initial model weights $x_{t-\tau}$, the local model weight drift away from $x_{t-\tau}$ is bounded by:
    \begin{equation}
        \mathbb{E}\left[\left\|x_{t-\tau, k}^i-x_{t-\tau}\right\|^{2}\right]
        \leq \frac{k}{2L^2}\left(\sigma_g^2+\mathbb{E}\left[\left\|\nabla F(x_{t-\tau}) \right\|^2\right]\right) +\frac{(k+1)}{24L^2}\sigma_l^2. 
    \end{equation}}\label{local drift}
\begin{proof}
    Given in Appendix \ref{proof of Theorem 1}.
\end{proof}

\paragraph{Corollary 1}(The upper bound of local gradients). \textit{When all conditions in Theorem 1 are satisfied, and $\Gamma^2 < \frac{3}{12K^2L^2M+K\ln{(K+1)}}$,
the local gradients $\Delta_i(x_{t-\tau,K})$ derived by client $i$ after $K$ epochs of training is bounded by:
    \begin{equation}
    \begin{aligned}
        \mathbb{E}\left[\left\| \Delta_i(x_{t-\tau,K}) \right\|^2\right] &\leq \frac{12K^2L^2M+K\ln{(K+1)}}{3L^2\Theta(\Gamma) }\mathbb{E}\left[\left\|\nabla F(x_t) \right\|^2  \right]
        \\
        &\qquad + \frac{72K^2L^2M + K\ln{(K+1)}}{72L^2\Theta(\Gamma)} \sigma_l^2 + \frac{24K^2L^2M+K\ln{(K+1)}}{6L^2\Theta(\Gamma)} \sigma_g^2
    \end{aligned}
    \end{equation}}\label{pseudo bounded}

\begin{proof}
    Given in Appendix \ref{proof of Corollary 1}.
\end{proof}
\paragraph{Theorem 2}(Convergence result). \textit{When $\eta_{i,k}$ and $\Gamma$ satisfies the conditions in Theorem 1 and Corollary 1, and  $\eta_g \leq \frac{1}{NKL}$, the global model update is then bounded by\footnote{It is the ergodic convergence and commonly used in non-convex optimization\cite{ghadimi2013stochastic,asgd,FedOPT,FedBuff}.}:
    \begin{equation}
        \begin{aligned}
            \frac{1}{T}\sum_{t=1}^{T} \mathbb{E}\left(\left\|\nabla F(x_t) \right\|^2\right) &\leq \frac{2(F(x_1)-F(x^{*}))}{\eta_g KNT}
            \\
            &\qquad + \frac{\eta_g}{N} \cdot\left( \frac{\ln{(K+1)} + 4K}{72L\Theta(\Gamma)}\sigma_l^2  +
            \frac{3\ln{(K+1)}+4K}{18L\Theta(\Gamma)}\sigma_g^2 \right)\label{convergence11}
        \end{aligned}
    \end{equation}
    where $x^{*}$ denote the optimal model weights. If $\eta_g = \mathcal{O}\left( \frac{1}{\sqrt{KNT}} \right)$, we have the following ergodic converge rate for the iteration of the proposed AsyncFedED algorithm:
    \begin{equation}
        \begin{aligned}
        \frac{1}{T}\sum_{t=1}^{T} \mathbb{E}\left(\left\|\nabla F(x_t) \right\|^2\right)
        &\leq \mathcal{O}\left( \frac{F(x_1)-F(x^{*})}{\sqrt{KNT}} \right) 
        + \mathcal{O}\left( \frac{\sigma_l^2+\sigma_g^2}{K^{\frac{3}{2}}N^{\frac{3}{2}}\sqrt{T}\Theta(\Gamma)} \right)\label{Convergence result}
        \end{aligned} 
    \end{equation}}\label{convergence result}
\begin{proof}
    Given in Appendix \ref{proof of Theorem 2}.
\end{proof}

\subsection{Discussions}
\paragraph{Effect of no. local epochs $\boldsymbol{K}$} We can see from Theorem 1 and Corollary 1, the upper bound of local model drift and local pseudo gradients are increasing with the rate of $\mathcal{O}(K)$ and $\mathcal{O}(K^2)$, which is in line with the fact that the more local epochs the clients run at each round, the more significant the local model update will be. However, we note that Corollary 1 is based on the condition $\Gamma^2 < \frac{3}{12K^2L^2M+K\ln{(K+1)}}$, which enforces the upper bound of the staleness $\Gamma$ to decrease with the increase of $K$. We also note that the convergence rate given in \eqref{convergence result} is not a monotonic decreasing function of $K$ due to the presence of $K$ in $\Theta(\Gamma)$. Hence, it is of critical importance to find a proper $K$ in order to guarantee the convergence of the proposed algorithm, which motivates the proposed adaptive scheme introduced in this paper.

\paragraph{Convergence rate in non-adaptive case} In non-adaptive case when $\eta_g$ is a fixed constant, the expectation given in (\ref{convergence11}) does not go to zero with the increase of $T$. Therefore, we can not prove the convergence of the AFL algorithm theoretically. In another non-adaptive case when $K$ is a constant, we can achieve a convergence rate of $\mathcal{O}(\frac{1}{\sqrt{T}})$ given by \eqref{convergence result}.

\section{Numerical results}\label{experimentalresults}

\subsection{Experimental Setting}\label{setting}
\paragraph{Training task and models}
We chose three most commonly used  Non-IID datasets for federated learning to evaluate the proposed algorithm: Synthetic-1-1\cite{li2019fair}, FEMNIST\cite{leaf} and Shakespeare text data\cite{FedAvg}. We also sample $10\%$ of each dataset randomly for testing and the rest for training. We employ different types of network structures were used for different data sets that accomplishes different tasks. Specifically, we use MLP for Synthetic-1-1, CNN for FEMNIST and RNN for Shakespeare text data. The models' structures are described in Appendix \ref{model structure}. 

\paragraph{Baselines}
We compare the proposed AsyncFedED algorithm with two SFL algorithms: FedAvg\cite{FedAvg} and FedProx\cite{FedProx}, as well as two AFL algorithms: FedAsync+Constant\cite{AsyncFed} and FedAsync+Hinge\cite{AsyncFed}. We tune hyperparameters by grid searching, where the corresponding sweep range and the values we chose are shown in Appendix \ref{tune hyper parameters}.

\paragraph{Training environment}
The hardware we run our experiments and the communication protocol we apply are described in \ref{training environment}. In addition, in order to simulate time-varying property of clients, we assume a suspension probability $P$ such that each client has a certain probability $P$ of being suspended, and the hang time is a random value distributed with regards to the maximum running time.

\subsection{Comparison results}

In this section, we present the comparison results of the proposed AsyncFedED with the baseline algorithms. We note that all the results presented here are averaged over five independent experimental runs. All the codes and raw results data can be found online\footnote{\href{https://github.com/Qiyr0/AsyncFedED}{https://github.com/Qiyr0/AsyncFedED}}.

\subsubsection{Convergence rate} 
\begin{figure}[H]
\centering
\subfigure[Synthetic-1-1]{\includegraphics[width=0.32\textwidth]{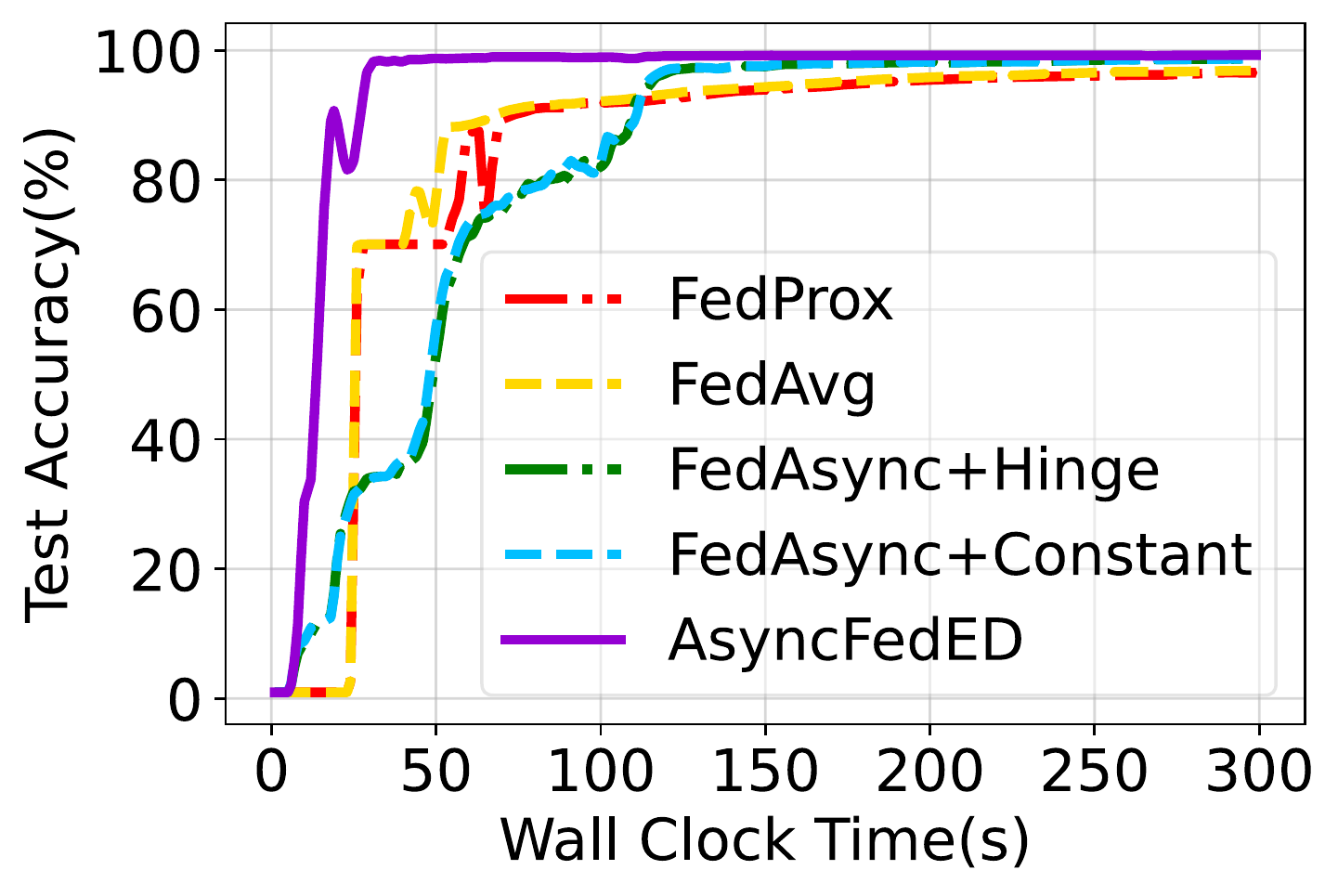}}
\subfigure[FEMNIST]{\includegraphics[width=0.32\textwidth]{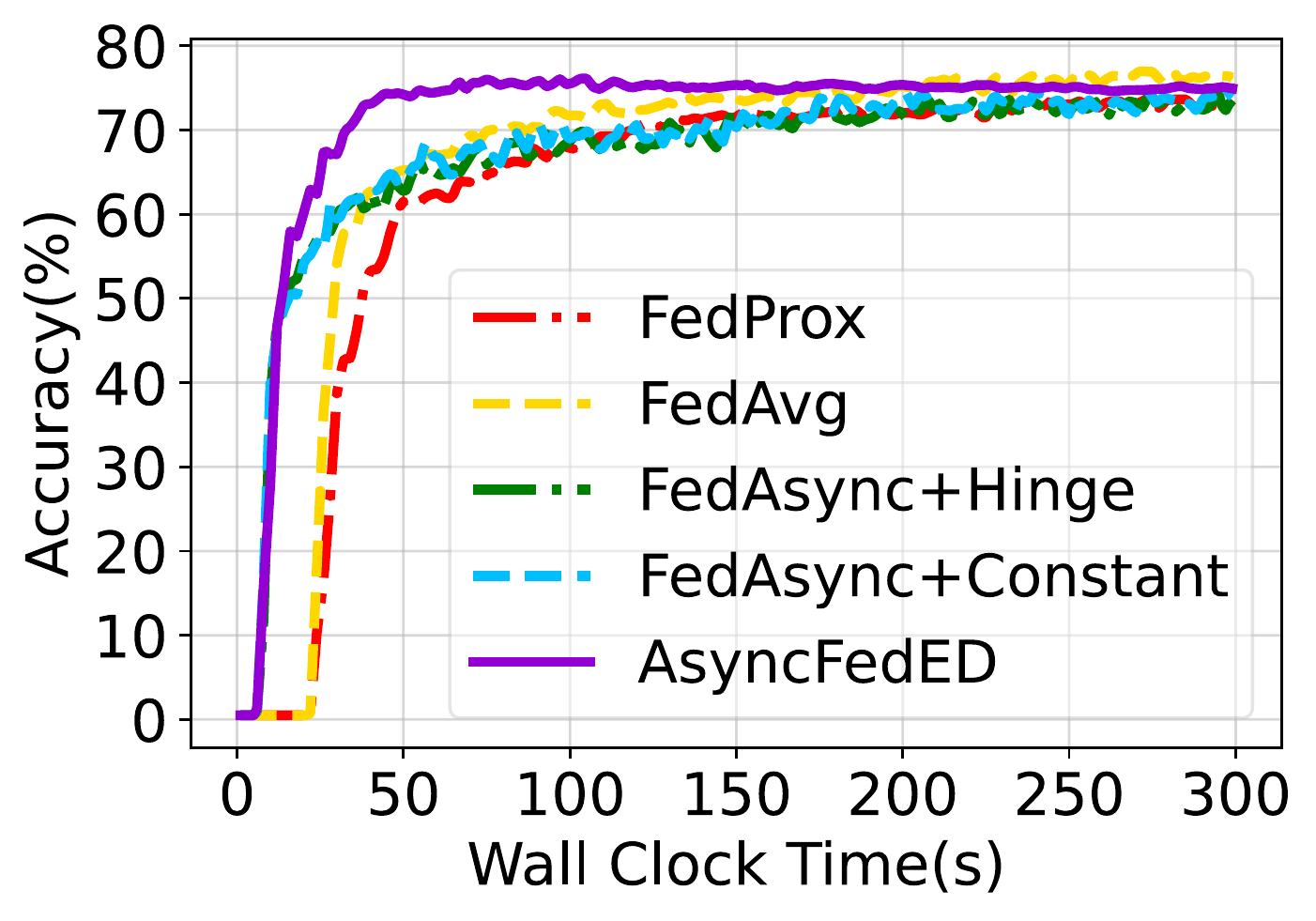}}  
\subfigure[ShakeSpeare text data]{\includegraphics[width=0.32\textwidth]{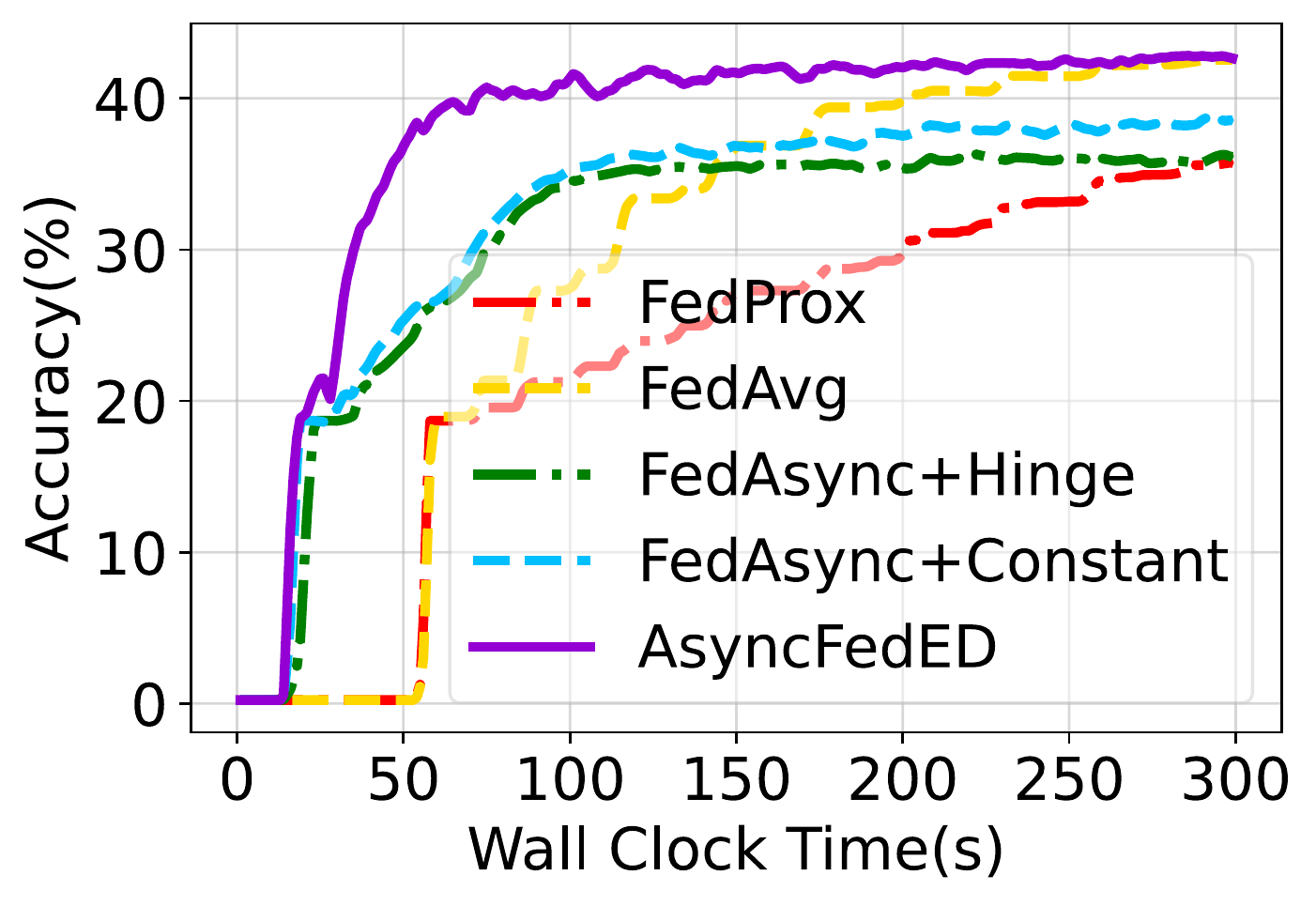}} 
\centering
\caption{Test accuracy v.s. training time curves when the suspension probability is set to $P=0.1$. } 
\label{convergenceProcess}
\end{figure}

We present the test accuracy v.s. running time curves in Fig. \ref{convergenceProcess} to numerically evaluate the convergence rate of different scheme. We set the client suspension probability to be $P=0.1$ while results with other values of $P$ can be found in Appendix \ref{allresults}. We can observe that AsyncFedED converges remarkably faster than all the baseline algorithm on all the three considered tasks.

\subsubsection{Robustness against clients suspension} \label{robustness}
\begin{figure}[H]
\centering
    \subfigure[Synthetic-1-1]{
        \centering
        \begin{minipage}[t]{0.31\textwidth}
            \centering
            \includegraphics[width=\textwidth]{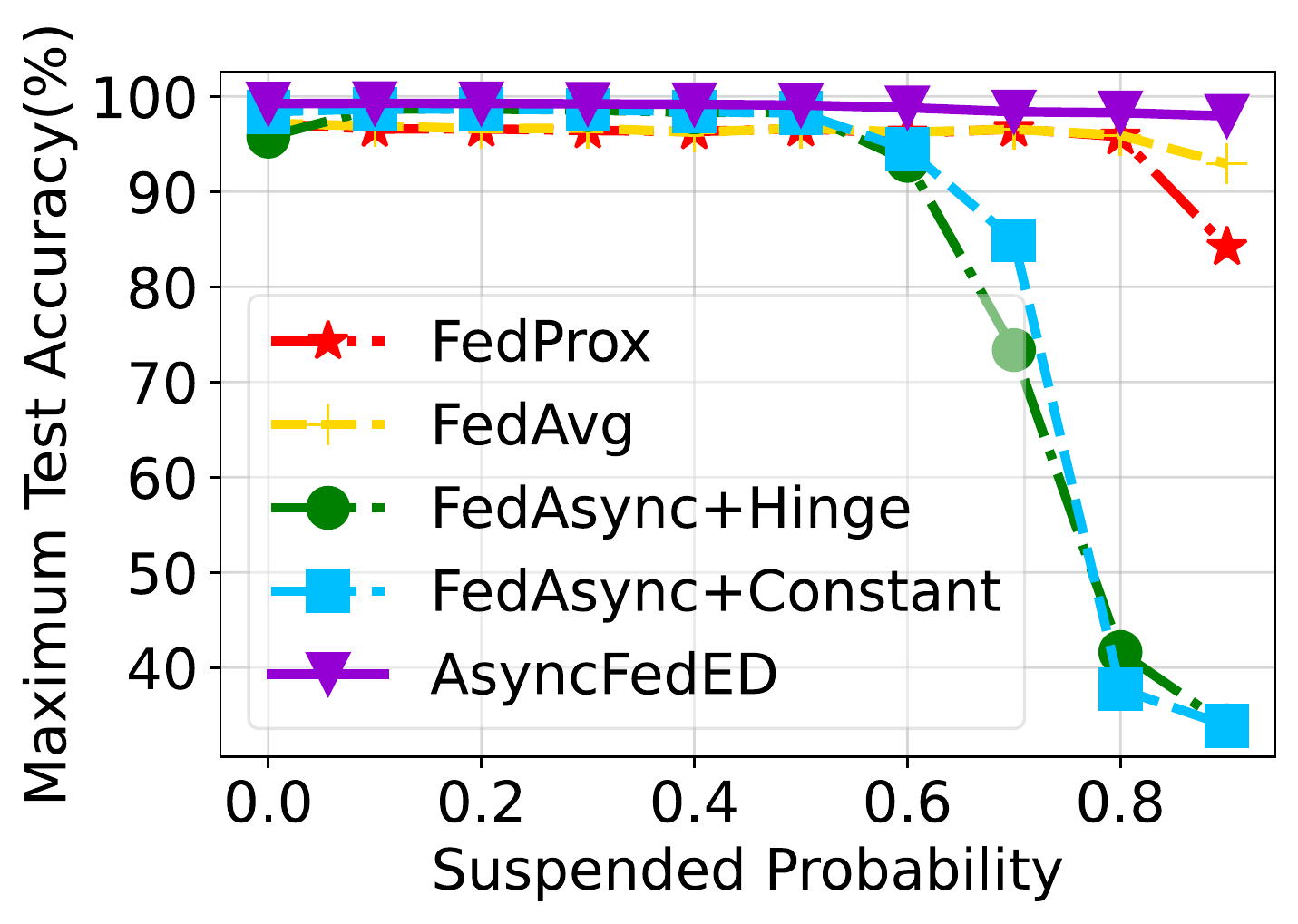} \\
            \includegraphics[width=\textwidth]{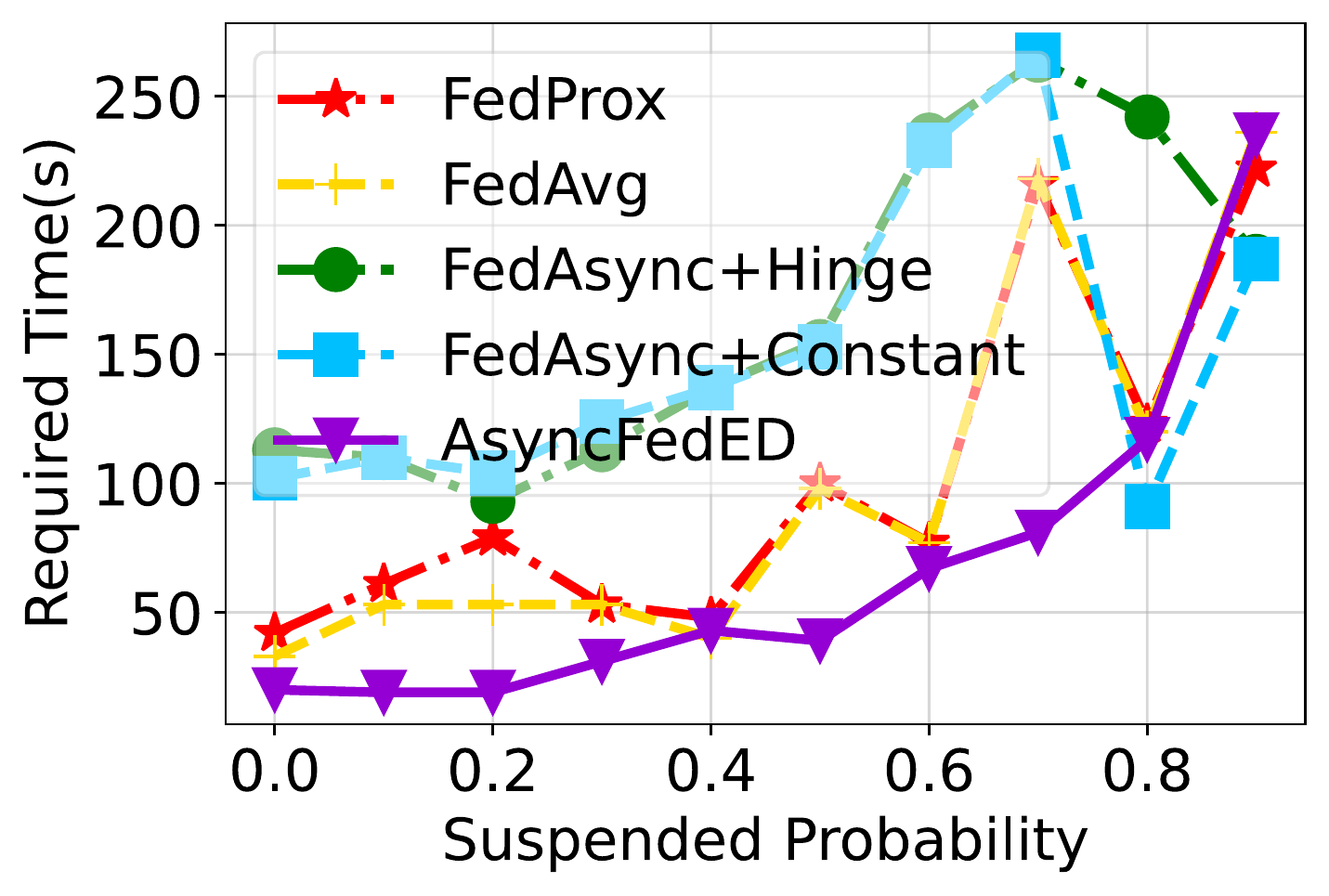}
        \end{minipage}
    }
    \subfigure[FEMNIST]{
        \centering
        \begin{minipage}[t]{0.31\textwidth}
            \centering
            \includegraphics[width=\textwidth]{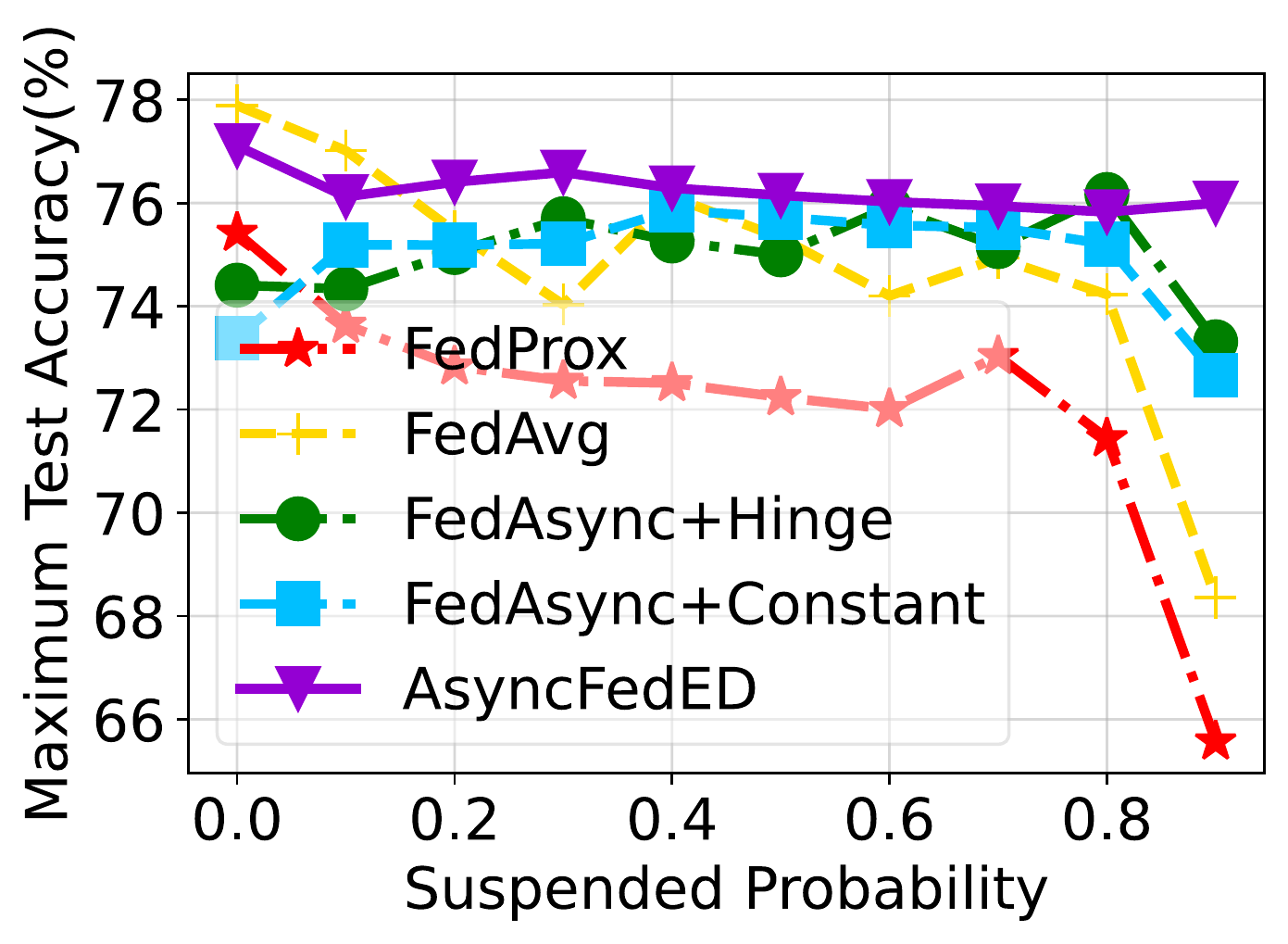} \\
            \includegraphics[width=\textwidth]{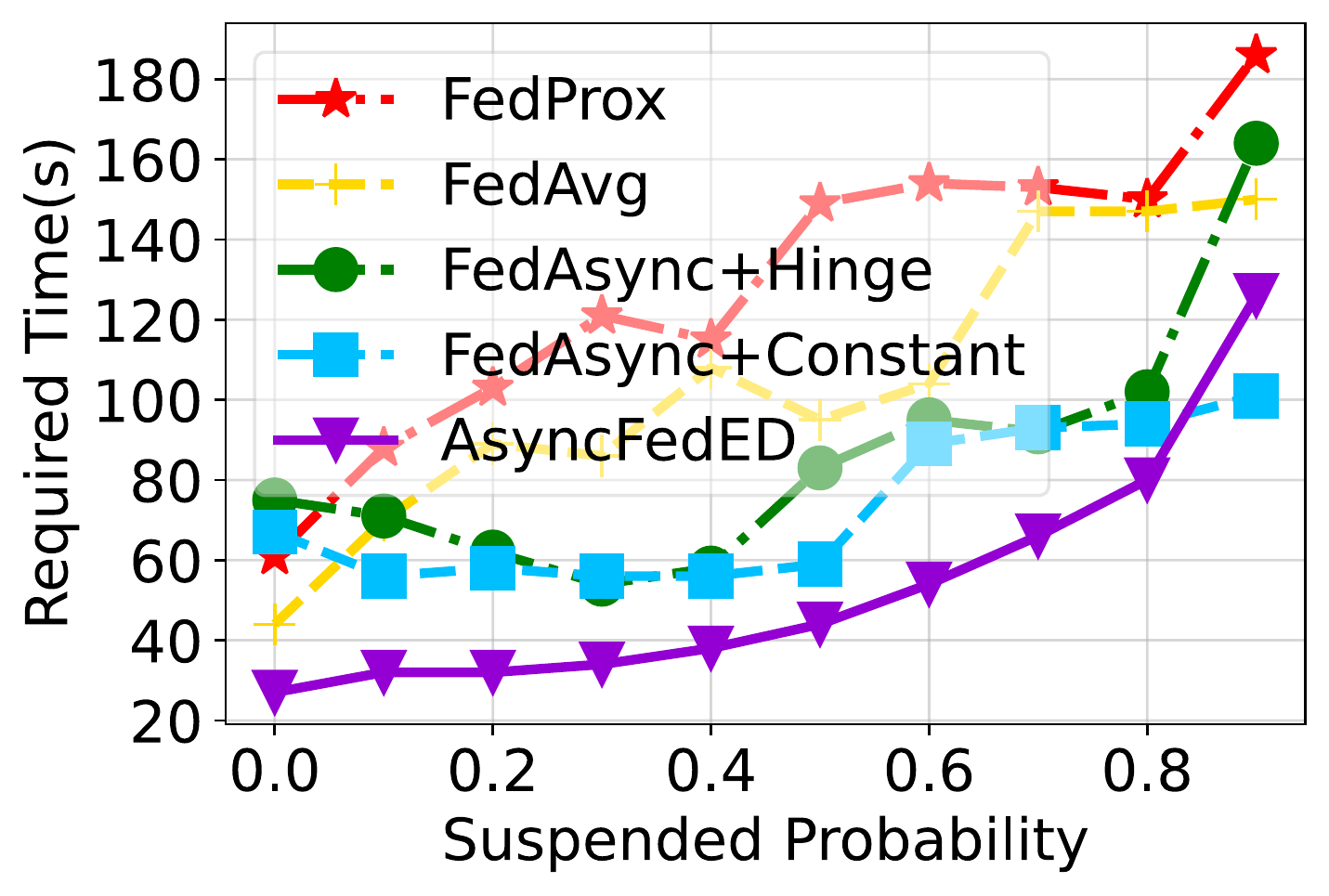}
        \end{minipage}
    }
    \subfigure[ShakeSpeare text data]{
        \centering
        \begin{minipage}[t]{0.31\textwidth}
            \centering
            \includegraphics[width=\textwidth]{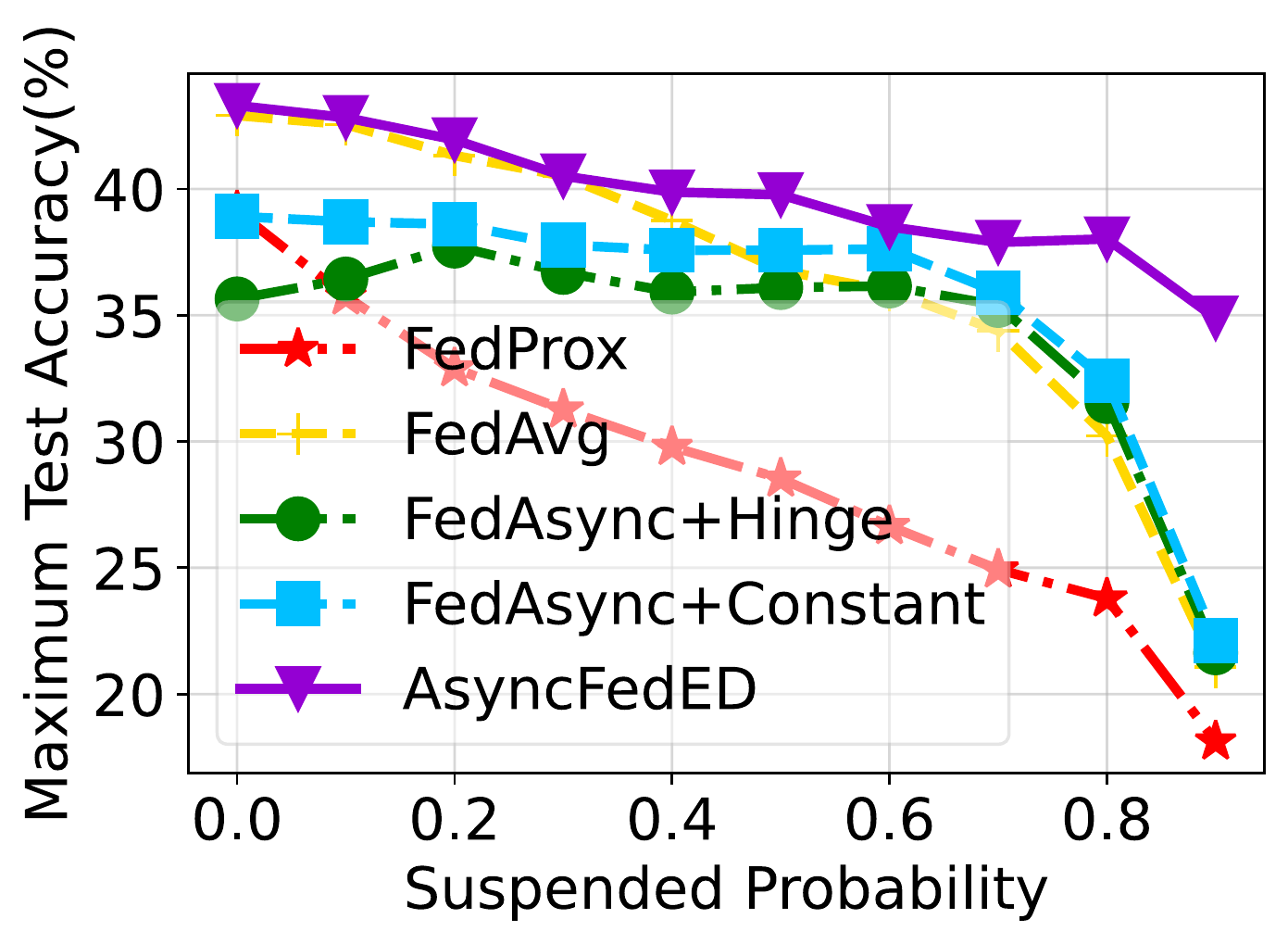} \\
            \includegraphics[width=\textwidth]{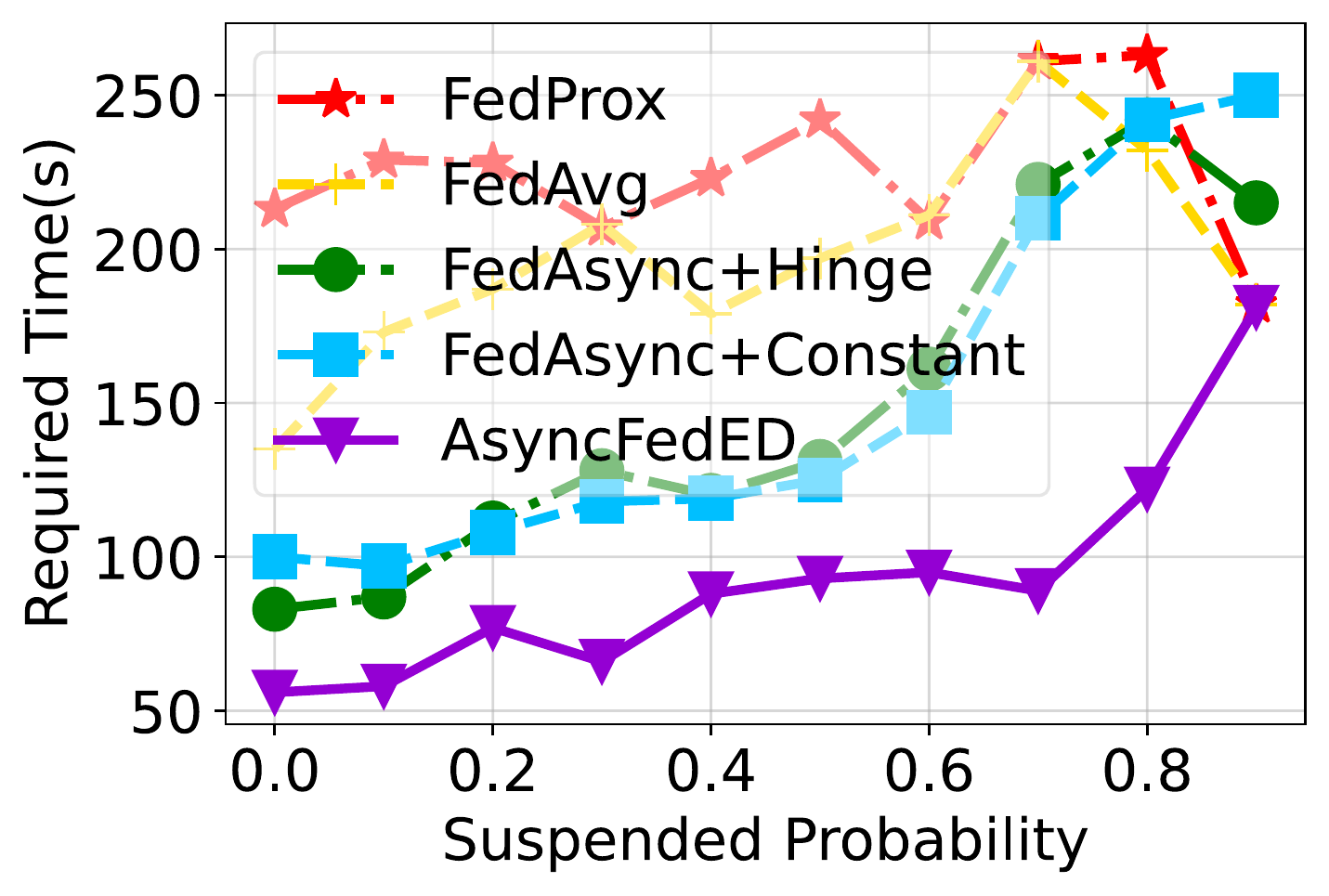}
        \end{minipage}
    }
\caption{The maximum test accuracy reached in 300s and the required time to reach 90\% of the maximum test accuracy with respect to different suspension probability $P$. } 
\label{robustnessResults}
\end{figure}

We present the maximum test accuracy achieved by different algorithm within a given time (300 seconds in these experiments) with regards to different $P$ in Fig. \ref{robustnessResults}, as well as the required time for each algorithm to reach 90\% of its maximum test accuracy, in order to evaluate the robustness of algorithms against clients' suspension. It can be observe from Fig. \ref{robustnessResults} that the proposed AsyncFedED outperforms the other baseline algorithm in terms of both the maximum test accuracy it reaches within a given time and the required time to reach a certain accuracy. We also note that the performance of the proposed algorithm remain quite stable when $P$ increases, while the existing algorithms, AsyncFed+Hinge and AsyncFed+Constant, declines drastically when $P$ is large, which validates the robustness of the proposed algorithm against the suspension of clients.

\subsubsection{Effectiveness of adaptive $K$}\label{EffectAdaptiveK}
\begin{figure}[H]
\centering
\subfigure[Synthetic-1-1]{\includegraphics[width=0.32\textwidth]{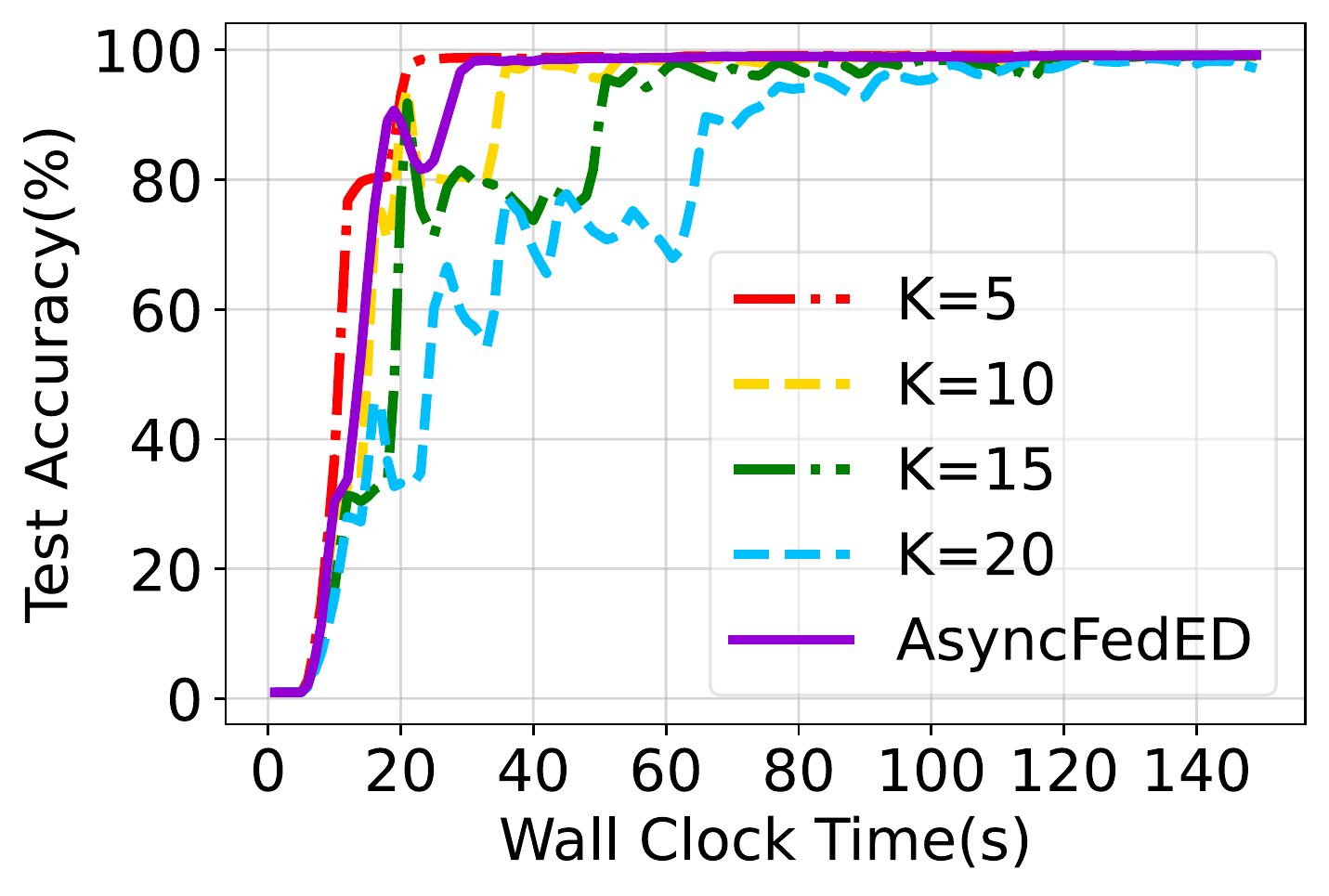}}
\subfigure[FEMNIST]{\includegraphics[width=0.32\textwidth]{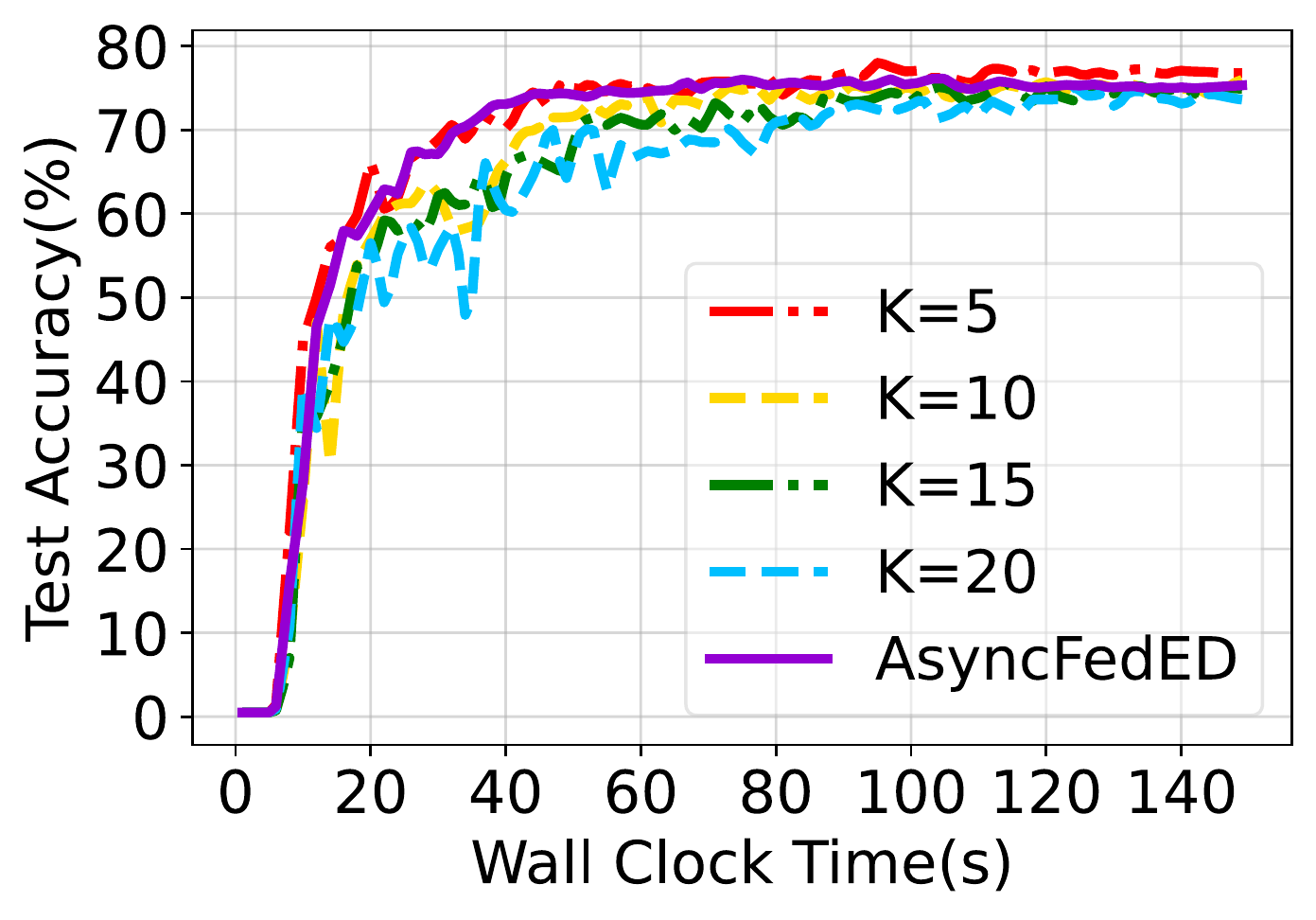}}  
\subfigure[ShakeSpeare text data]{\includegraphics[width=0.32\textwidth]{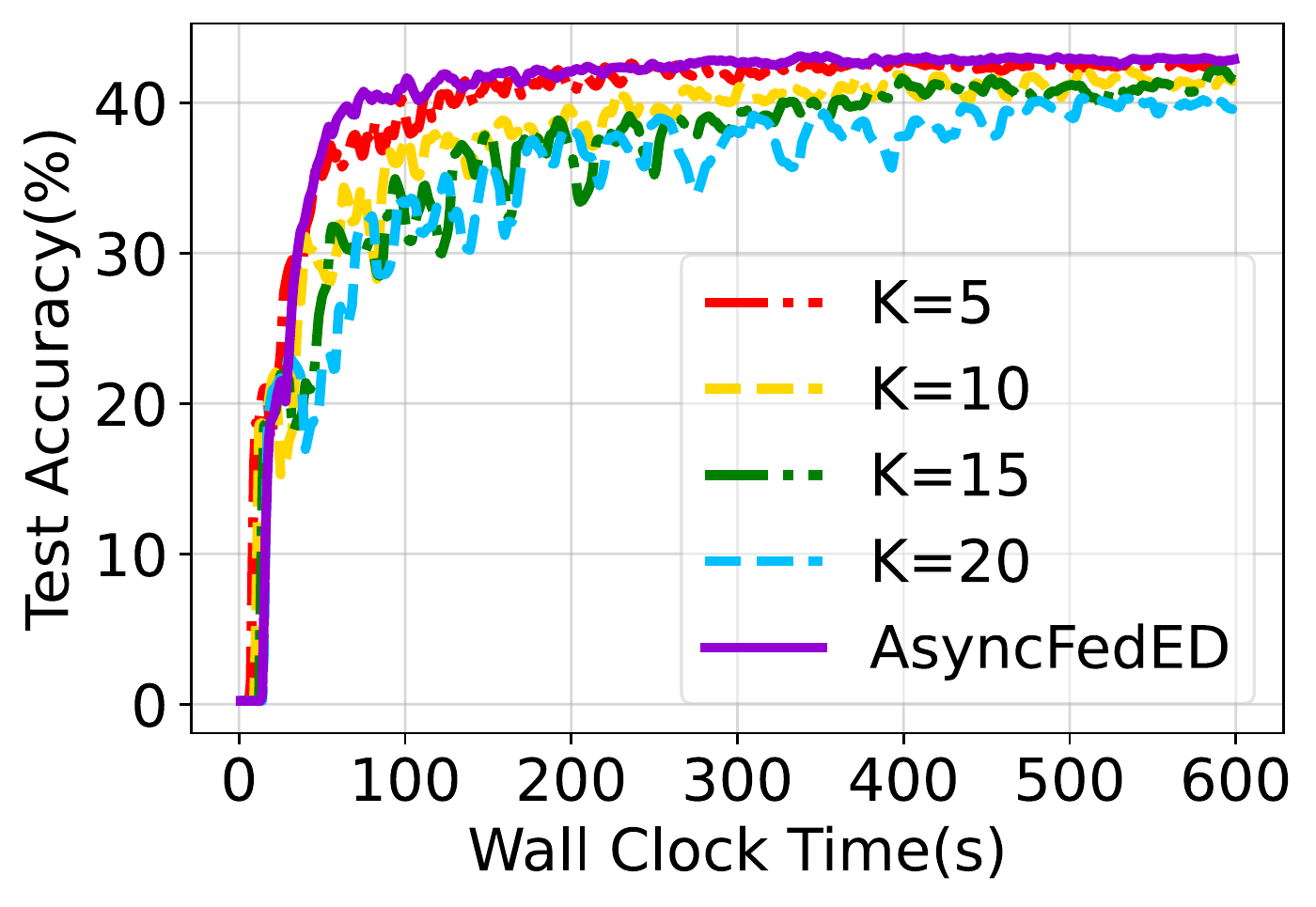}} 
\centering
\caption{Effectiveness of adaptive $K$. AsyncFedED shows more smooth convergence progress in all three data sets.} 
\label{effectiveofK}
\end{figure}

To verify the effectiveness of adaptive $K$, we compare the performance of the proposed AsyncFedED algorithm with the adaptive $K$ scheme and the constant $K$ scheme with $K$ is set to be 5, 10, 15, 20. We present the test accuracy v.s. training time curves in  Due to the setting of initial local epoch $K=10$ in Fig. \ref{effectiveofK} AsyncFedED and constant local, where we can see that the adaptive $K$ scheme converges faster and smoother than the fixed $k$ schemes, which proves the effectiveness of adaptive no. of local epochs. 

\section{Conclusions}\label{conclusions}
In this paper, we proposed an AFL framework with  adaptive weight aggregation algorithm referred to as AsyncFedED, in which the aggregation parameters are set adaptively according to the staleness of the received model updates.  We also proposed an update rule of the number of local epochs, which automatically balances the staleness of the model updates from different clients despite the devices heterogeneity. We have proven the convergence of the proposed algorithm theoretically and numerically shown that the proposed AFL framework outperform the existing methods in terms of convergence rate and the robustness to the device failures. Although the considered AFL framework preserves data privacy by training in a collaborative way without out data sharing, privacy still may leak through the uploaded model updates to the server. Moreover, although the proposed algorithm is robust to the device failure, a malicious client could still hurt the convergence of the learning model. Therefore, a selection mechanism used to identify a group of safe devices that protect data privacy and model convergence is a future direction to investigate.

{
\bibliography{reference}
}

\section*{Checklist}


\begin{enumerate}

\item For all authors...
\begin{enumerate}
  \item Do the main claims made in the abstract and introduction accurately reflect the paper's contributions and scope?
    \answerYes{}
  \item Did you describe the limitations of your work?
    \answerYes{See section \ref{conclusions}.}
  \item Did you discuss any potential negative societal impacts of your work?
    \answerYes{See section \ref{conclusions}.}
  \item Have you read the ethics review guidelines and ensured that your paper conforms to them?
    \answerYes{}
\end{enumerate}

\item If you are including theoretical results...
\begin{enumerate}
  \item Did you state the full set of assumptions of all theoretical results?
    \answerYes{See Section \ref{assumptions}.}
\item Did you include complete proofs of all theoretical results?
    \answerYes{See Appendix \ref{proof details}.}
\end{enumerate}

\item If you ran experiments...
\begin{enumerate}
  \item Did you include the code, data, and instructions needed to reproduce the main experimental results (either in the supplemental material or as a URL)?
    \answerNo{The raw result data was included online but the code will be include in the camera ready version.}
  \item Did you specify all the training details (e.g., data splits, hyperparameters, how they were chosen)?
    \answerYes{See Section \ref{setting} and Appendix \ref{tune hyper parameters}.}
    \item Did you report error bars (e.g., with respect to the random seed after running experiments multiple times)?
    \answerYes{All experiments are repeated at least five times.}
    \item Did you include the total amount of compute and the type of resources used (e.g., type of GPUs, internal cluster, or cloud provider)?
    \answerYes{See Appendix \ref{training environment}}
\end{enumerate}

\item If you are using existing assets (e.g., code, data, models) or curating/releasing new assets...
\begin{enumerate}
  \item If your work uses existing assets, did you cite the creators?
    \answerYes{}
  \item Did you mention the license of the assets?
    \answerYes{}
  \item Did you include any new assets either in the supplemental material or as a URL?
    \answerNA{}
  \item Did you discuss whether and how consent was obtained from people whose data you're using/curating?
    \answerNA{}
  \item Did you discuss whether the data you are using/curating contains personally identifiable information or offensive content?
    \answerNA{}
\end{enumerate}

\item If you used crowdsourcing or conducted research with human subjects...
\begin{enumerate}
  \item Did you include the full text of instructions given to participants and screenshots, if applicable?
    \answerNA{}
  \item Did you describe any potential participant risks, with links to Institutional Review Board (IRB) approvals, if applicable?
    \answerNA{}
  \item Did you include the estimated hourly wage paid to participants and the total amount spent on participant compensation?
    \answerNA{}
\end{enumerate}

\end{enumerate}


\appendix
\newpage
\begin{center}
    \begin{minipage}[t]{0.98\textwidth}
    \centering
        \LARGE \textbf{Appendix to "AsyncFedED: Asynchronous Federated Learning with Euclidean Distance based Adaptive Weight Aggregation"}
    \end{minipage}
\end{center}
\vspace{30pt}

\section{Proof details}\label{proof details}
For ease of presentation, we rewrite the $\nabla f_i(\cdot;\xi)$ as $g(\cdot)$, $\mathbb{E}_{\xi_{i,k}}\left[{\nabla f_{i}(x;\xi_i) | x}\right]$ as $\mathbb{E}_{\xi_{i,k}}[{g(x)}]$, and $x_{t-\tau,k}^i$ as $x_{t-\tau,k}$. We further define $M:=\frac{1}{K}\sum_{k=1}^{K}\eta_{i,k}^2$, which is the average of the squared local learning rate over $K$ epoches, $N :=  
\frac{1}{K}\sum_{k=1}^{K}\eta_{i,k}$, which is the average of the learning rate, $\Theta(\Gamma) := 1 - \left(\frac{\Gamma^2K\ln{(K+1)}}{3}+4K^2L^2M\Gamma^2 \right)$, where we recall the $\Gamma$ is the upper bound of the staleness.
\subsection{Auxiliary lemmas}
We first present the following auxiliary lemmas, which are essential for the convergence analysis. 
\paragraph{Lemma 1}For any vector $x_i \in \mathbb{R}^{d},i=1,2,...,n$, we have:
\begin{equation}
    \left\|\sum_{i=1}^{n}x_i \right\|^2 \leq n \sum_{i=1}^{n}\left\|x_i \right\|^2.
\end{equation}
\begin{proof}
    This lemma follows the definition of Jensen's inequality.
\end{proof}

\paragraph{Lemma 2} \textit{If Assumption 1 is satisfied, the global loss function $F(x)$ is non-convex and has L-Lipschitz gradient.}
\begin{proof}
    This lemma straightforwardly follows from the definition of $F(x)=\frac{1}{m}\sum_{i=1}^{m}f_i(x)$.
\end{proof}

\paragraph{Lemma 3} \textit{Based on Assumption 1 and 4, the global gradient at $x_{t-\tau}$ can be bounded by:
\begin{equation}
    \left\| \nabla F(x_{t-\tau}) \right\|^2 \leq 2L^2\Gamma^2 \left\| \Delta_i(x_{t-\tau,K}) \right\|^2 + 2 \left\| \nabla F(x_{t}) \right\|^2.
\end{equation}}
\begin{proof}
\begin{subequations}
    \begin{align}
    \left\| \nabla F(x_{t-\tau}) \right\|^2 & = \left\| \nabla F(x_{t-\tau}) - \nabla F(x_{t}) + \nabla F(x_{t}) \right\|^2
    \\
    &\leq 2L^2\left\| x_{t} - x_{t-\tau} \right\|^2 + 2 \left\| \nabla F(x_{t}) \right\|^2 \label{15b}
    \\
    &\leq 2L^2\Gamma^2 \left\| \Delta_i(x_{t-\tau,K}) \right\|^2 + 2 \left\| \nabla F(x_{t}) \right\|^2\label{15c}.
\end{align}
\end{subequations}
where (\ref{15b}) results from Lemma 1 and Assumption 1, and (\ref{15c}) can be derived from Assumption 4: $\gamma(i,\tau) = \dfrac{\|x_{t} - x_{t-\tau} \|}{\|\Delta_i(x_{t-\tau,K})\|} \leq \Gamma$.
\end{proof}

\subsection{Proof of Theorem 1}\label{proof of Theorem 1}
After client $i$ downloading the global model $x_{t-\tau}$, it performs one SGD step given by:
\begin{equation}
    x_{t-\tau,k} = x_{t-\tau,k-1} - \eta_{i,k}\cdot g(x_{t-\tau,k-1}),
\end{equation}
where $k \geq 1$. From the fact that $\| a+b\|^2=\| a\|^2+\| b\|^2+2\langle a,b \rangle$ and take the expectation with respect to the randomness of mini-batch, we have
\begin{subequations}\label{submitR_1}
    \begin{align}
    & \mathbb{E}_{\xi_{i,k}}\left[ \left\|x_{t-\tau, k}-x_{t-\tau}\right\|^{2}\right] 
    \\
    = &\mathbb{E}_{\xi_{i,k}}\left[\left\|x_{t -\tau, k-1}-\eta_{i,k} \cdot g\left(x_{t-\tau, k-1} \right)-x_{t-\tau}\right\|^{2} \right]
    \\
    = &\mathbb{E}_{\xi_{i,k}}\left[\left\| x_{t-\tau, k-1}-x_{t-\tau}-\eta_{i,k} \nabla f\left(x_{t-\tau, k-1}\right)  +\eta_{i,k} \nabla f\left(x_{t-\tau, k-1}\right) - \eta_{i,k} g\left(x_{t-\tau, k-1}\right) \right \|^{2} \right]
    \\
    = &\mathbb{E}_{\xi_{i,k}}\left[ \left\|x_{t-\tau, k-1}-x_{t-\tau}-\eta_{i,k} \nabla f\left(x_{t-\tau, k-1}\right)\right\|^2 + \left\|\eta_{i,k} \nabla f\left(x_{t-\tau, k-1}\right) - \eta_{i,k} g\left(x_{t-\tau, k-1}\right) \right\|^2 \right] \label{17d}
    \\
    &\leq \mathbb{E}_{\xi_{i,k}}\left[\underbrace{\left\|x_{t-\tau, k-1}-x_{t-\tau}-\eta_{i,k} \nabla f\left(x_{t-\tau, k-1}\right)\right\|^2}_{R_1}\right] 
    + \eta_{i,k}^2\sigma_l^2,    \label{17e}
    \end{align}
\end{subequations}
where (\ref{17d}) is due to the unbiased stochastic gradients in Assumption 2, such that $\mathbb{E}_{\xi_{i,k}}[g(x_{t-\tau})] = \nabla f(x_{t-\tau})$:
\begin{subequations}
    \begin{align}
        &2\mathbb{E}_{\xi_{i,k}}\left[ \langle x_{t-\tau, k-1}-x_{t-\tau}-\eta_{i,k} \nabla f\left(x_{t-\tau, k-1}\right),\eta_{i,k} \nabla f\left(x_{t-\tau, k-1}\right) - \eta_{i,k} g\left(x_{t-\tau, k-1}\right)\rangle\right]
        \\
        = &2\langle x_{t-\tau, k-1}-x_{t-\tau}-\eta_{i,k} \nabla f\left(x_{t-\tau, k-1}\right),  \eta_{i,k} \mathbb{E}_{\xi_{i,k}}\left[(\nabla f\left(x_{t-\tau, k-1}\right) - \eta_{i,k} g\left(x_{t-\tau, k-1}\right))\right]\rangle
        \\
        = &0.
    \end{align}
\end{subequations}
(\ref{17e}) also results from Assumption 2: $\| \nabla f(x) - g(x) \|^2 \leq \sigma_l^2$.

Similarly, by $\| a+b\|^2=\| a\|^2+\| b\|^2+2\langle a,b \rangle$, we have:
\begin{subequations}
    \begin{align}
    \mathbb{E}_{\xi_{i,k}}\left[R_1\right] &= \mathbb{E}_{\xi_{i,k}}\left[ \left\|x_{t-\tau, k-1}-x_{t-\tau}-\eta_{i,k} \nabla f\left(x_{t-\tau, k-1}\right)\right\|\right] 
    \\
    &= \mathbb{E}_{\xi_{i,k}}\left[\left\|x_{t-\tau, k-1}-x_{t-\tau}\right\|^2\right] +\eta_{i,k}^2\cdot\mathbb{E}_{\xi_{i,k}}\left[\underbrace{\left\|\nabla f(x_{t-\tau,k-1})\right\|^2}_{R_{11}}\right]    \notag
    \\
    &\qquad +\mathbb{E}_{\xi_{i,k}}\left[\underbrace{2\langle x_{t-\tau, k-1}-x_{t-\tau}, -\eta_{i,k}\cdot \nabla f(x_{t-\tau,k-1}) \rangle}_{R_{12}}\right].
    \end{align}
\end{subequations}

Based on Lemma 1 and Assumption 1 and 3, $R_{11}$ can be bounded by:
\begin{subequations}
    \begin{align}
    \mathbb{E}_{\xi_{i,k}}\left[ R_{11}\right]&= \mathbb{E}_{\xi_{i,k}}\left[\left\|\nabla f(x_{t-\tau,k-1})\right\|^2\right] 
    \\
    &= \mathbb{E}_{\xi_{i,k}}\left[\left\|\nabla f(x_{t-\tau,k-1})-\nabla f(x_{t-\tau})  +\nabla f(x_{t-\tau}) -\nabla F(x_{t-\tau})+\nabla F(x_{t-\tau})\right\|^2 \right]
    \\
    &\leq  \mathbb{E}_{\xi_{i,k}}\left[3L^2 \cdot \left\|x_{t-\tau, k-1}-x_{t-\tau}\right\|^{2}+3\sigma_g^2 + 3\left\|\nabla F(x_{t-\tau})\right\|^{2}\right].
    \end{align}
\end{subequations}

For the term of $R_{12}$,  we have:
\begin{subequations}\label{R12}
    \begin{align}
    \mathbb{E}_{\xi_{i,k}}\left[R_{12}\right] &= 2\mathbb{E}_{\xi_{i,k}}\left[\langle x_{t-\tau, k-1}-x_{t-\tau}, -\eta_{i,k}\cdot \nabla f(x_{t-\tau,k-1}) \rangle \right]
    \\
    &= 2\mathbb{E}_{\xi_{i,k}}[ \langle \frac{1}{\widetilde{K}}(x_{t-\tau, k-1}-x_{t-\tau}),
    \eta_{i,k}\widetilde{K} \left[\left(\nabla f(x_{t-\tau})-\nabla f(x_{t-\tau,k-1})\right) \right.\notag
    \\
    &\qquad \left. +\left(\nabla F(x_{t-\tau}) -\nabla f(x_{t-\tau})\right) - \nabla F(x_{t-\tau}) \right] \rangle ]
    \\
    &\leq \frac{1}{\widetilde{K}^2}\cdot\mathbb{E}_{\xi_{i,k}}\left[\left\|x_{t-\tau, k-1}-x_{t-\tau}\right\|^{2} \right] 
     + 3\eta_{i,k}^2L^2\widetilde{K}^2\cdot\mathbb{E}_{\xi_{i,k}}\left[\left\|x_{t-\tau, k-1}-x_{t-\tau}\right\|^{2} \right]    \notag
    \\
    &\qquad +3\eta_{i,k}^2\widetilde{K}^2\cdot\left(\sigma_g^2+\mathbb{E}_{\xi_{i,k}}\left[\left\|\nabla F(x_{t-\tau}) \right\|^2\right]\right),
    \end{align}
\end{subequations}
where $\Tilde{K}$ is an auxiliary variable that can be any value. (\ref{R12}) is derived due to the fact that $2\langle  a,b\rangle \leq \|a \|^2 + \| b\|^2$ and based on Lemma 1. Then with the upper bounds of $R_{11}$ and $R_{12}$, it yields from Eq.(\ref{submitR_1}):

\begin{equation}\label{overresult}
    \begin{aligned}
     \mathbb{E}_{\xi_{i,k}}\left[\left\|x_{t-\tau, k}-x_{t-\tau}\right\|^{2}\right] &\leq \left[1+\frac{1}{\widetilde{K}^2}+3(\widetilde{K}^2+1)\eta_{i,k}^2L^2\right]\cdot\mathbb{E}_{\xi_{i,k}}\left[\left\|x_{t-\tau, k-1}-x_{t-\tau}\right\|^{2}\right]
     \\
     &\qquad +3\left(\widetilde{K}^2+1\right)\eta_{i,k}^2\cdot\left(\sigma_g^2+\mathbb{E}_{\xi_{i,k}}\left[\left\|\nabla F(x_{t-\tau}) \right\|^2\right]\right) + \eta_{i,k}^2\sigma_l^2.
    \end{aligned}
\end{equation}

Let $\widetilde{K}^2=2k$ and $\eta_{i,k}^2\leq\dfrac{1}{3(\widetilde{K}+1)\widetilde{K}L^2}$, which leads to $\eta_{i,k}^2 \leq\dfrac{1}{6(2k+1)kL^2}$. Then, we rewrite Eq.(\ref{overresult}) by:
\begin{subequations}
    \begin{align}
    &\quad\:\: \mathbb{E}_{\xi_{i,k}}\left[\left\|x_{t-\tau, k}-x_{t-\tau}\right\|^{2}\right]
    \\
    &\leq \underbrace{(1+\frac{1}{k})}_{A_k}\cdot\mathbb{E}_{\xi_{i,k}}\left[\left\|x_{t-\tau, k-1}-x_{t-\tau}\right\|^{2}\right]
     +\underbrace{\frac{1}{2kL^2}\left[\sigma_g^2+\mathbb{E}_{\xi_{i,k}}\left[\left\|\nabla F(x_{t-\tau}) \right\|^2\right]\right] + \eta_{i,k}^2\sigma_l^2}_{B_k} 
    \\
    &= \sum_{j=1}^{k-1}\left(B_j\prod_{r=j+1}^{k}A_r \right)+B_k 
    \\
    &= \frac{k+1}{2L^2}\left(\sigma_g^2+\mathbb{E}_{\xi_{i,k}}\left[\left\|\nabla F(x_{t-\tau}) \right\|^2\right]\right)\cdot \sum_{j=1}^{k}\frac{1}{j}\cdot\frac{1}{j+1}  
    +(k+1)\sigma_l^2\cdot\sum_{j=1}^{k}\eta_{i,j}^2\cdot\frac{1}{j+1} 
    \\
    &\leq \frac{k+1}{2L^2}\left(\sigma_g^2+\mathbb{E}_{\xi_{i,k}}\left[\left\|\nabla F(x_{t-\tau}) \right\|^2\right]\right)\cdot \sum_{j=1}^{k}\frac{1}{j}\cdot\frac{1}{j+1} 
     +\frac{(k+1)}{6L^2}\sigma_l^2\cdot\sum_{j=1}^{k}\frac{1}{j(j+1)(2j+1)} 
    \\
    &\leq \frac{k+1}{2L^2}\left(\sigma_g^2+\mathbb{E}_{\xi_{i,k}}\left[\left\|\nabla F(x_{t-\tau}) \right\|^2\right]\right)\cdot \sum_{j=1}^{k}\frac{1}{j}\cdot\frac{1}{j+1} 
    +\frac{(k+1)}{6L^2}\sigma_l^2\cdot\sum_{j=1}^{k}\frac{1}{j(j+1)(j+2)} \label{23f}
    \\
    &= \frac{k}{2L^2}\left(\sigma_g^2+\mathbb{E}_{\xi_{i,k}}\left[\left\|\nabla F(x_{t-\tau}) \right\|^2\right]\right) +\frac{(k^2+3k)}{24(k+2)L^2}\sigma_l^2
    \\
    &\leq \frac{k}{2L^2}\left(\sigma_g^2+\left\|\nabla F(x_{t-\tau}) \right\|^2\right) +\frac{(k+1)}{24L^2}\sigma_l^2,\label{23h}
    \end{align}
\end{subequations}

where (\ref{23f}) results from the fact that $\frac{1}{j(j+1)(2j+1)} \leq \frac{1}{j(j+1)(j+2)}$, $\forall j \geq 1$, and \eqref{23h} is due to the fact that $\dfrac{k^2+3k}{k+2} < k+1$, and by taking the total expectation with respect to the randomness from the mini-batch and the  participated clients, i.e., $\mathbb{E}(\cdot) = \mathbb{E}_{i \in [m]}\mathbb{E}_{\xi_{i,k}}(\cdot)$. Thus we complete the proof of Theorem 1.

\subsection{Proof of Corollary 1}\label{proof of Corollary 1}
From the definition of $\Delta_i(x_{t-\tau,K})$, we have: 
\begin{equation}
\begin{aligned}
    \Delta_i(x_{t-\tau,K}) &= x_{t-\tau,K} - x_{t-\tau} 
    \\
    &= -\sum_{k=1}^{K}\eta_{i,k}\cdot g(x_{t-\tau,k-1}).
\end{aligned}
\end{equation}
Similar to the analysis of Eq.(\ref{submitR_1}) based on Lemma 1 and Assumption 2, and taking conditional expectation with respect to the randomness of mini-batch, we have:
\begin{subequations}\label{Deltatau}
\begin{align}
    &\quad\:\: \mathbb{E}_{\xi_{i,k}}\left[\left\| \Delta_i(x_{t-\tau,K}) \right\|^2\right] 
    \\
    &= \mathbb{E}_{\xi_{i,k}}\left[\left\|\sum_{k=1}^{K}\eta_{i,k} g(x_{t-\tau,k-1})\right\|^2\right]
    \\
    &\leq \mathbb{E}_{\xi_{i,k}}\left[ K \sum_{k=1}^{K}\eta_{i,k}^2\left\| g(x_{t-\tau,k-1})\right\|^2\right]  \label{25c}
    \\
    &= \mathbb{E}_{\xi_{i,k}}\left[ K \sum_{k=1}^{K}\eta_{i,k}^2\left\| g(x_{t-\tau,k-1}) - \nabla f(x_{t-\tau,k-1}) + \nabla f(x_{t-\tau,k-1})\right\|^2 \right]
    \\
    &= \mathbb{E}_{\xi_{i,k}}\left[ K \sum_{k=1}^{K}\eta_{i,k}^2 \left[ \left\| g(x_{t-\tau,k-1}) - \nabla f(x_{t-\tau,k-1}) \right\|^2+\left\|  \nabla f(x_{t-\tau,k-1})\right\|^2 \right]\right]
    \\
    &\leq K^2M\sigma_l^2 + \mathbb{E}_{\xi_{i,k}}\left[\underbrace{K \sum_{k=1}^{K}\eta_{i,k}^2 \left\|  \nabla f(x_{t-\tau,k-1}) \right\|^2}_{T_{1}} \right]   \label{25f}
\end{align}
\end{subequations}
where $T_{1}$ can be bounded by:
\begin{subequations}\label{T1!}
    \begin{align}
    \mathbb{E}_{\xi_{i,k}}\left[T_{1}\right] &=\mathbb{E}_{\xi_{i,k}}\left[ K \sum_{k=1}^{K}\eta_{i,k}^2  \left\| \nabla f(x_{t-\tau,k-1}) \right\|^2\right]
    \\
    &= \mathbb{E}_{\xi_{i,k}} \left[K\sum_{k=1}^{K}\eta_{i,k}^2  \left\|  \nabla f(x_{t-\tau,k-1}) - \nabla f(x_{t-\tau}) \right.\right. \notag
    \\
    &\qquad \left.\left. + \nabla f(x_{t-\tau}) - \nabla F(x_{t-\tau})+\nabla F(x_{t-\tau}) - \nabla F(x_{t}) + \nabla F(x_{t})\right\|^2\right]
    \\
    &\leq \mathbb{E}_{\xi_{i,k}}\left[4K\sum_{k=1}^{K}\eta_{i,k}^2  \left\|  \nabla f(x_{t-\tau,k-1}) - \nabla f(x_{t-\tau}) \right\|^2 \right]   \notag
    \\
    &\qquad + \mathbb{E}_{\xi_{i,k}}\left[4K\sum_{k=1}^{K}\eta_{i,k}^2  \left\|  \nabla f(x_{t-\tau}) - \nabla F(x_{t-\tau})\right\|^2\right] \notag
    \\
    &\qquad +\mathbb{E}_{\xi_{i,k}}\left[4K\sum_{k=1}^{K}\eta_{i,k}^2  \left\| \nabla F(x_{t-\tau}) - \nabla F(x_{t})\right\|^2 \right]   \notag
     \\
    &\qquad +\mathbb{E}_{\xi_{i,k}}\left[4K\sum_{k=1}^{K}\eta_{i,k}^2  \left\| \nabla F(x_{t})\right\|^2\right]       \label{26c}
    \\
    &\leq \mathbb{E}_{\xi_{i,k}}\left[4KL^2 \underbrace{\sum_{k=1}^{K}\eta_{i,k}^2 \left\| x_{t-\tau,k-1} -x_{t-\tau} \right\|^2 }_{T_{11}}\right] \notag
    \\
    &\qquad + 4K^2M\sigma_g^2 + 4K^2L^2M\Gamma^2 \mathbb{E}_{\xi_{i,k}}\left[\left\| \Delta_i(x_{t-\tau,K}) \right\|^2 \right]+ 4K^2M\mathbb{E}_{\xi_{i,k}}\left[\left\|\nabla F(x_t) \right\|^2\right],  \label{26d}
    \end{align}
\end{subequations}
where (\ref{25c}) and (\ref{26c}) both result from the Lemma 1, (\ref{25f}) is due to Assumption 2, and (\ref{26d}) is derived from Assumption 1,3 and 4.

With Theorem 1, the expectation of $T_{11}$ is given by
\begin{subequations}\label{T_11}
    \begin{align}
    &\quad\:\: \mathbb{E}_{\xi_{i,k}}\left[T_{11}\right] 
    \\
    &=  \sum_{k=1}^{K}\eta_{i,k}^2 \mathbb{E}_{\xi_{i,k}}\left[ \left\| x_{t-\tau,k-1} -x_{t-\tau} \right\|^2\right]
    \\
    &\leq \sum_{k=1}^{K}\frac{1}{6(2k+1)kL^2}\cdot\frac{k-1}{2L^2}\left(\sigma_g^2+\mathbb{E}_{\xi_{i,k}}\left[\left\|\nabla F(x_{t-\tau}) \right\|^2\right]\right)
     +\sum_{k=1}^{K}\frac{1}{6(2k+1)kL^2}\cdot\frac{k}{24L^2}\sigma_l^2
    \\
    &= \frac{1}{12L^4}\left( \sigma_g^2+\mathbb{E}_{\xi_{i,k}}\left[\left\|\nabla F(x_{t-\tau}) \right\|^2\right] \right) \cdot \sum_{k=1}^{K}\frac{k-1}{(2k+1)k} + \frac{1}{144L^4}\sigma_l^2 \cdot \sum_{k=1}^{K}\frac{k}{(2k+1)k}
    \\
    &\leq \frac{\ln{(K+1)}}{24L^4}\left( \sigma_g^2+\mathbb{E}_{\xi_{i,k}}\left[\left\|\nabla F(x_{t-\tau}) \right\|^2\right] \right) + \frac{\ln{(K+1)}}{288L^4}\sigma_l^2   \label{27e}
    \\
    &\leq \frac{\ln{(K+1)}}{12L^4} \mathbb{E}_{\xi_{i,k}}\left[\left\|\nabla F(x_{t-\tau}) \right\|^2\right] + \frac{\Gamma^2\ln{(K+1)}}{12L^2} \mathbb{E}_{\xi_{i,k}}\left[\left\| \Delta_i(x_{t-\tau,K}) \right\|^2\right] \notag
    \\
    &\qquad + \frac{\ln{(K+1)}}{24L^4}\sigma_g^2 + \frac{\ln{(K+1)}}{288L^4}\sigma_l^2, \label{27f}
    \end{align}
\end{subequations}
where (\ref{27e}) results from $\sum_{k=1}^{K}\frac{k-1}{(2k+1)k} < \sum_{k=1}^{K}\frac{k}{(2k+1)k} \leq \frac{\ln{(K+1)}}{2}$, and (\ref{27f}) results from Lemma 3. It then follows:
\begin{equation}
    \begin{aligned}
    \mathbb{E}_{\xi_{i,k}}\left[T_{1}\right] &\leq \left( \frac{K\ln{(K+1)}}{3L^2} + 4K^2M \right)\cdot\mathbb{E}_{\xi_{i,k}}\left[\left\|\nabla F(x_t) \right\|^2  \right]
    \\
    &\qquad   + \left(\frac{\Gamma^2K\ln{(K+1)}}{3}+4K^2L^2M\Gamma^2 \right) \cdot\mathbb{E}_{\xi_{i,k}}\left[\left\| \Delta_i(x_{t-\tau,K}) \right\|^2\right]
    \\
    &\qquad + \left(\frac{K\ln{(K+1)}}{72L^2}\right)\sigma_l^2 + \left(\frac{K\ln{(K+1)}}{6L^2}+4K^2M \right)\sigma_g^2.
    \end{aligned}
\end{equation}

With this upper bound of $T_1$,  we can derive from Eq.(\ref{Deltatau}):
\begin{equation}
    \begin{aligned}
    &\left( 1 - \left(\frac{\Gamma^2K\ln{(K+1)}}{3}+4K^2L^2M\Gamma^2 \right)  \right) \cdot\mathbb{E}_{\xi_{i,k}}\left[\left\| \Delta_i(x_{t-\tau,K}) \right\|^2\right]
    \\
    \leq & \frac{12K^2L^2M+K\ln{(K+1)}}{3L^2 } \cdot\mathbb{E}_{\xi_{i,k}}\left[\left\|\nabla F(x_t) \right\|^2 \right] 
    + \frac{72K^2L^2M + K\ln{(K+1)}}{72L^2}\sigma_l^2 
    \\
    &\qquad + \frac{24K^2L^2M+K\ln{(K+1)}}{6L^2} \sigma_g^2.
    \end{aligned}
\end{equation}

Define $ \Theta(\Gamma):= 1 - \left(\frac{\Gamma^2K\ln{(K+1)}}{3}+4K^2L^2M\Gamma^2 \right) $. $\Theta(\Gamma)>0$ leads to $\Gamma^2 < \frac{3}{12K^2L^2M+K\ln{(K+1)}}$, after sorting the terms and taking the total expectation (with respect to the randomness from the mini-batch and the clients randomly participated in, i.e., $\mathbb{E} = \mathbb{E}_{i \in [m]}\mathbb{E}_{\xi_{i,k}}$) on both sides, we complete the proof of Corollary 1.

\subsection{Proof of Theorem 2}\label{proof of Theorem 2}


Assume that at the global model iteration $t$, client $i, i\in [m]$ upload its local update result $\Delta_i(x_{t-\tau,K})$to the server. And the update at the server can be denoted by:
\begin{equation}
    x_{t+1} = x_t + \eta_g \cdot \Delta_i(x_{t-\tau,K}). 
\end{equation}
From Lemma 2 and the triangle inequality, we have:
\begin{equation}\label{lipchitz}
    \begin{aligned}
    F\left(x_{t+1}\right)-F\left(x_{t}\right) & \leq\left\langle\nabla F\left(x_{t}\right), x_{t+1}-x_{t}\right\rangle+\frac{L}{2}\left\|x_{t+1}-x_{t}\right\|^{2} 
    \\
    & = \left\langle\nabla F\left(x_{t}\right), \eta_g \Delta_i(x_{t-\tau,K})  \right\rangle+\frac{L\eta_g^2}{2}\left\| \Delta_i(x_{t-\tau,K}) \right\|^{2}
    \\
    & = -\frac{K\eta_g}{N}\langle N\nabla F(x_{t}), -\frac{1}{K}\Delta_i(x_{t-\tau,K}) \rangle +\frac{L\eta_g^2}{2}\left\| \Delta_i(x_{t-\tau,K}) \right\|^{2}. 
    \end{aligned}
\end{equation}

From the fact that $2\langle a,b\rangle = \left\|a \right\|^2+\left\|b \right\|^2-\left\|a-b \right\|^2$, and $\Delta_i(x_{t-\tau,K}) = -\sum_{k=1}^{K}\eta_{i,k}g(x_{t-\tau,k-1})$, take  conditional expectation with respect to the randomness of mini-batch on both sides of Eq.(\ref{lipchitz}), it yields:
\begin{subequations}
    \begin{align}
    &\quad\:\: \mathbb{E}_{\xi_{i,k}}\left[ F\left(x_{t+1}\right)-F\left(x_{t}\right)\right] 
    \\
    & \leq -\frac{KN\eta_g}{2}\cdot\mathbb{E}_{\xi_{i,k}}\left[\left\| \nabla F(x_{t}) \right\|^2 \right] + \left(\frac{L\eta_g^2}{2}-\frac{\eta_g}{2NK}\right)\cdot\mathbb{E}_{\xi_{i,k}}\left[\left\|\Delta_i(x_{t-\tau,K}) \right\|^{2}\right] \notag
    \\
    &\qquad + \frac{K\eta_g}{2N} \cdot\mathbb{E}_{\xi_{i,k}}\left[\left\|N \nabla F(x_{t}) + \frac{1}{K} \Delta_i(x_{t-\tau})\right\|^2 \right] 
    \\
    & = -\frac{KN\eta_g}{2}\cdot\mathbb{E}_{\xi_{i,k}}\left[\left\| \nabla F(x_{t}) \right\|^2 \right]
     + \left(\frac{L\eta_g^2}{2}-\frac{\eta_g}{2NK}\right)\cdot\mathbb{E}_{\xi_{i,k}}\left[\left\|\Delta_i(x_{t-\tau,K}) \right\|^{2}\right] \notag
    \\
    &\qquad + \frac{K\eta_g}{2N} \cdot\mathbb{E}_{\xi_{i,k}}\left[\underbrace{\left\|N \nabla F(x_{t}) - \frac{1}{K} \sum_{k=1}^{K}\eta_{i,k} g(x_{t-\tau,k-1})\right\|^2}_{T_2} \right].
    \end{align} \label{All}
\end{subequations}

Similar to the analysis of Eq.(\ref{submitR_1}) based on Lemma 1 and Assumption 2, $T_2$ can be bounded by:
\begin{subequations}
    \begin{align}
    \mathbb{E}_{\xi_{i,k}}\left[T_2\right] &= \mathbb{E}_{\xi_{i,k}}\left[\left\|N \nabla F(x_{t}) - \frac{1}{K} \sum_{k=1}^{K}\eta_{i,k} g(x_{t-\tau,k-1})\right\|^2 \right]
    \\
    &= \mathbb{E}_{\xi_{i,k}}\left[\left\|N \nabla F(x_{t}) - \frac{1}{K} \sum_{k=1}^{K}\eta_{i,k} \nabla f(x_{t-\tau,k-1}) \right.\right. \notag
    \\
    &\qquad \left.\left.+ \frac{1}{K} \sum_{k=1}^{K}\eta_{i,k}\nabla f(x_{t-\tau,k-1}) -\frac{1}{K} \sum_{k=1}^{K}\eta_{i,k} g(x_{t-\tau,k-1})\right\|^2 \right]
    \\
    &= \mathbb{E}_{\xi_{i,k}}\left[\left\|N \nabla F(x_{t}) - \frac{1}{K} \sum_{k=1}^{K}\eta_{i,k} \nabla f(x_{t-\tau,k-1})\right\|^2\right]   \notag
    \\
    &\qquad +\mathbb{E}_{\xi_{i,k}}\left[ \left\|\frac{1}{K} \sum_{k=1}^{K}\eta_{i,k} \nabla f(x_{t-\tau,k-1}) -\frac{1}{K} \sum_{k=1}^{K}\eta_{i,k} g(x_{t-\tau,k-1})\right\|^2 \right]
    \\
    &\leq \mathbb{E}_{\xi_{i,k}}\left[\underbrace{\left\|N \nabla F(x_{t}) - \frac{1}{K} \sum_{k=1}^{K}\eta_{i,k} \nabla f(x_{t-\tau,k-1})\right\|^2 }_{T_{21}}\right] +M\sigma_l^2.
    \end{align}
\end{subequations}
For term $T_{21}$, applying Lemma 1, we have:
\begin{subequations}\label{T21_}
    \begin{align}
    &\quad\:\:\mathbb{E}_{\xi_{i,k}}\left[T_{21}\right] 
    \\
    &= \mathbb{E}_{\xi_{i,k}}\left[\left\|N \nabla F(x_{t}) - \frac{1}{K}\sum_{k=1}^{K}\eta_{i,k} \nabla f(x_{t-\tau,k-1})\right\|^2 \right]
    \\
    &= \mathbb{E}_{\xi_{i,k}}\left[\left\| N\nabla F(x_{t}) - N \nabla F(x_{t-\tau})  + N \nabla F(x_{t-\tau}) - N \nabla f(x_{t-\tau})\right.\right.        \notag
    \\
    &\qquad \left.\left. + \frac{1}{K}\sum_{k=1}^{K}\eta_{i,k}\nabla f(x_{t-\tau}) - \frac{1}{K}\sum_{k=1}^{K}\eta_{i,k} \nabla f(x_{t-\tau,k-1})\right\|^2 \right]
    \\
    &\leq 3\mathbb{E}_{\xi_{i,k}}\left[ \left\| N\nabla F(x_{t}) - N\nabla F(x_{t-\tau}) \right\|^2 \right]
    + 3\mathbb{E}_{\xi_{i,k}}\left[ \left\|  N\nabla F(x_{t-\tau}) - N\nabla f(x_{t-\tau})\right\|^2 \right]     \notag 
    \\
    &\qquad + 3\mathbb{E}_{\xi_{i,k}}\left[\left\|  \frac{1}{K}\sum_{k=1}^{K}\eta_{i,k}\nabla f(x_{t-\tau}) - \frac{1}{K}\sum_{k=1}^{K}\eta_{i,k} \nabla f(x_{t-\tau,k-1})\right\|^2\right]\label{35d}
    \\
    &\leq 3N^2L^2\Gamma^2 \cdot\mathbb{E}_{\xi_{i,k}}\left[\left\| \Delta_i(x_{t-\tau,K}) \right\|^{2}\right] +3N^2\sigma_g^2 
    + \frac{3L^2}{K}\cdot\sum_{k=1}^{K}\eta_{i,k}^2\cdot\mathbb{E}_{\xi_{i,k}}\left[\left\|x_{t-\tau,k-1} - x_{t-\tau} \right\|^2 \right]     \label{34e}
    \\
    &\leq \frac{\ln{(K+1)}}{4KL^2}\cdot\mathbb{E}_{\xi_{i,k}}\left[ \left\|\nabla F(x_{t})\right\|^2 \right]+ \left( \frac{\Gamma^2\ln{(K+1)}}{4KL^2}+3N^2L^2\Gamma^2 \right) \cdot\mathbb{E}_{\xi_{i,k}}\left[\left\| \Delta_i(x_{t-\tau,K}) \right\|^{2}\right] \notag
    \\
    &\qquad + \left(\frac{\ln{(K+1)}}{8KL^2} + 3N^2 \right)\sigma_g^2 + \frac{\ln{(K+1)}}{96KL^2}\sigma_l^2     \label{34f}
    \end{align}
\end{subequations}
In (\ref{35d}), the first term can be bounded by Assumption 1 and Assumption 4, the second term can be bounded by Assumption 3, the third term can be bounded by the upper bound of term $T_{11}$, and (\ref{34e}),(\ref{34f}) showed their upper bounds. Submit Eq.(\ref{Deltatau}) to term $T_2$ and sort terms on the right side, we can get:
\begin{equation}
    \begin{aligned}
    \mathbb{E}_{\xi_{i,k}}\left[T_{2}\right] 
    &\leq \frac{\ln{(K+1)}}{4KL^2}\cdot\mathbb{E}_{\xi_{i,k}}\left[\left\|\nabla F(x_{t})\right\|^2\right] + \left(\frac{\ln{(K+1)}}{96KL^2} + M \right)\sigma_l^2  
    \\
    &\:\:\:\: + \left(\frac{\ln{(K+1)}}{8KL^2}+3N^2 \right)\sigma_g^2 + \left(\frac{\Gamma^2\ln{(K+1)}}{4KL^2}+3N^2L^2\Gamma^2\right)\mathbb{E}_{\xi_{i,k}}\left[\left\|\Delta_i(x_{t-\tau,K}) \right\|^{2}\right] \label{T2}.
    \end{aligned}
\end{equation}
If $\eta_{i,k}^2 \leq \dfrac{1}{6(2k+1)kL^2}$, from the Cauchy-Schwarz inequality, we can get $N^2\leq M\leq \dfrac{1}{K}\sum_{k=1}^{K=1} \frac{1}{6(2k+1)kL^2} \leq \dfrac{1}{18L^2}$. If $\eta_g \leq \frac{1}{NLK}$, taking the total expectation (with respect to the randomness from the mini-batch and the clients randomly participated in, i.e., $\mathbb{E} = \mathbb{E}_{i \in [m]}\mathbb{E}_{\xi_{i,k}}$) on both side, we can get:

\begin{equation}\label{finalF}
    \begin{aligned}
        \mathbb{E}\left[F\left(x_{t+1}\right)-F\left(x_{t}\right)\right] & \leq -\frac{KN\eta_g}{2}\mathbb{E}\left[\left\| \nabla F(x_{t}) \right\|^2\right] + \frac{L\eta_g^2}{2} \cdot \frac{K\ln{(K+1)}+4K^2}{72L^2\Theta(\Gamma) } \sigma_l^2 
        \\
        &\qquad + \frac{L\eta_g^2}{2} \cdot \frac{3K\ln{(K+1)}+4K^2}{18L^2\Theta(\Gamma) } \sigma_g^2.
    \end{aligned}
\end{equation}
For $t=1 \to T$, we can obtain:
\begin{equation}
    \begin{aligned}
    \frac{KN\eta_g}{2} \cdot\mathbb{E}\left[\sum_{t=1}^{T} \left\| \nabla F(x_t) \right\|^2\right] &\leq F\left(x_{1}\right)-\mathbb{E}\left[F\left(x_{T+1}\right)\right] + T\cdot\frac{L\eta_g^2}{2} \cdot \frac{K\ln{(K+1)}+4K^2}{72L^2\Theta(\Gamma) } \sigma_l^2 
    \\
    &\qquad + T\cdot \frac{L\eta_g^2}{2} \cdot \frac{3K\ln{(K+1)}+4K^2}{18L^2\Theta(\Gamma) } \sigma_g^2.
    \end{aligned}
\end{equation}
When $T\to\infty, \mathbb{E}\left[ F\left(x_{T+1}\right)\right] = F\left(x^{*}\right)$, after simplifying the terms on both side we can complete the proof of Theorem 2.

\section{Experimental Details}\label{experimentaldetails}

\subsection{Training tasks and models}\label{model structure}
\paragraph{Synthetic-1-1}
It is a synthetic data set built by \citet{li2019fair}. Local data samples in each client $i$ are generated by the rule of $ y = \arg \max ({\rm softmax}(W_ix_i+b_i))$. The difference of the $(W_i,b_i)$ distribution in each client is controlled by the  parameter of $\alpha$, and the difference of $x_i$ distribution in each client is controlled by the parameter of $\beta$. $(\alpha,\beta) = (0,0)$ means that the data distribution between clients is IID, and increase the value of $(\alpha,\beta)$ will strengthen the Non-IIDness.
We chose $(\alpha,\beta) = (1,1)$ for our experiments and set the total number of clients as 10, as well as the number of samples in client $i$ follows a power law. Finally, a multilayer perceptron containing three fully connected layers was constructed for this classification task.

\paragraph{FEMNIST}
The full name of FEMNIST is Federated-MNIST and it was built by \citet{leaf} for federated learning algorithm testing. FEMNIST contains additional 26 uppercase and 26 lowercase letters compared to the original MNIST dataset, which was designed for handwritten character recognition. It contains 805263 samples and is divided into 3500 sub-datasets based Non-IID distribution, which can be distributed to 3500 clients. We chose 10 of these sub-datasets randomly for our experiments, and the random seed we chose can be found in the code. We use a basic convolutional neural network for this task, which contains two convolution layers and a pooling layer, as well as one fully connected layer.

\paragraph{ShakeSpeare text data}
This data set is built from \textit{The Complete Works of William Shakespeare} by \citet{FedAvg}, where dialog contents of each role in these plays are respectively assigned to one client for the task of next character prediction. This data set contains 422615 samples and is divided into 1129 sub-datasets. The data distribution is naturally Non-IID and we randomly choose 10 of these sub-datasets. We use a recurrent neural network for this task, which contains one embedding layer, two LSTM layers, and one fully-connected layer.

\subsection{Training Environment}\label{training environment}
We use multiple cores of a single computer which contains 2 ${\rm Intel}^{\circledR}$ ${\rm Xeon}^{\circledR}$ Silver 4114T CPUs and 4 ${\rm NVidia}^{\circledR}$ GeForce RTX 2080 SUPER GPUs for simulations. 
We implemented our distributed learning framework in Pytorch\cite{paszke2019pytorch}. Each of the server and clients is an independent process, and occupies a CPU core, respectively, so that the procedures of them do not interfere with each other. In particular, the server process has the highest priority to ensure the global update. 

Each client communicates with the server via the TCP/IP protocol for data transmission. The transmitting time can be calculated by:
$$
transmitting\ time = \frac{the\ total\ size\ of\ model\ parameters}{transmission\ speed}\times coefficient
$$
where in the simulation the $transmission\ speed$ is a fixed constant and the $coefficient$ is a random number following the normal distribution.

Further, instead of choosing a part of all clients for global model aggregation, we chose all clients to perform local training and set a certain probability $P$ of each client being suspended to simulate time-varying property. The hang time is a random value distributed with respect to the maximum running time. $P=0$ means that all clients are connected with server steadily.  Clients becomes more volatile as the value of $P$ increases.

\subsection{All results with different $P$}\label{allresults}
We present the test accuracy v.s. running time curves with regards to different values of $P=0,0.1,0.2,\cdots,0.9$ in Figure 5-14. We can observe that AsyncFedED converges remarkably faster than all the baseline algorithm on all the three considered tasks for different values of $P$.

\begin{figure}[H]
\centering
\subfigure[Synthetic-1-1]{\includegraphics[width=0.32\textwidth]{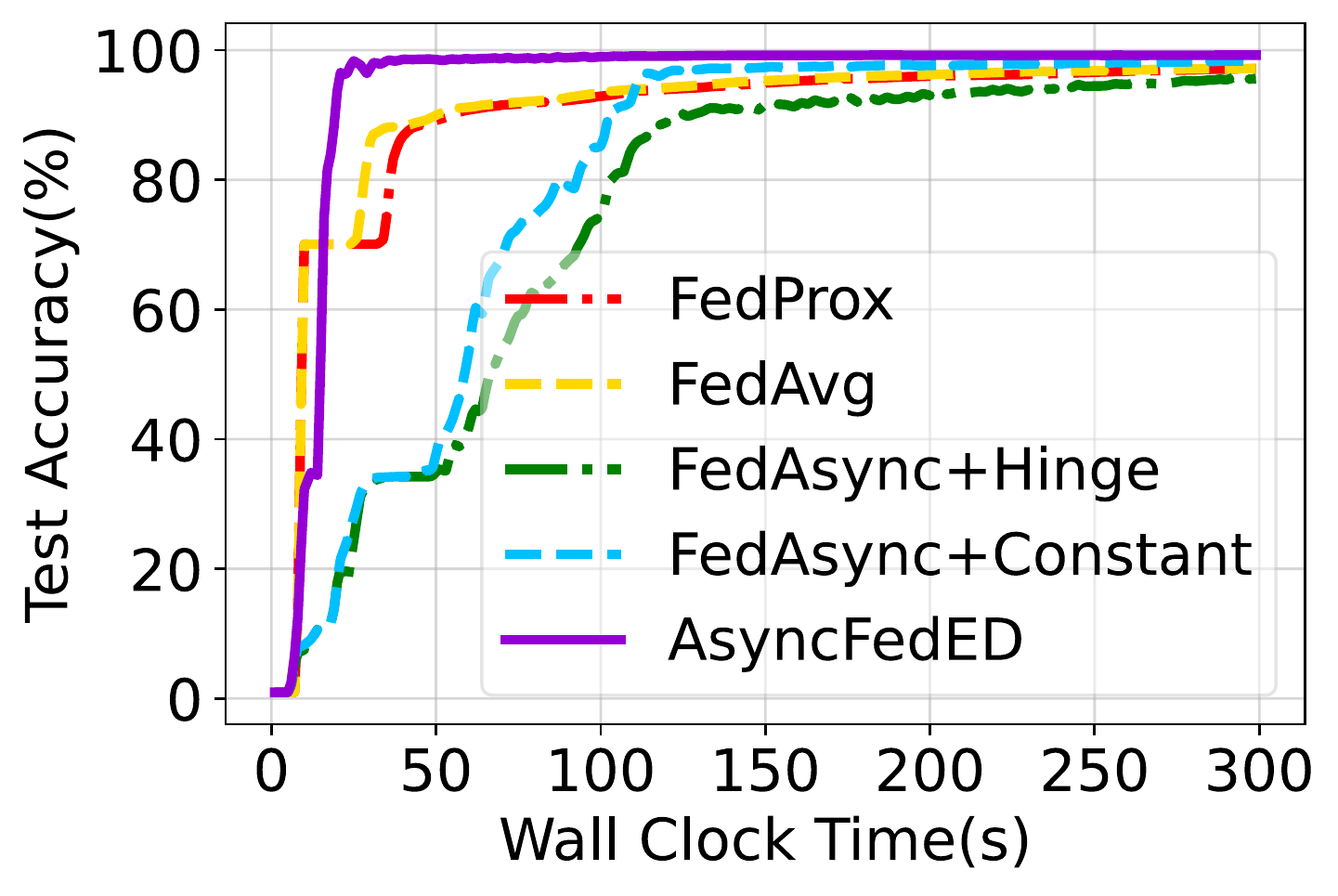}}
\subfigure[FEMNIST]{\includegraphics[width=0.32\textwidth]{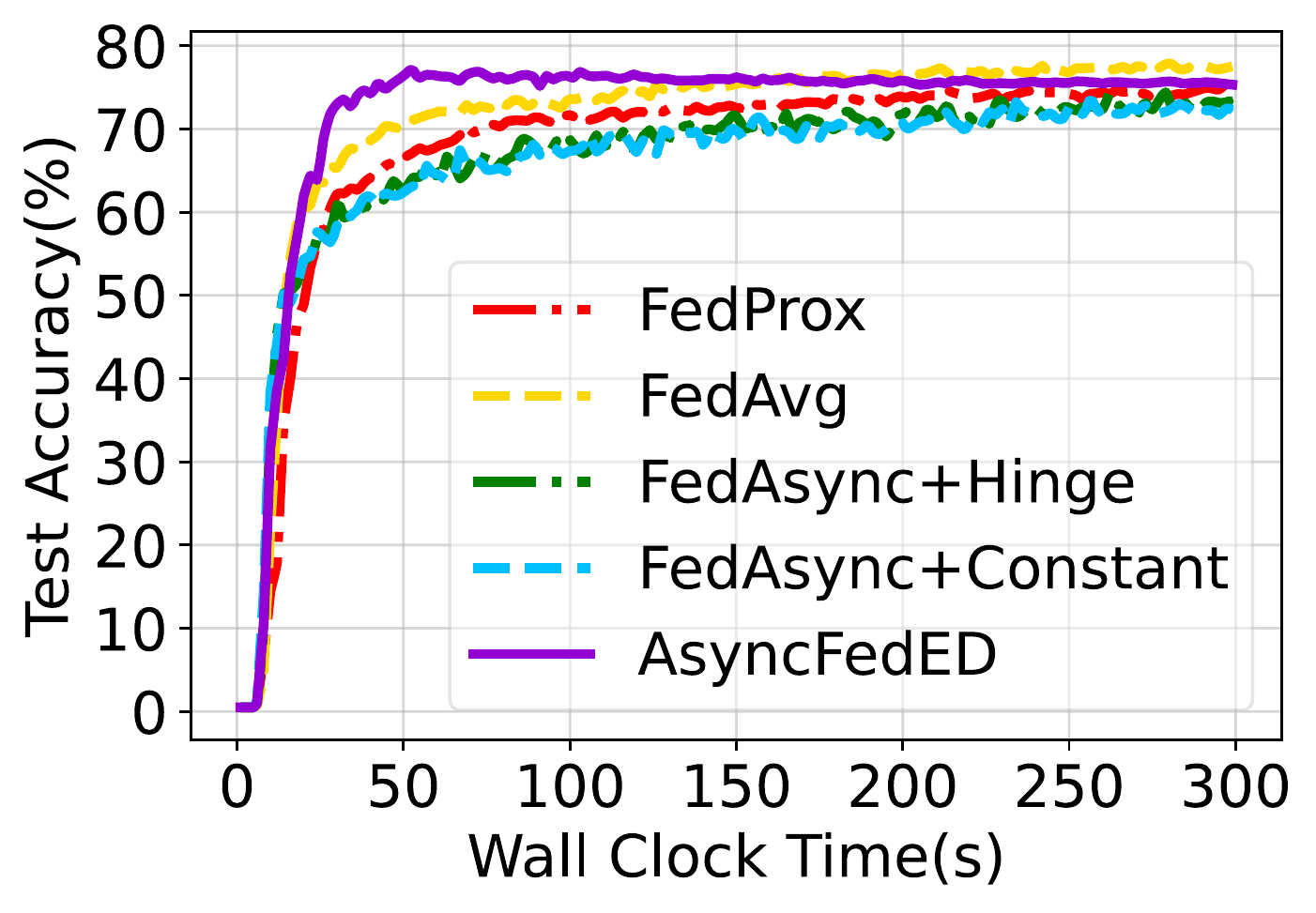}}  
\subfigure[ShakeSpeare text data]{\includegraphics[width=0.32\textwidth]{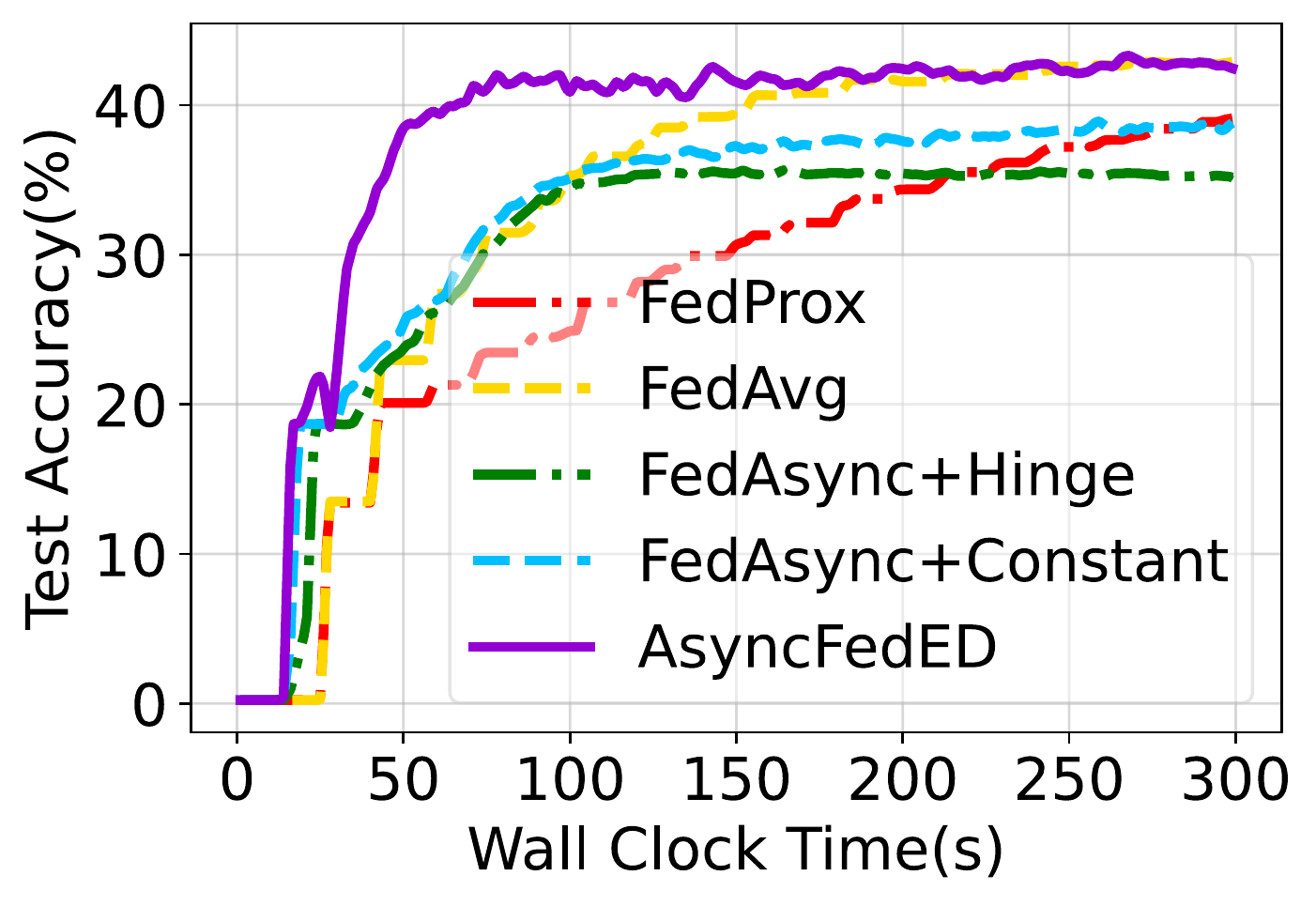}} 
\centering
\vspace{-0.3cm}
\caption{Test accuracy v.s. training time curves when the suspension probability is set to $P=0$. } 
\end{figure}
\vspace{-0.5cm}

\begin{figure}[H]
\centering
\subfigure[Synthetic-1-1]{\includegraphics[width=0.32\textwidth]{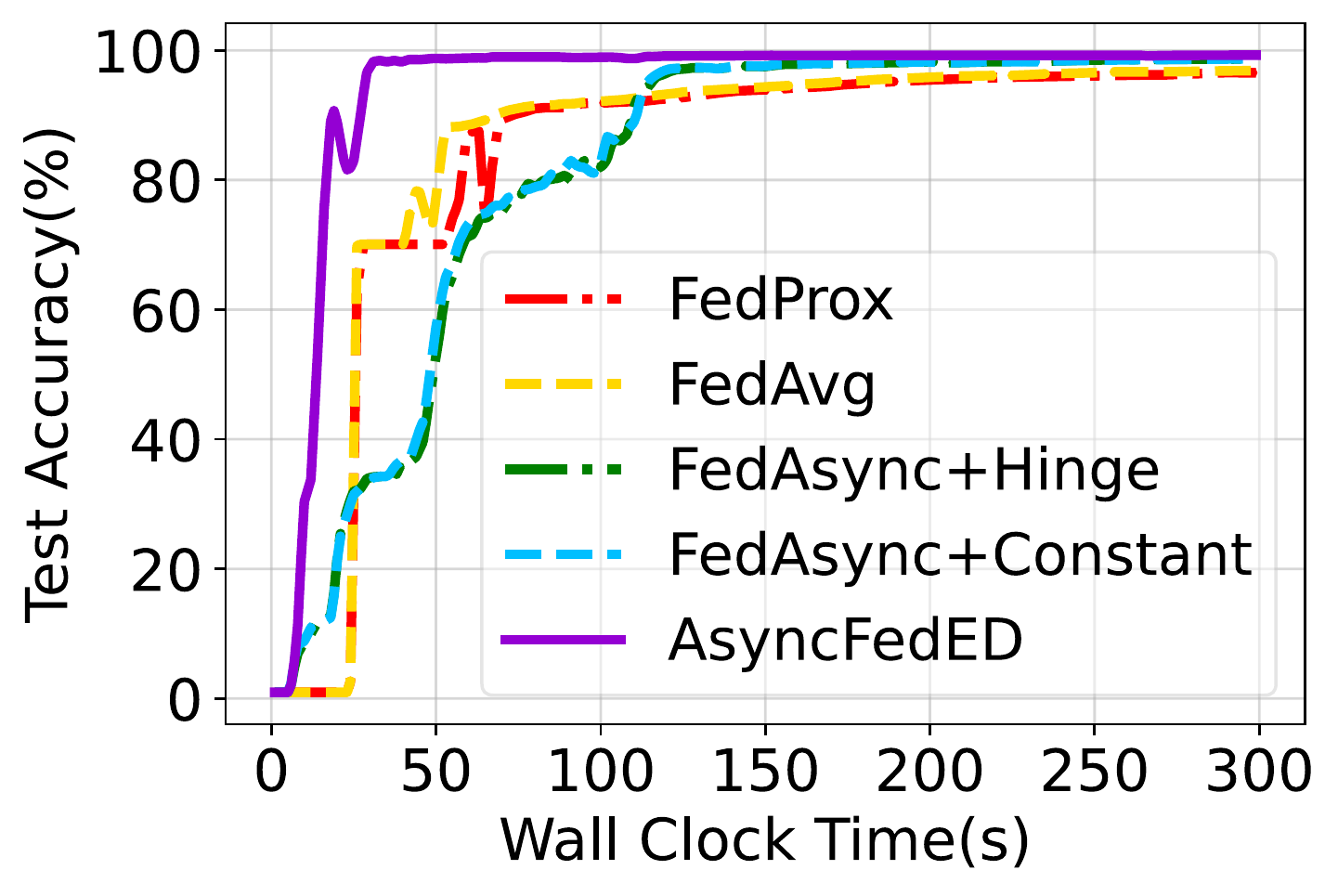}}
\subfigure[FEMNIST]{\includegraphics[width=0.32\textwidth]{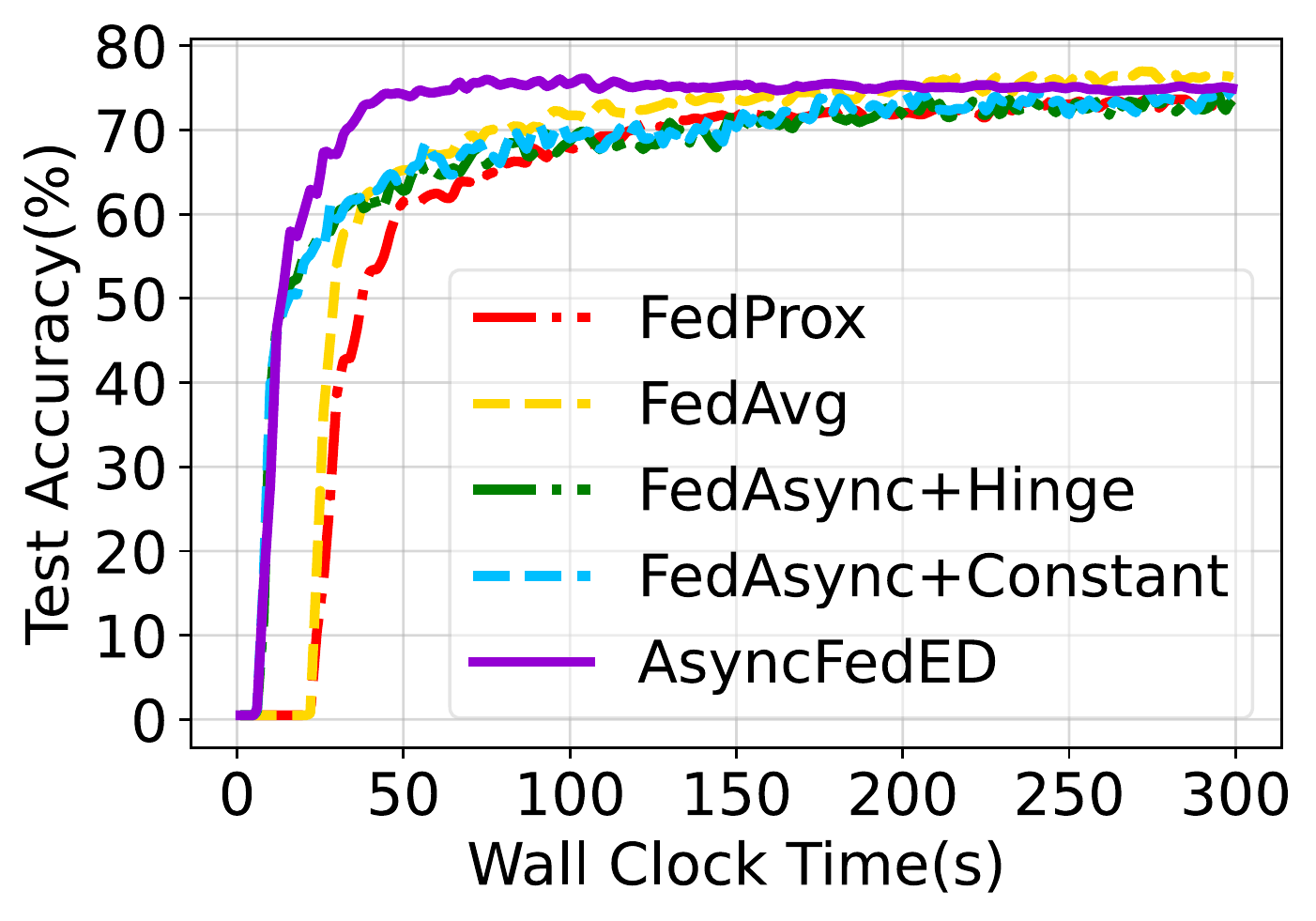}} 
\subfigure[ShakeSpeare text data]{\includegraphics[width=0.32\textwidth]{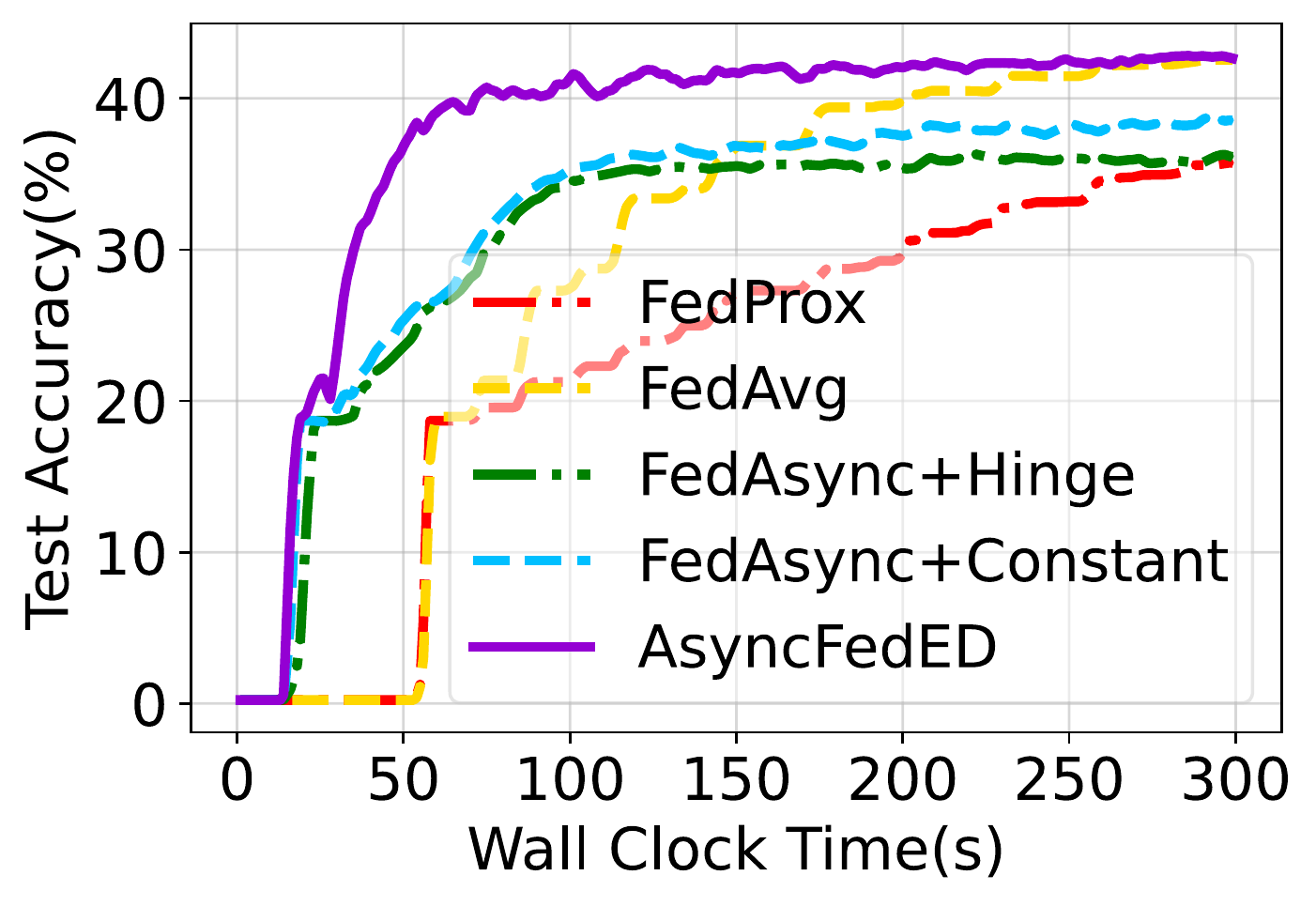}} 
\centering
\vspace{-0.3cm}
\caption{Test accuracy v.s. training time curves when the suspension probability is set to $P=0.1$. } 
\end{figure}
\vspace{-0.5cm}

\begin{figure}[H]
\centering
\subfigure[Synthetic-1-1]{\includegraphics[width=0.32\textwidth]{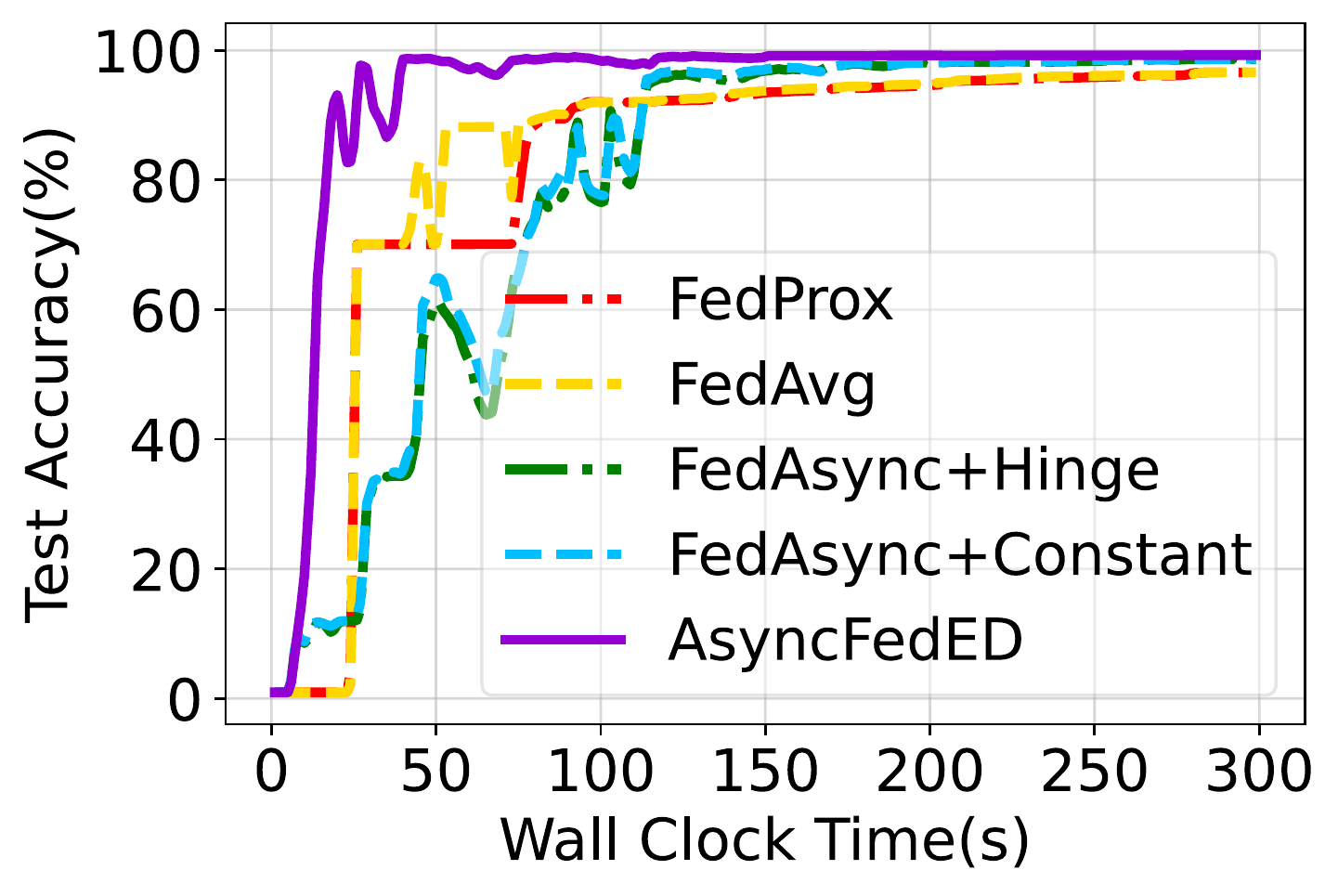}}
\subfigure[FEMNIST]{\includegraphics[width=0.32\textwidth]{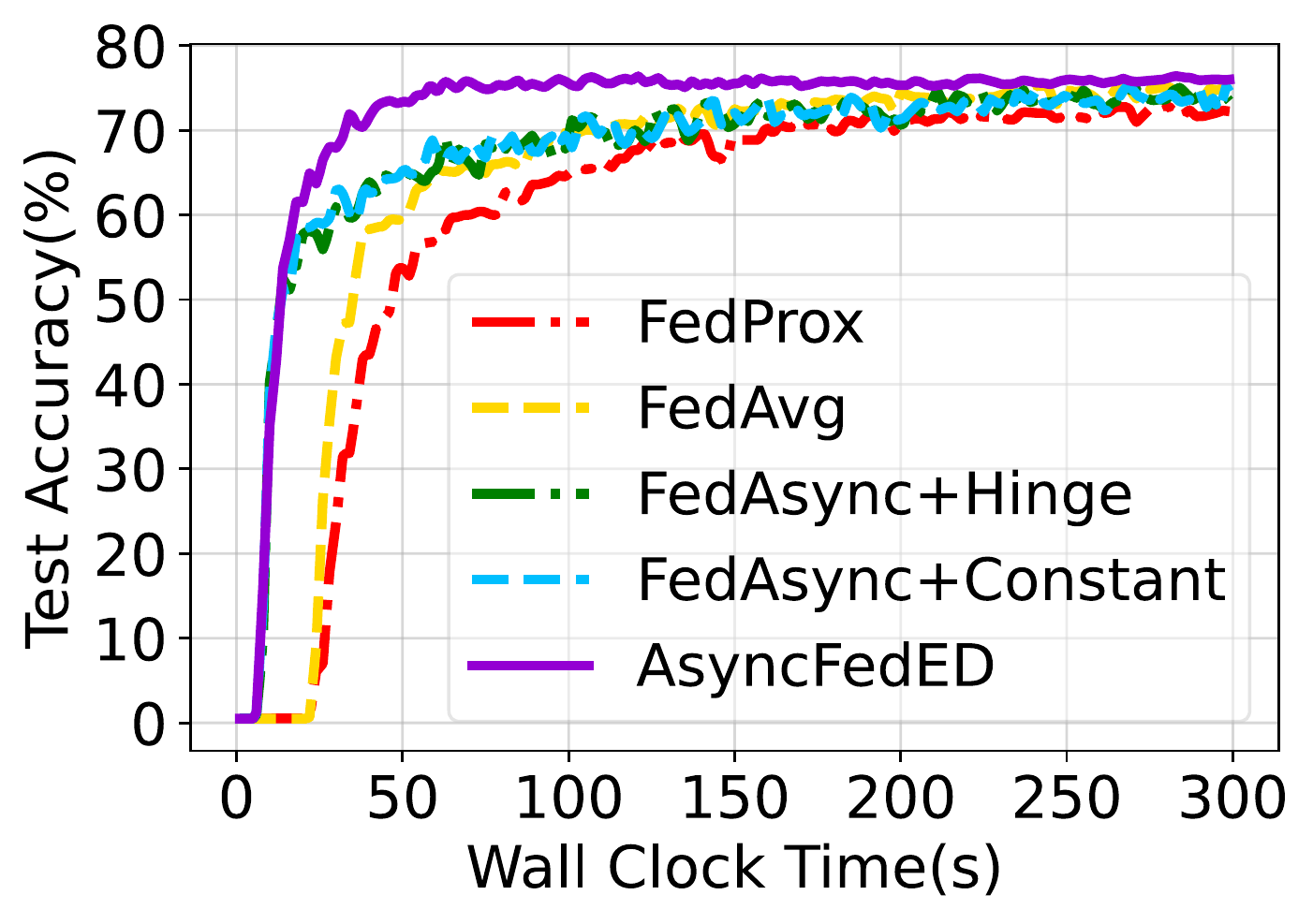}} 
\subfigure[ShakeSpeare text data]{\includegraphics[width=0.32\textwidth]{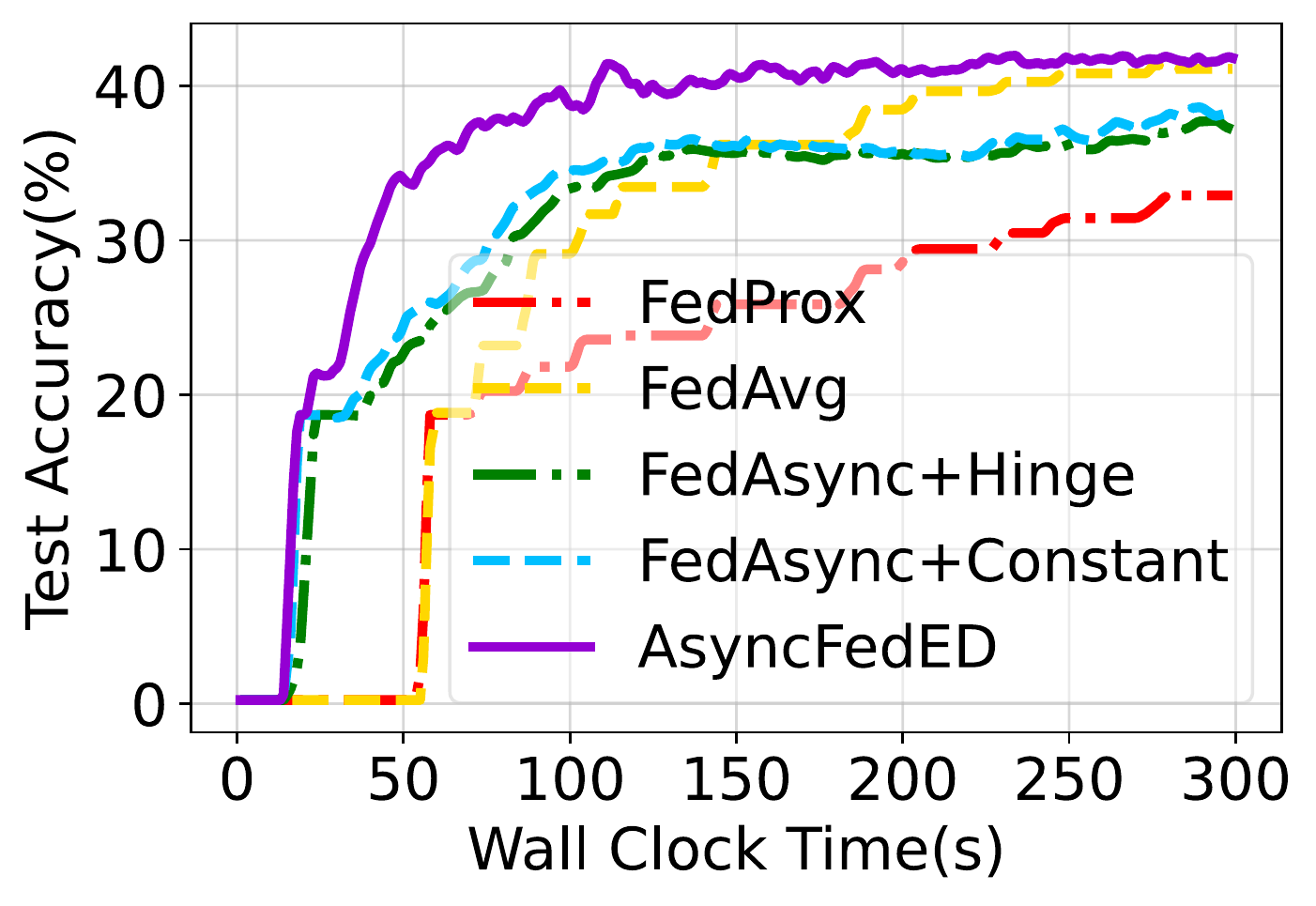}} 
\centering
\vspace{-0.3cm}
\caption{Test accuracy v.s. training time curves when the suspension probability is set to $P=0.2$. } 
\end{figure}
\vspace{-0.5cm}

\begin{figure}[H]
\centering
\subfigure[Synthetic-1-1]{\includegraphics[width=0.32\textwidth]{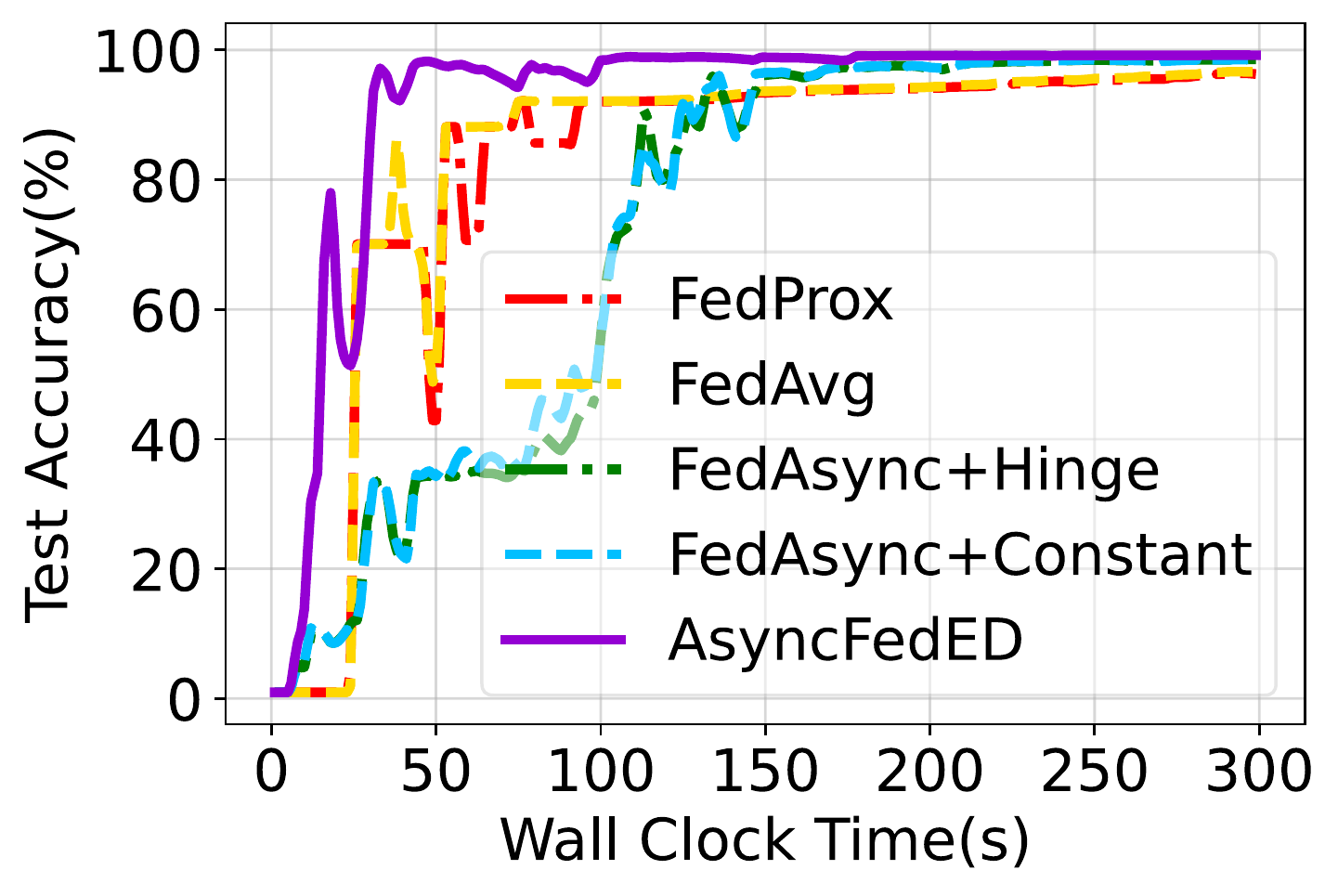}}
\subfigure[FEMNIST]{\includegraphics[width=0.32\textwidth]{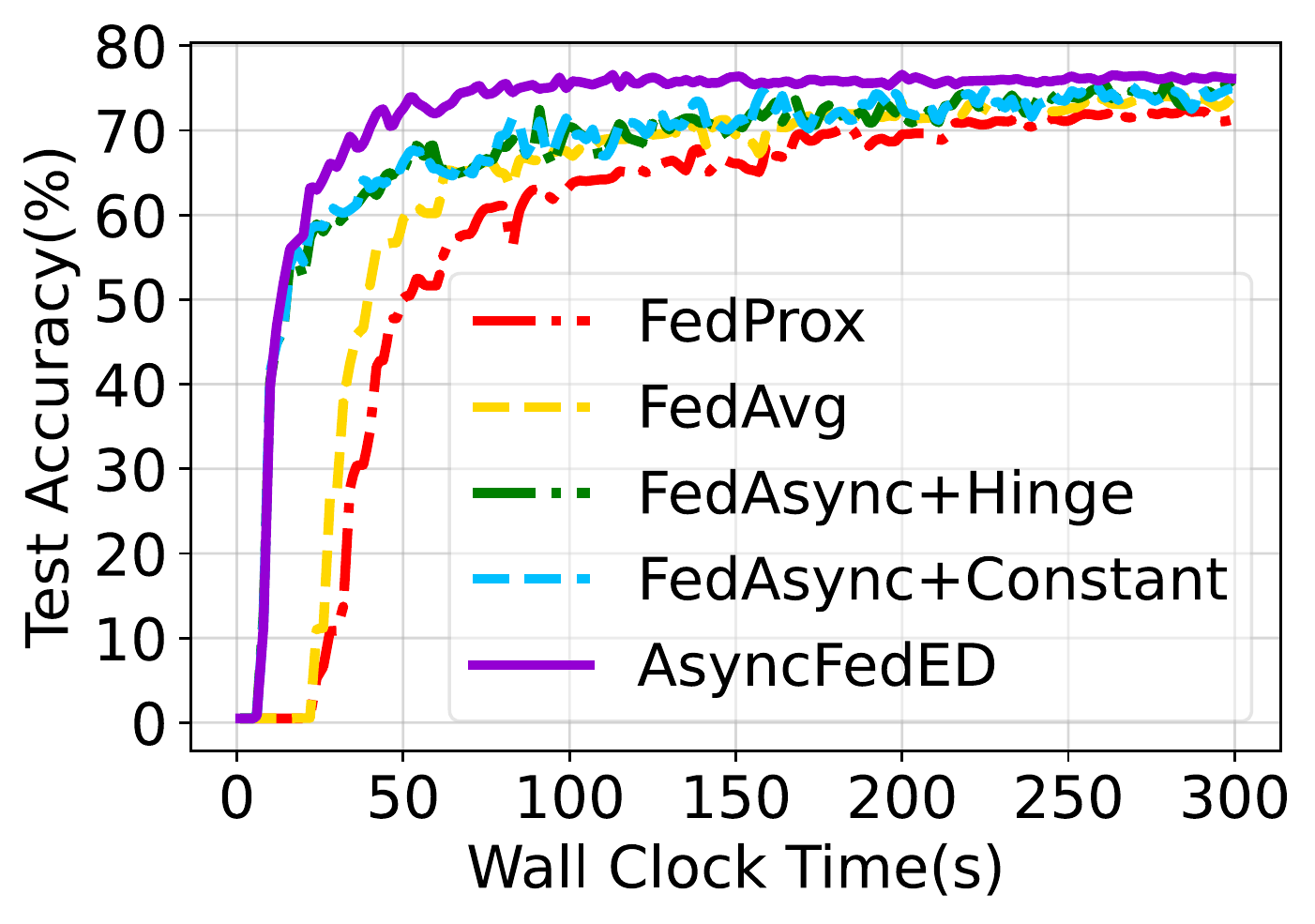}} 
\subfigure[ShakeSpeare text data]{\includegraphics[width=0.32\textwidth]{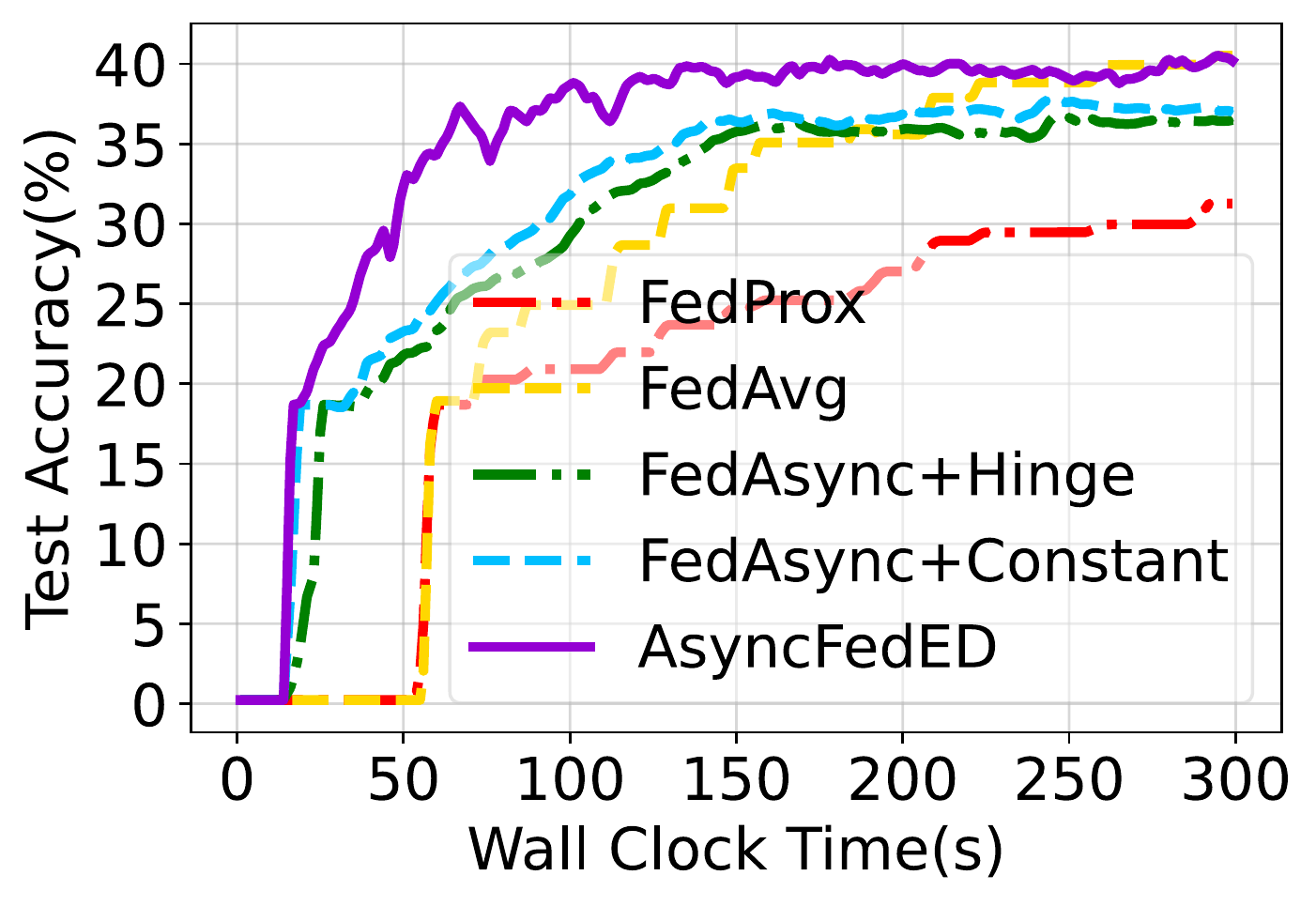}} 
\centering
\vspace{-0.3cm}
\caption{Test accuracy v.s. training time curves when the suspension probability is set to $P=0.3$. } 
\end{figure}
\vspace{-0.5cm}

\begin{figure}[H]
\centering
\subfigure[Synthetic-1-1]{\includegraphics[width=0.32\textwidth]{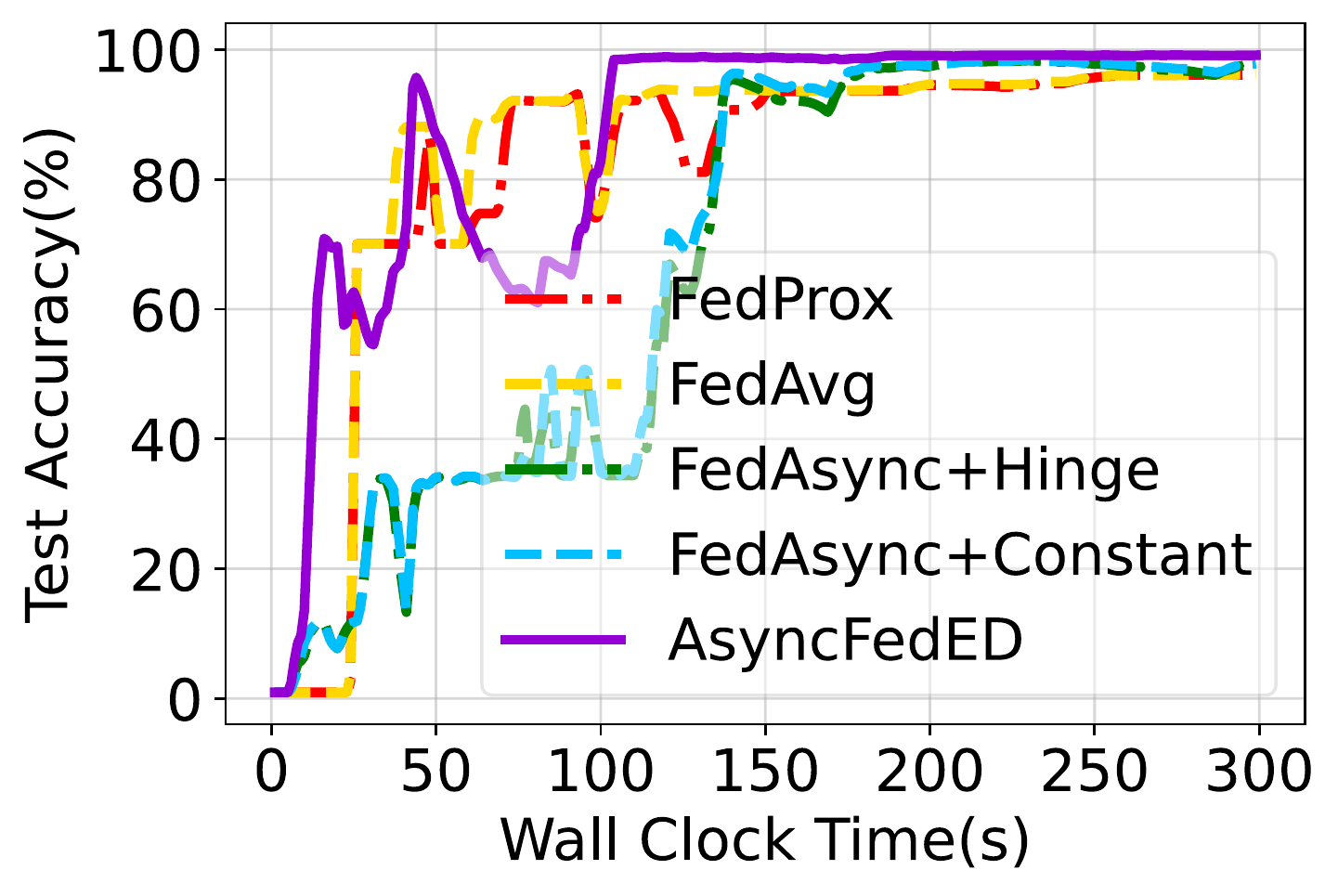}}
\subfigure[FEMNIST]{\includegraphics[width=0.32\textwidth]{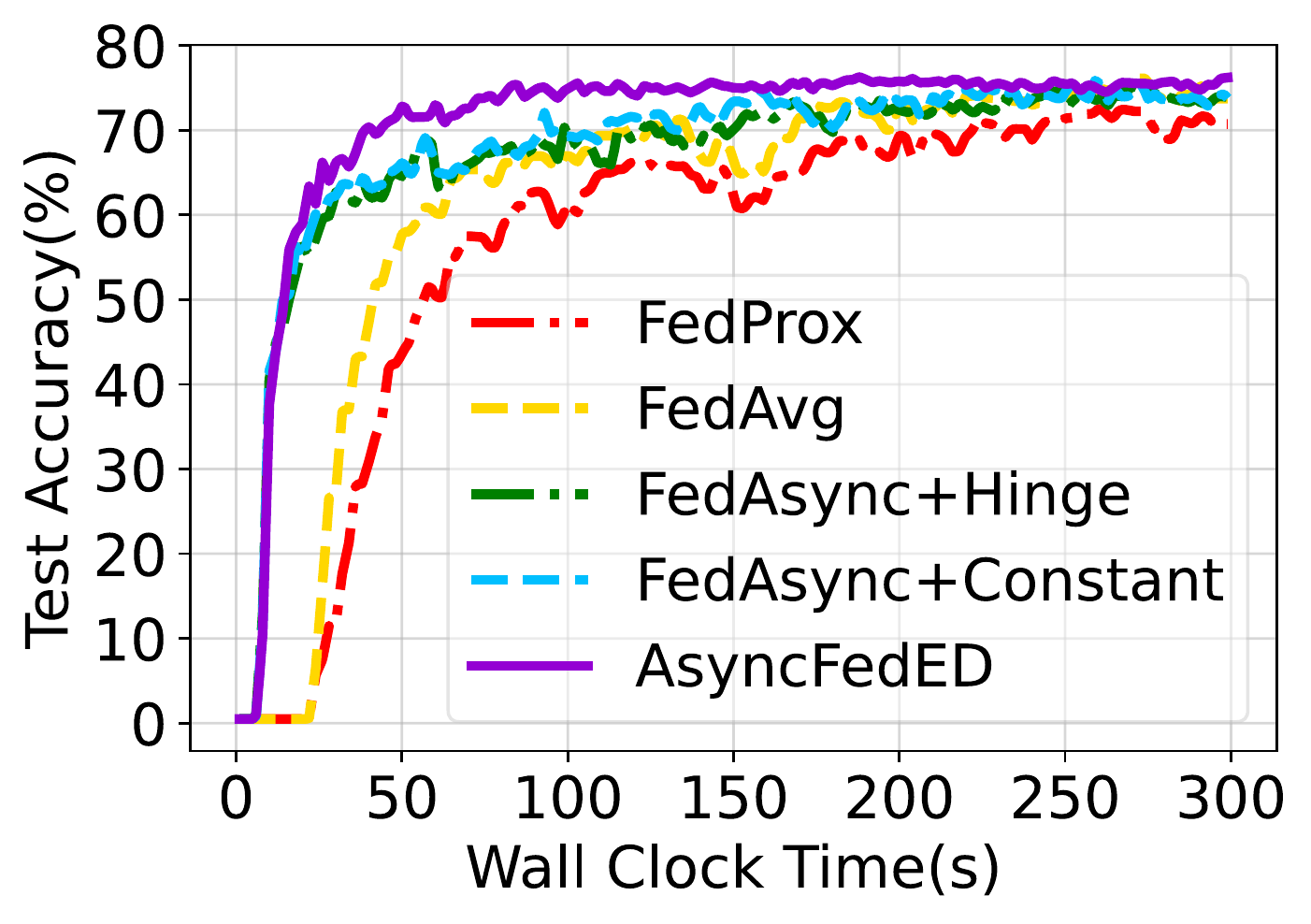}} 
\subfigure[ShakeSpeare text data]{\includegraphics[width=0.32\textwidth]{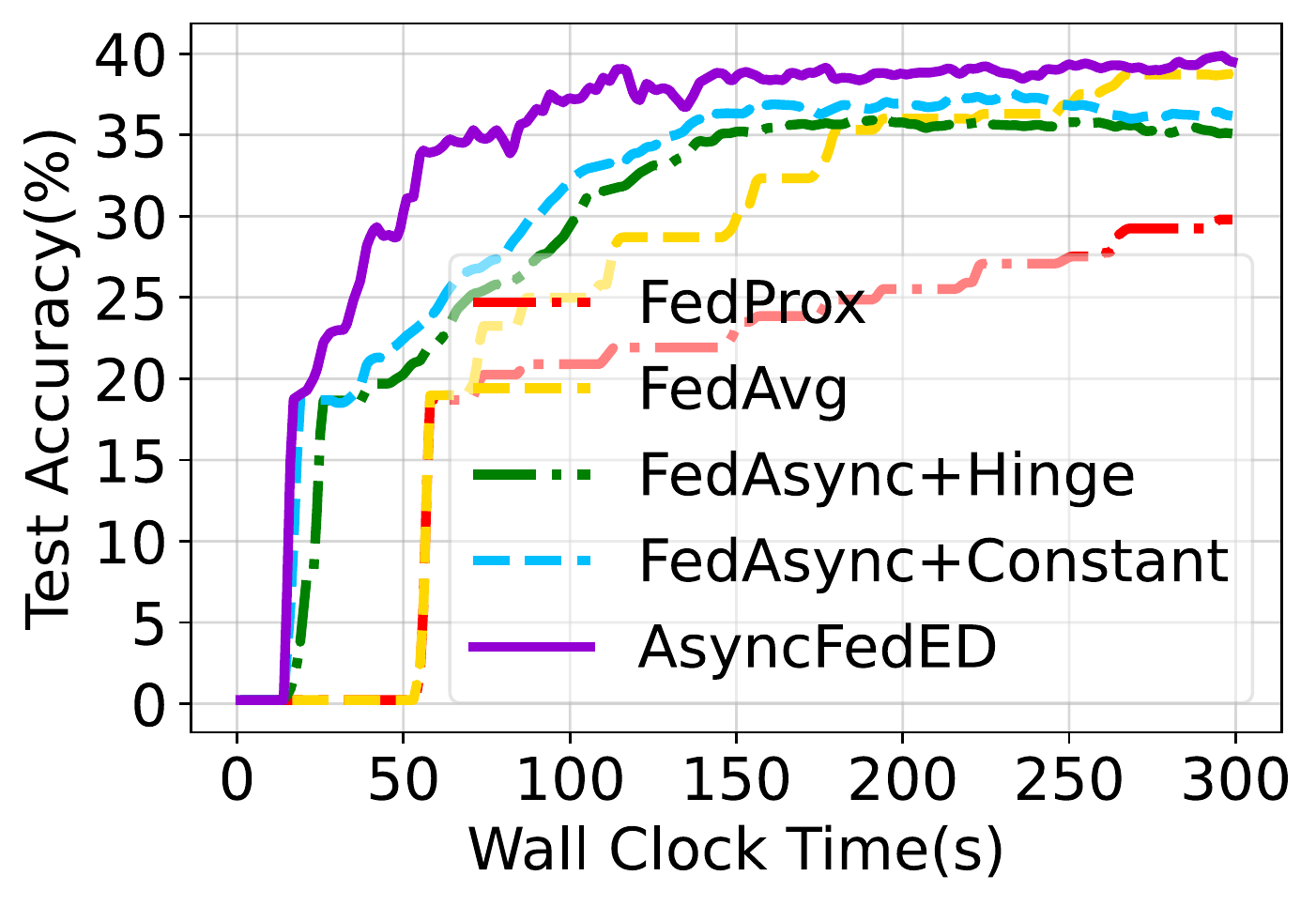}} 
\centering
\vspace{-0.3cm}
\caption{Test accuracy v.s. training time curves when the suspension probability is set to $P=0.4$. } 
\end{figure}
\vspace{-0.5cm}

\begin{figure}[H]
\centering
\subfigure[Synthetic-1-1]{\includegraphics[width=0.32\textwidth]{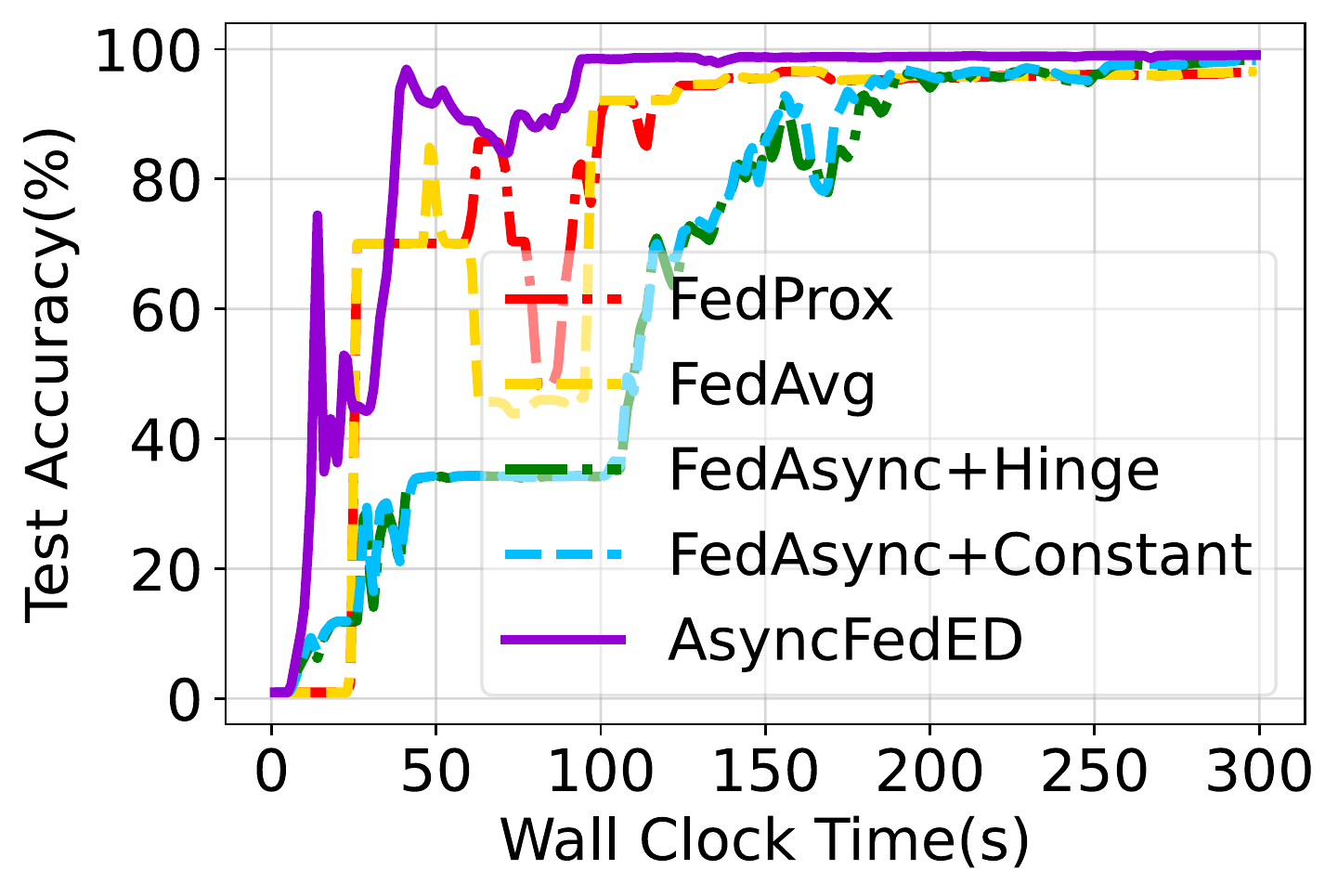}}
\subfigure[FEMNIST]{\includegraphics[width=0.32\textwidth]{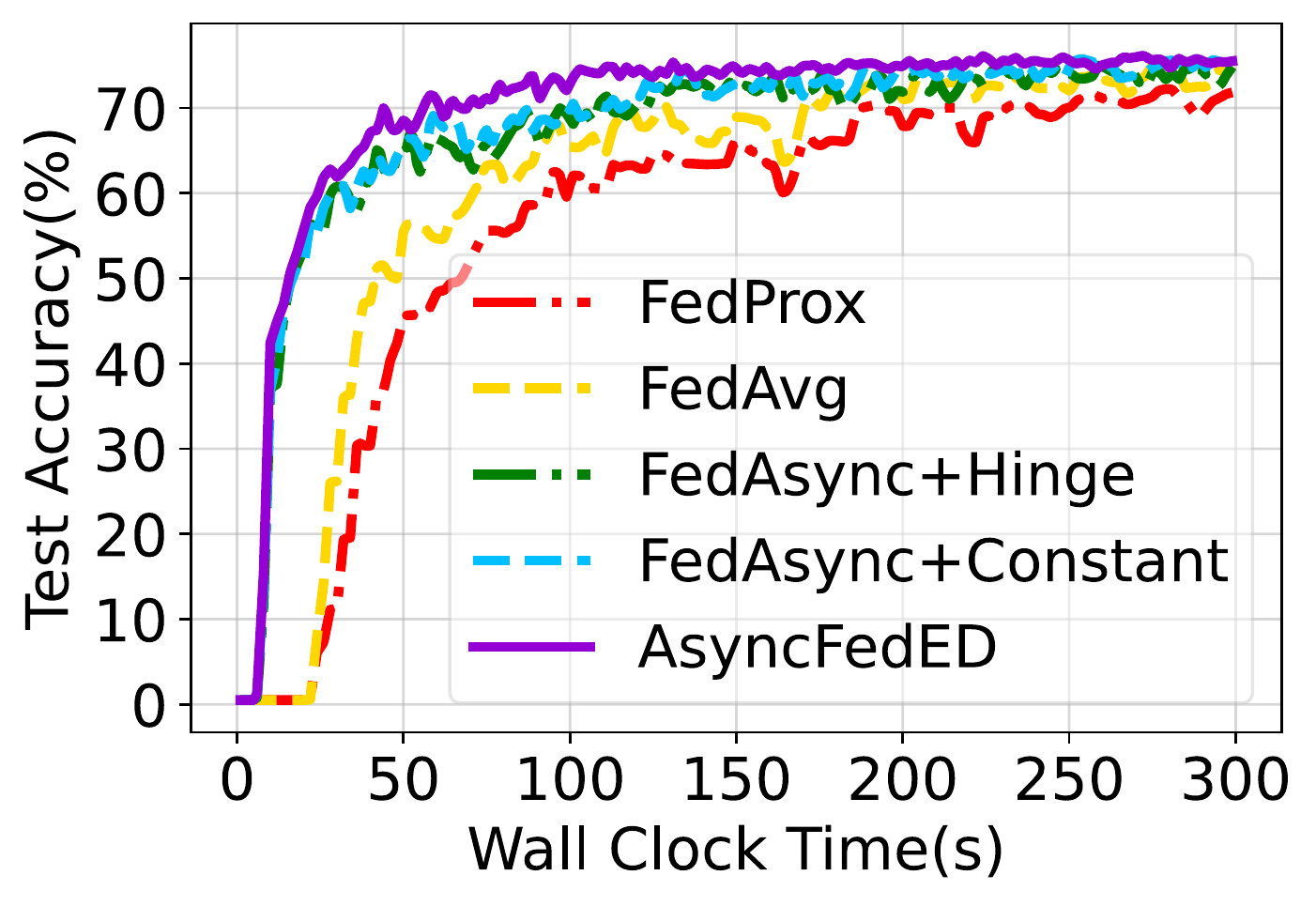}} 
\subfigure[ShakeSpeare text data]{\includegraphics[width=0.32\textwidth]{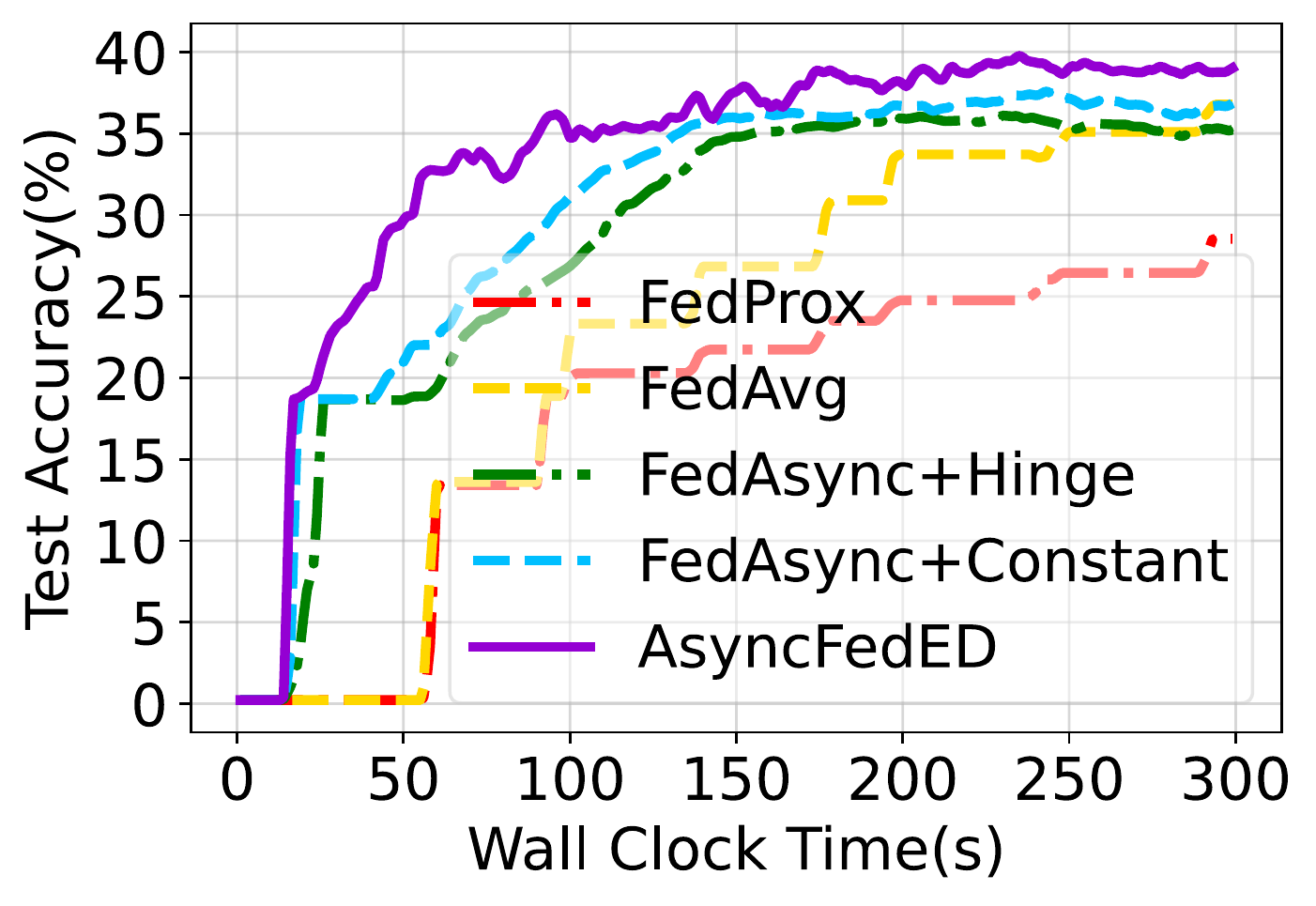}} 
\centering
\vspace{-0.3cm}
\caption{Test accuracy v.s. training time curves when the suspension probability is set to $P=0.5$. } 
\end{figure}
\vspace{-0.5cm}

\begin{figure}[H]
\centering
\subfigure[Synthetic-1-1]{\includegraphics[width=0.32\textwidth]{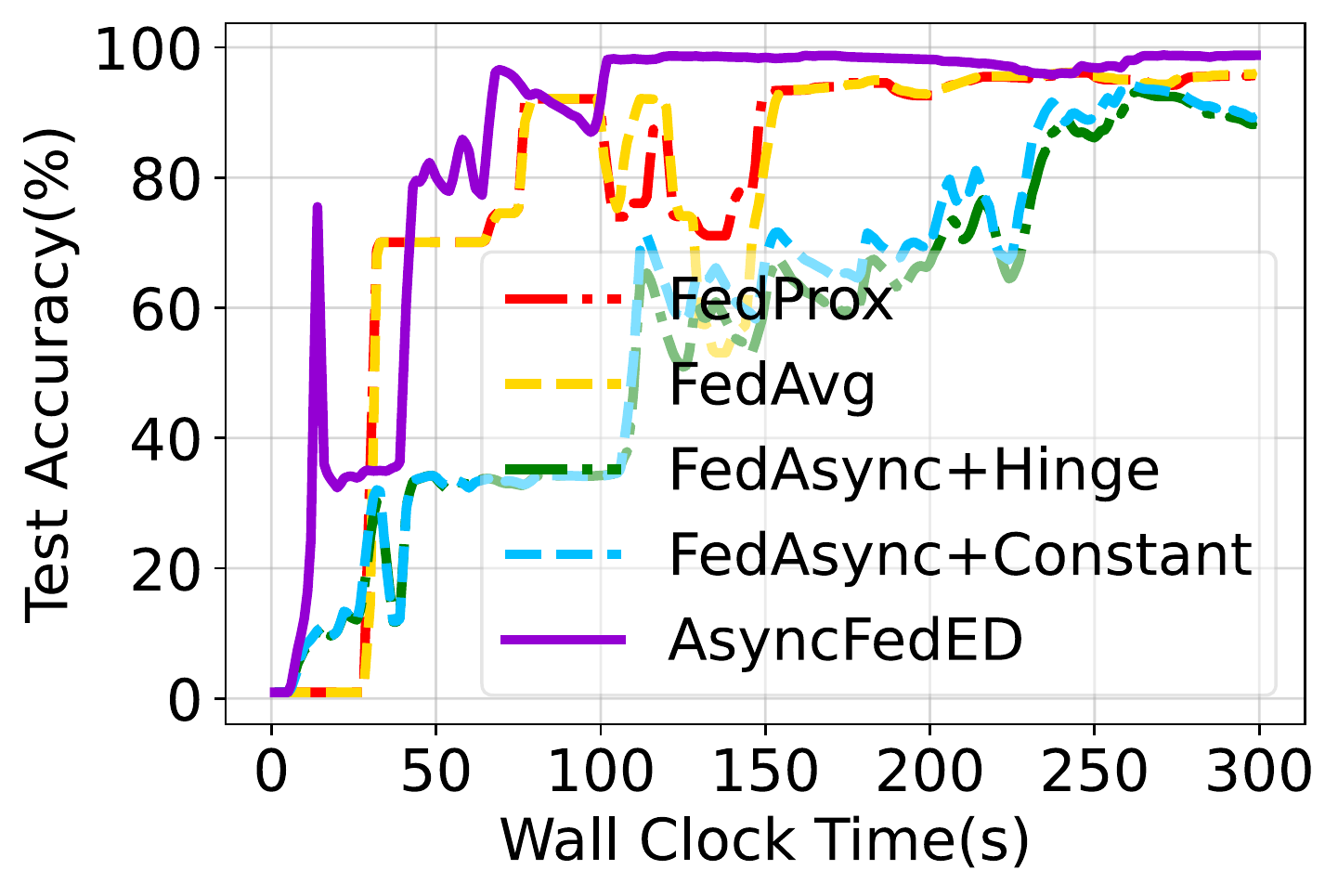}}
\subfigure[FEMNIST]{\includegraphics[width=0.32\textwidth]{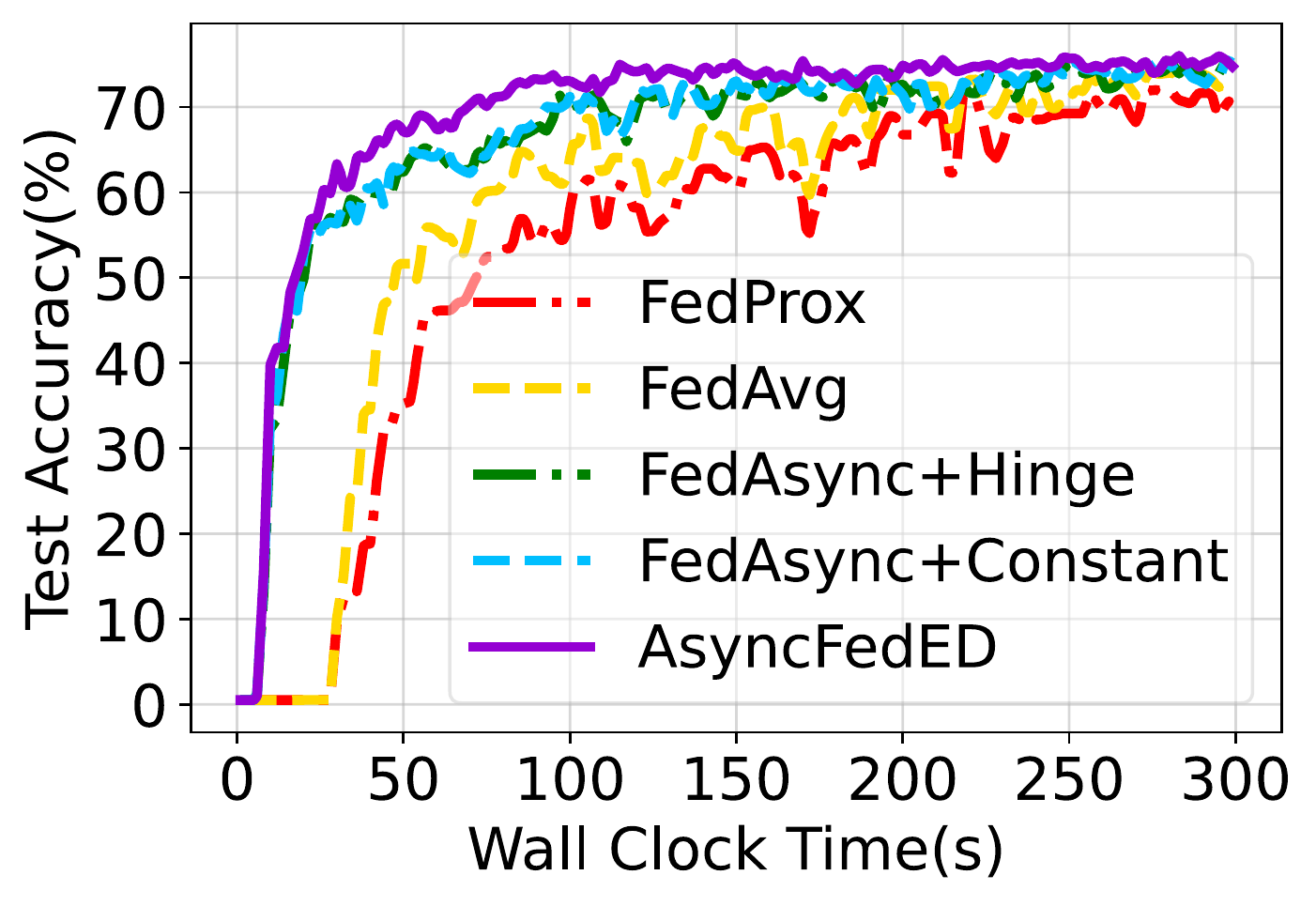}} 
\subfigure[ShakeSpeare text data]{\includegraphics[width=0.32\textwidth]{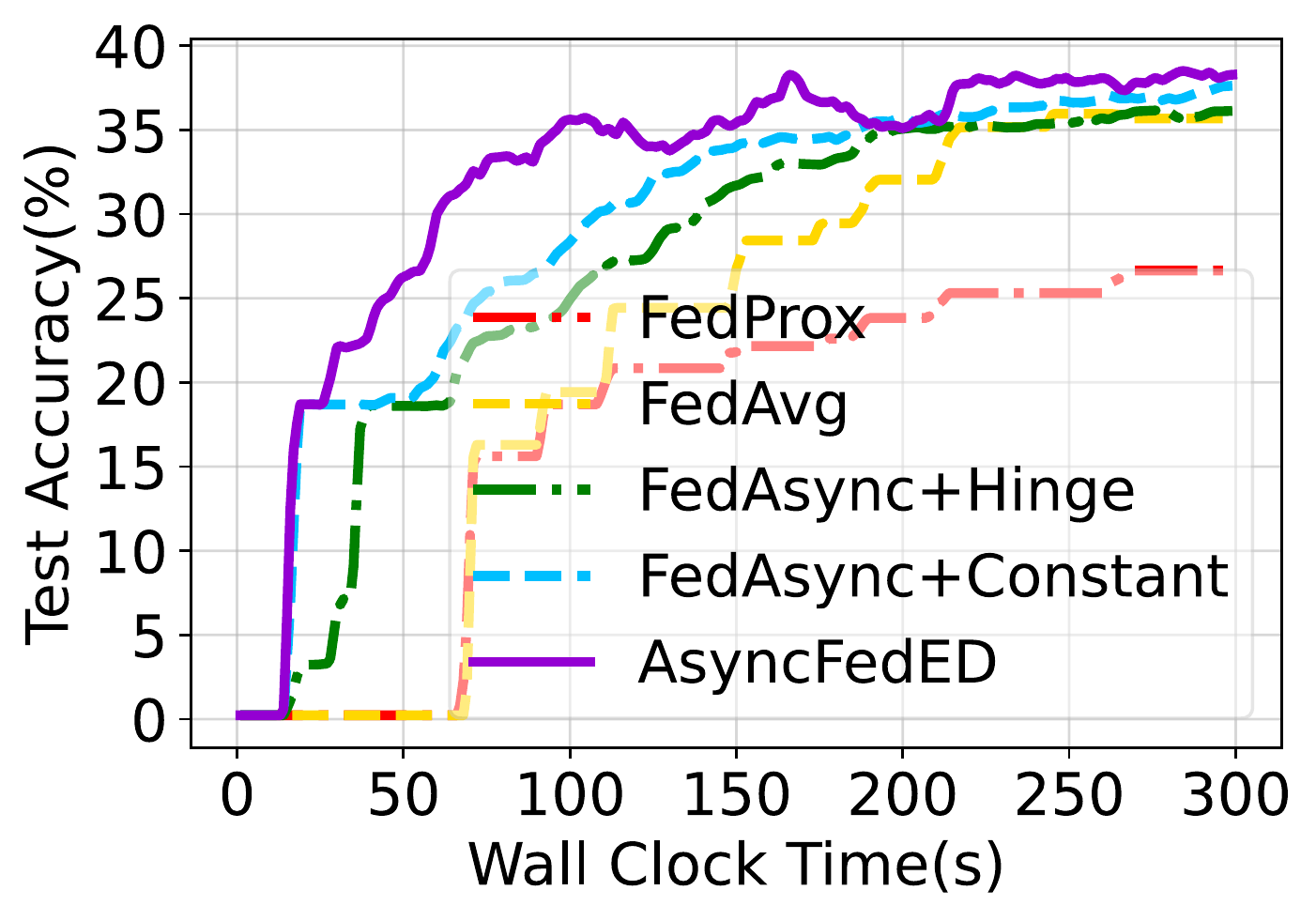}} 
\centering
\vspace{-0.3cm}
\caption{Test accuracy v.s. training time curves when the suspension probability is set to $P=0.6$. } 
\end{figure}
\vspace{-0.5cm}

\begin{figure}[H]
\centering
\subfigure[Synthetic-1-1]{\includegraphics[width=0.32\textwidth]{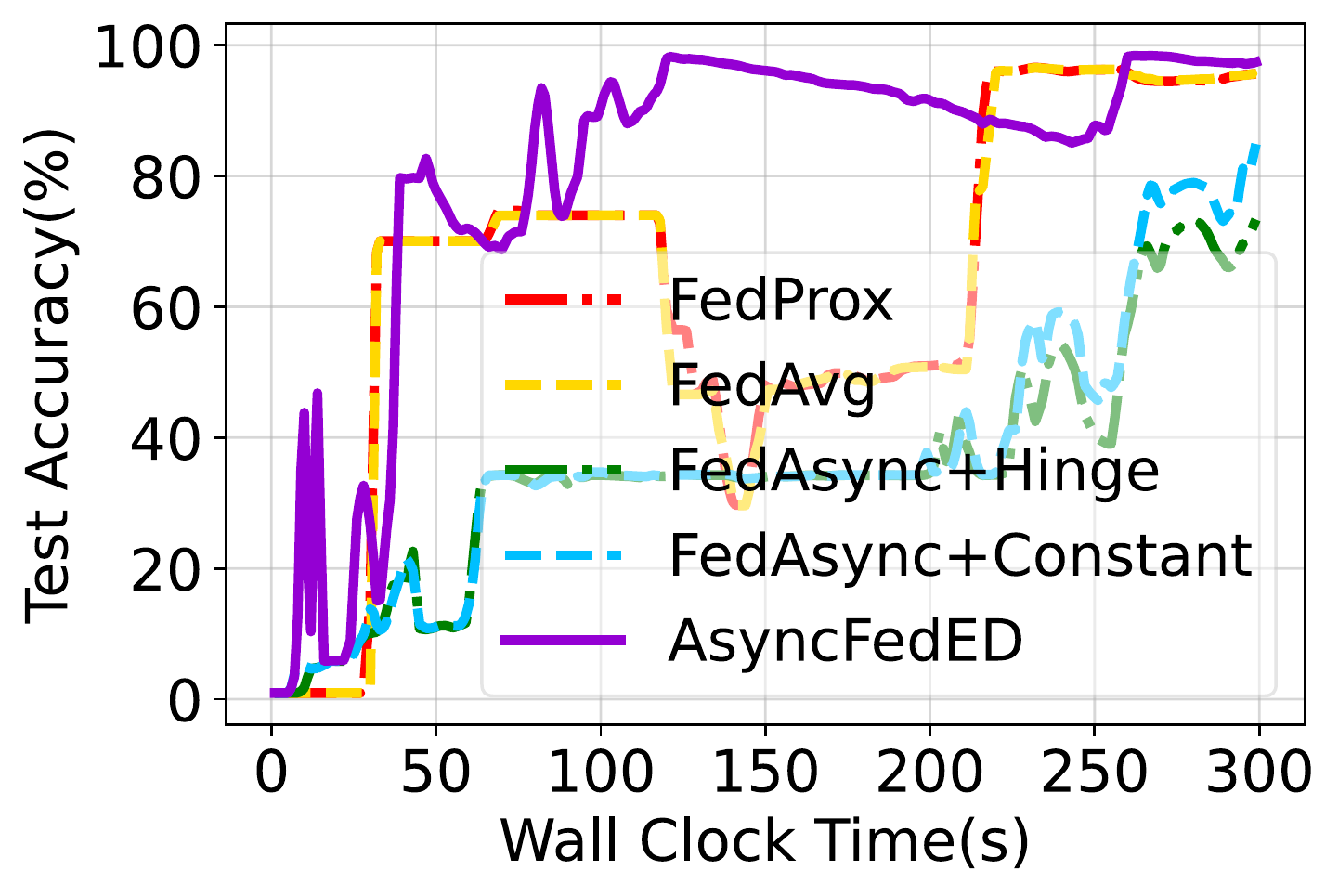}}
\subfigure[FEMNIST]{\includegraphics[width=0.32\textwidth]{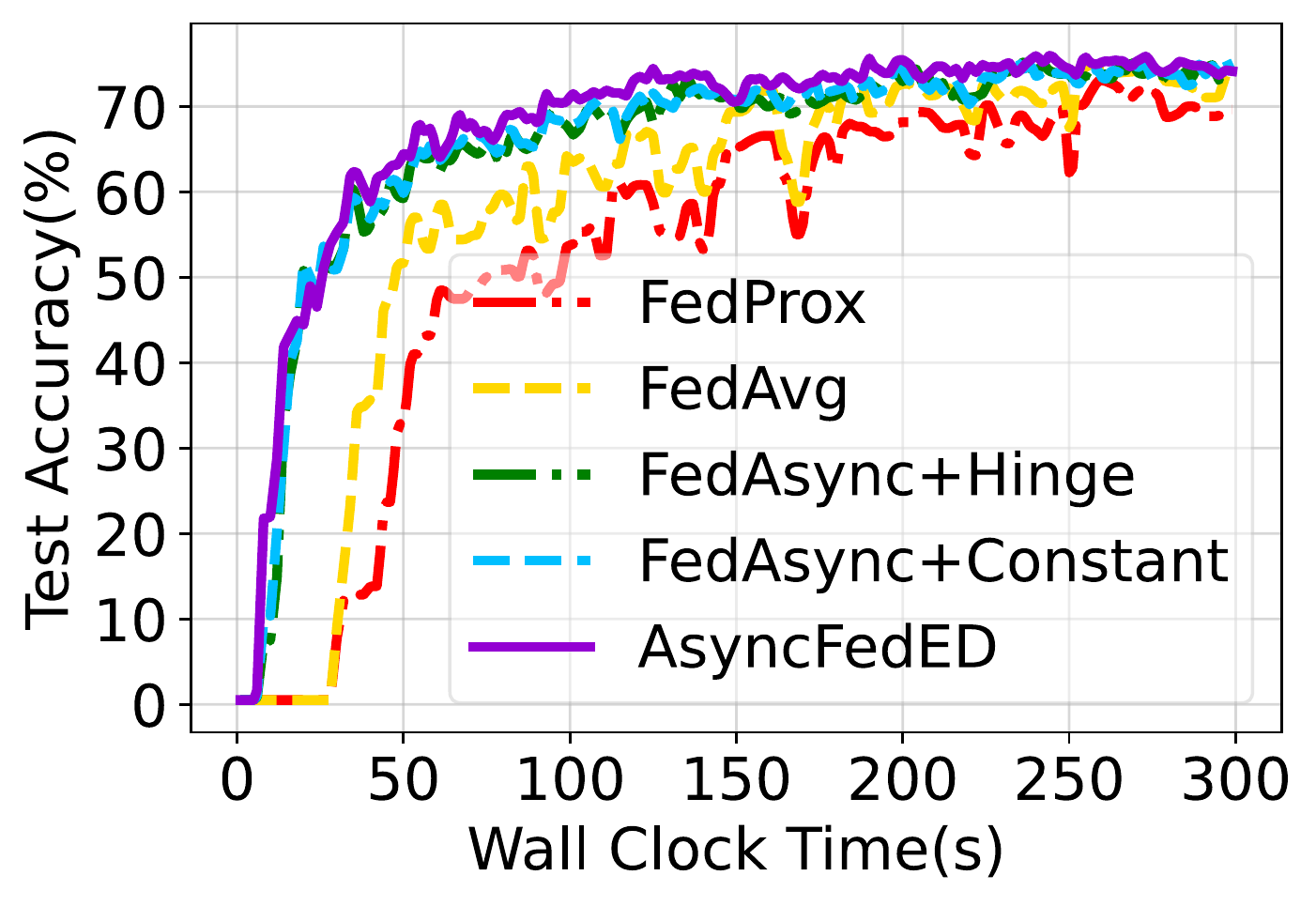}} 
\subfigure[ShakeSpeare text data]{\includegraphics[width=0.32\textwidth]{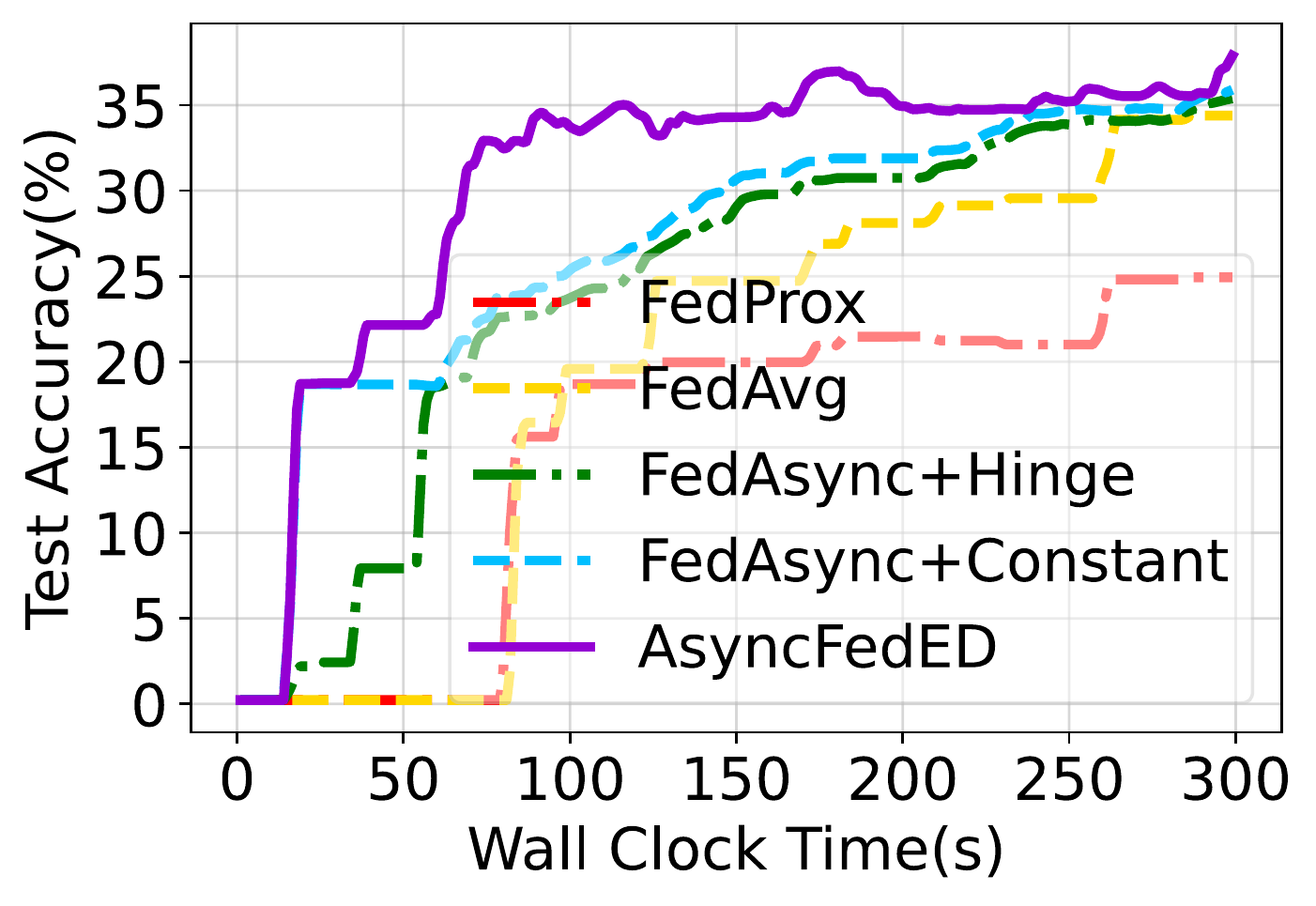}} 
\centering
\vspace{-0.3cm}
\caption{Test accuracy v.s. training time curves when the suspension probability is set to $P=0.7$. } 
\end{figure}
\vspace{-0.5cm}

\begin{figure}[H]
\centering
\subfigure[Synthetic-1-1]{\includegraphics[width=0.32\textwidth]{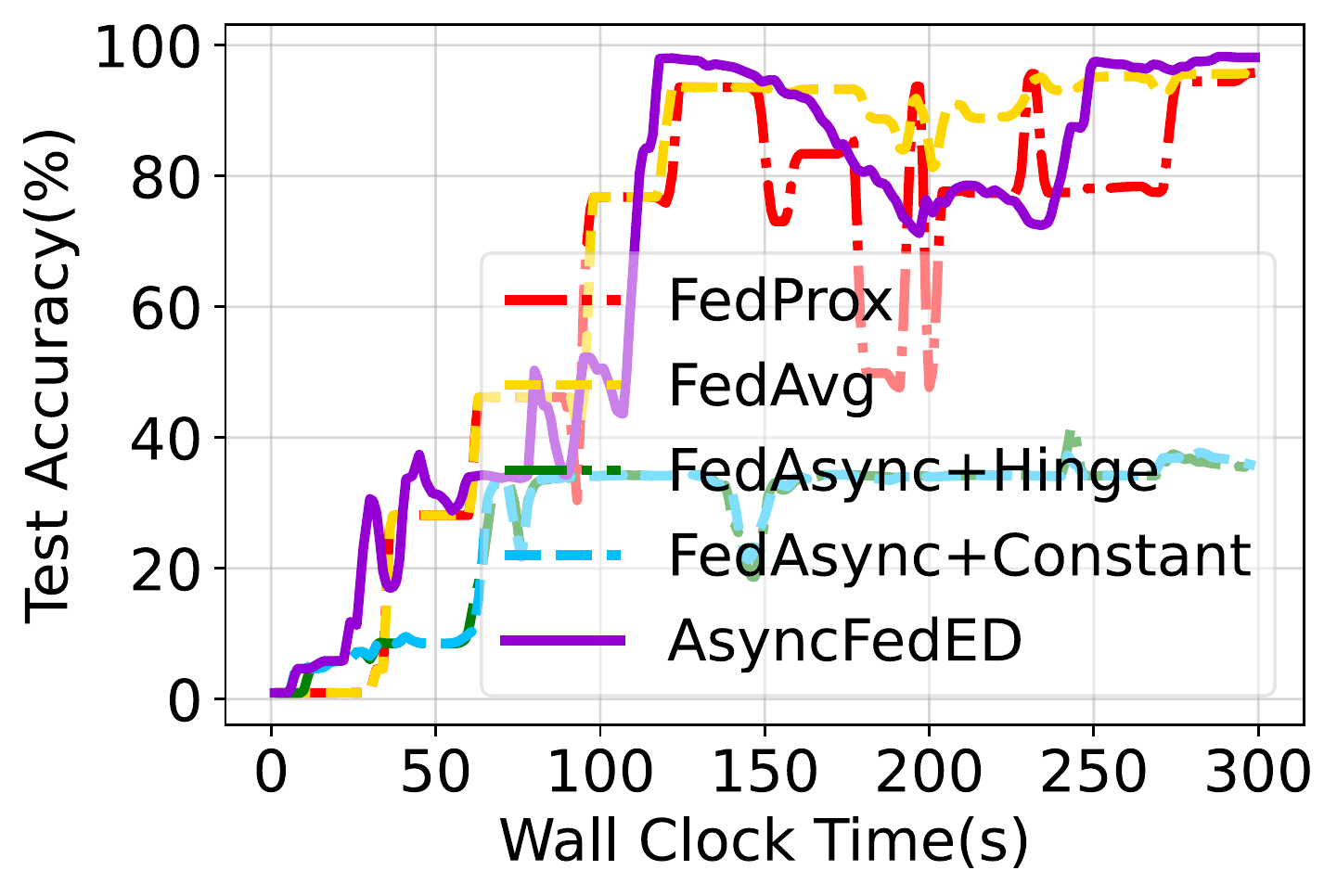}}
\subfigure[FEMNIST]{\includegraphics[width=0.32\textwidth]{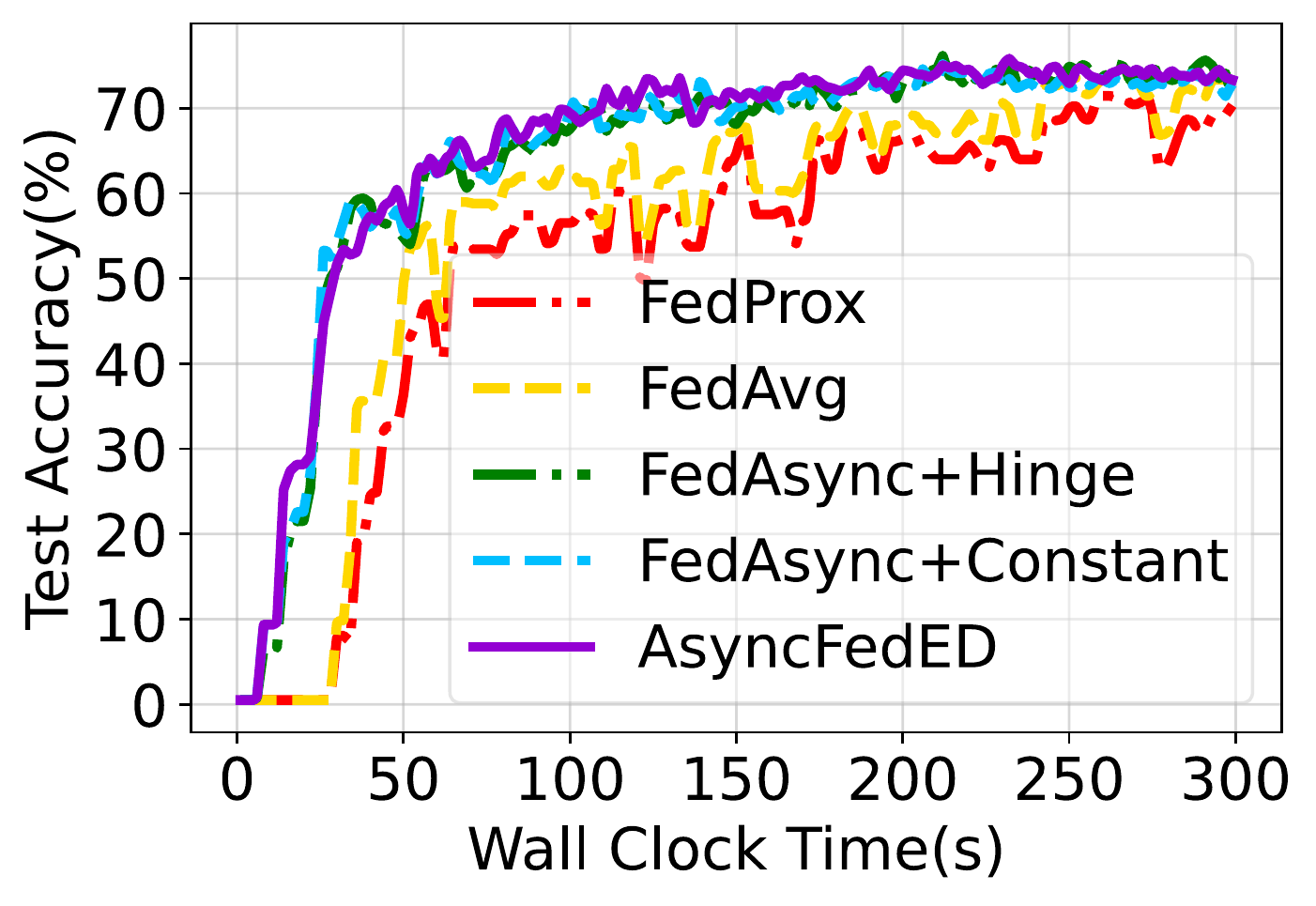}} 
\subfigure[ShakeSpeare text data]{\includegraphics[width=0.32\textwidth]{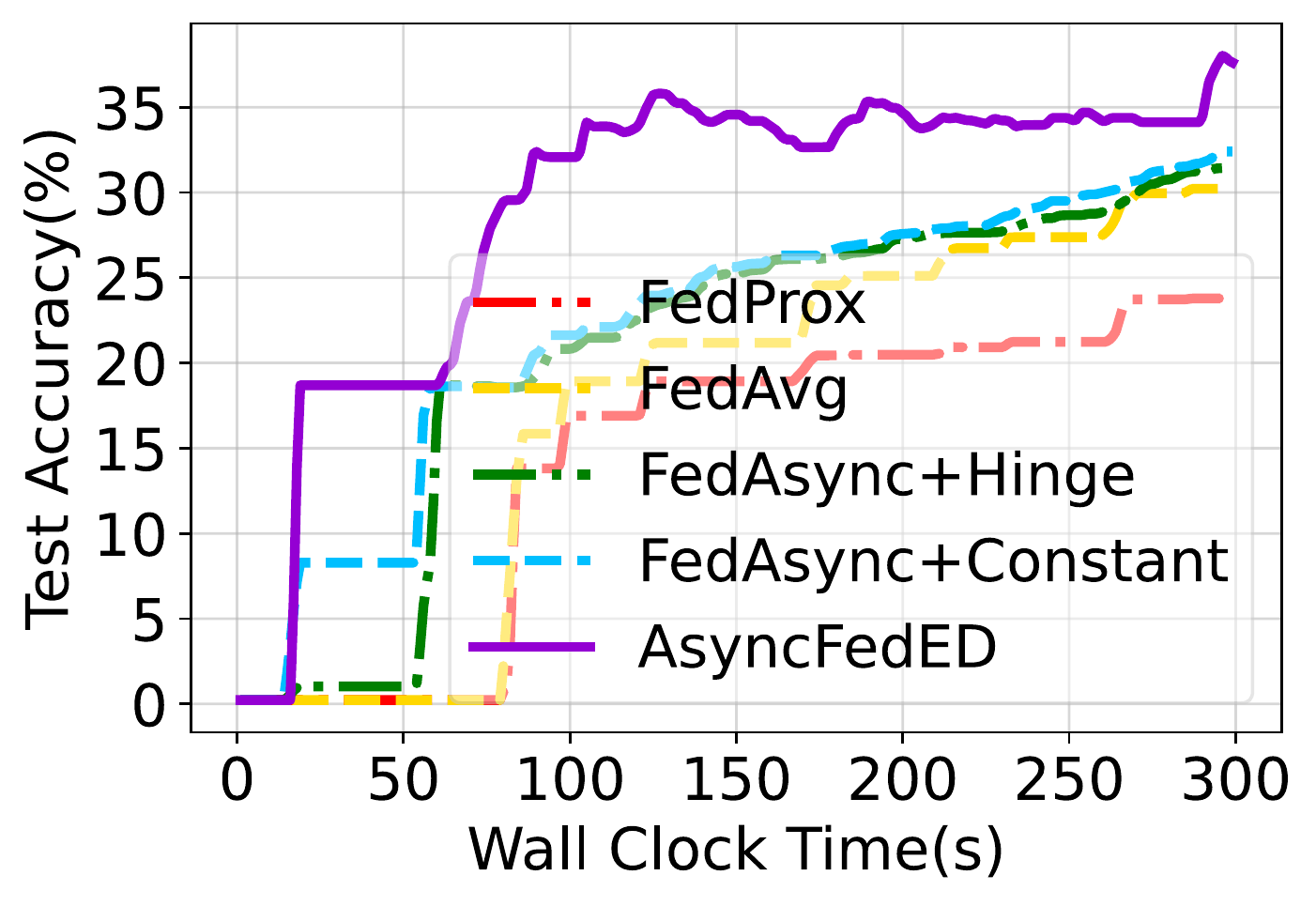}} 
\centering
\vspace{-0.3cm}
\caption{Test accuracy v.s. training time curves when the suspension probability is set to $P=0.8$. } 
\end{figure}
\vspace{-0.5cm}

\begin{figure}[H]
\centering
\subfigure[Synthetic-1-1]{\includegraphics[width=0.32\textwidth]{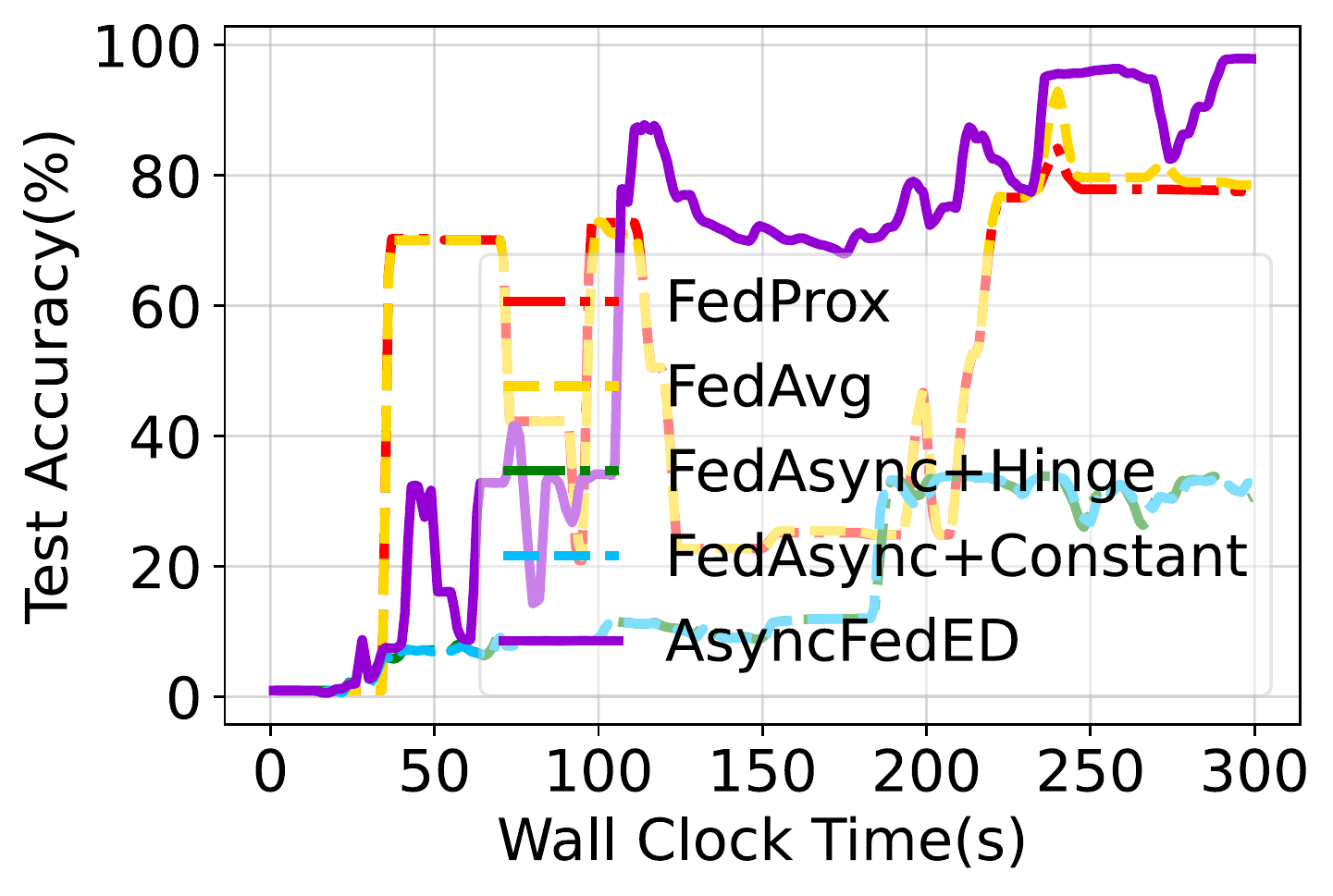}}
\subfigure[FEMNIST]{\includegraphics[width=0.32\textwidth]{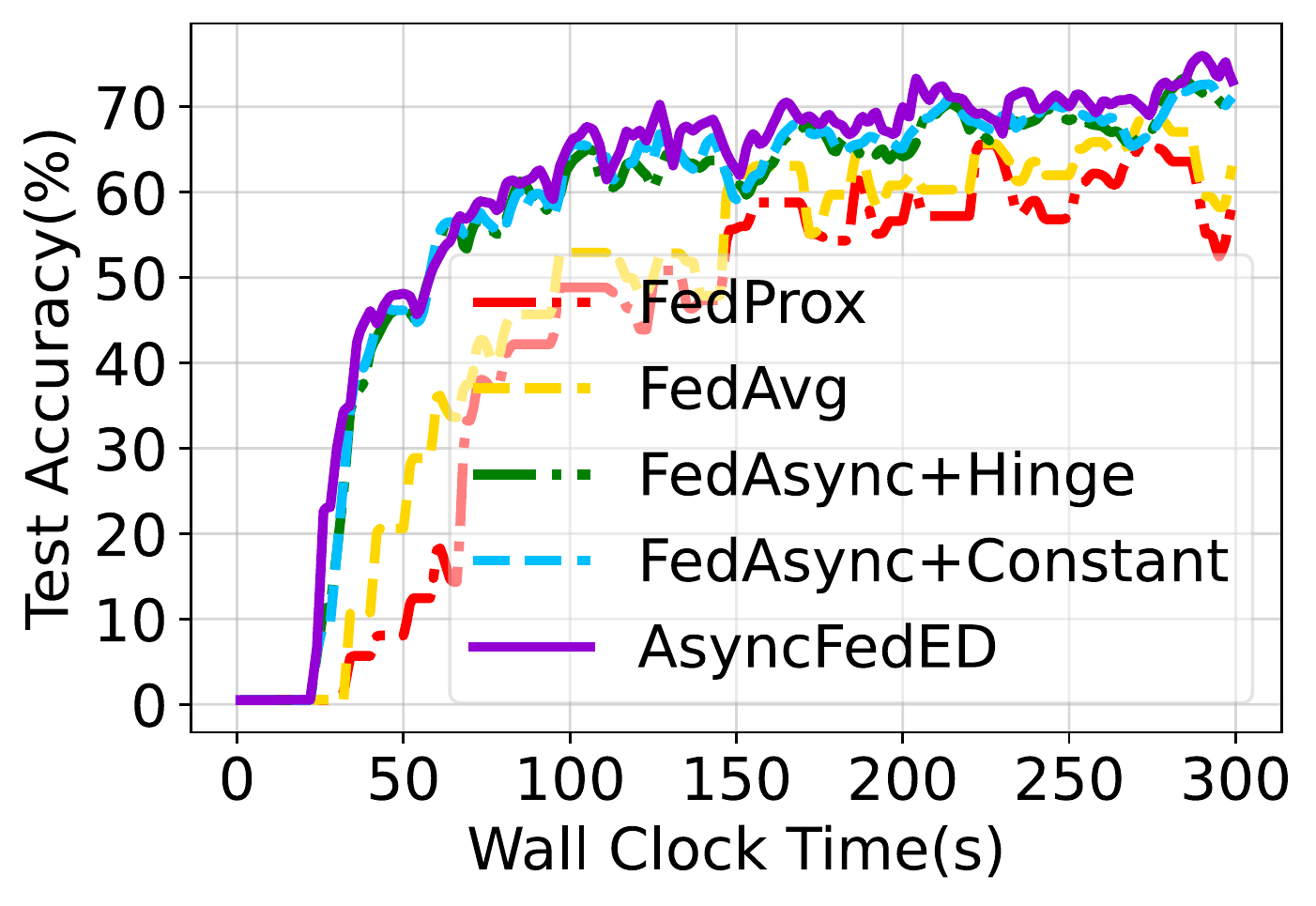}} 
\subfigure[ShakeSpeare text data]{\includegraphics[width=0.32\textwidth]{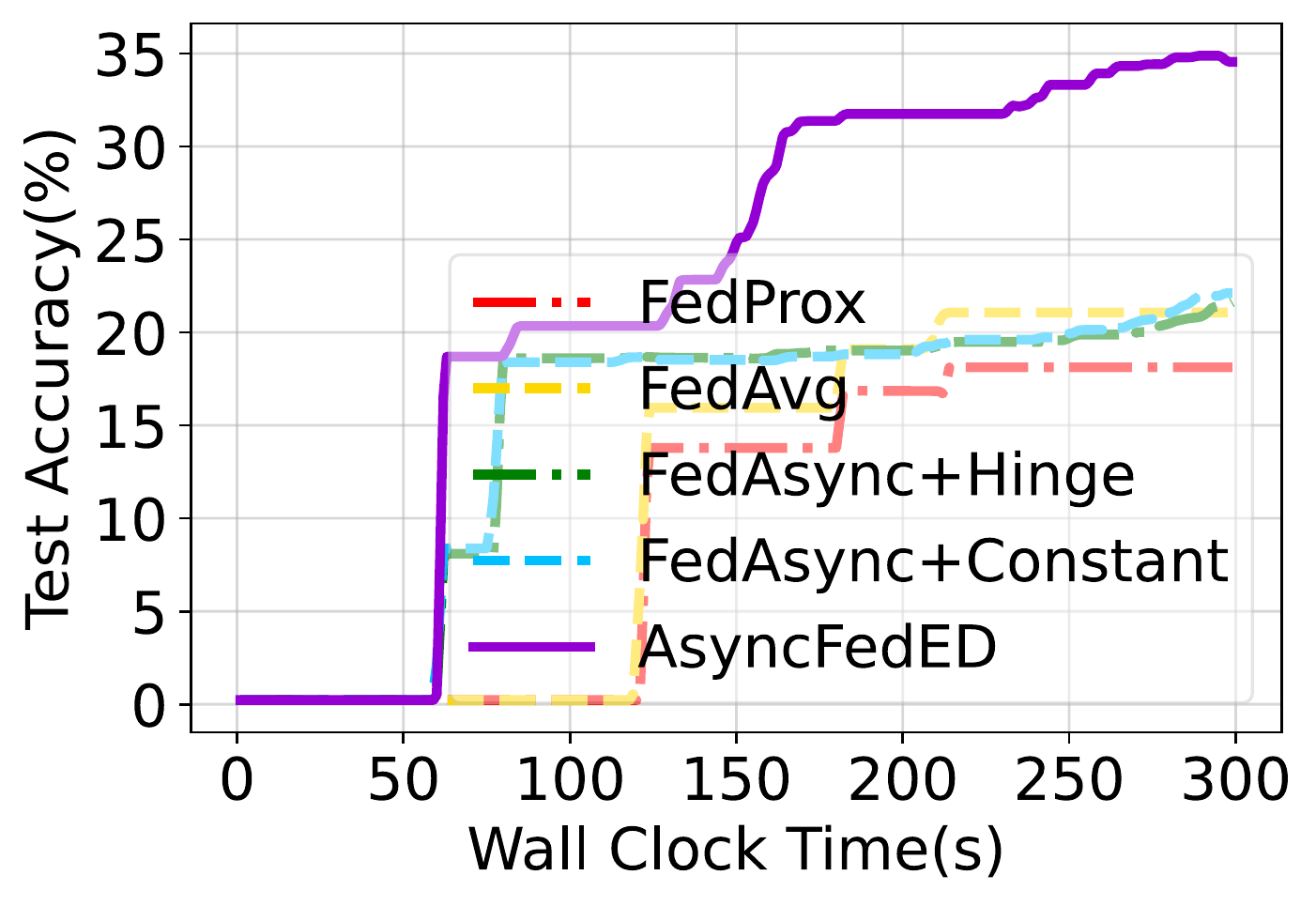}} 
\centering
\vspace{-0.3cm}
\caption{Test accuracy v.s. training time curves when the suspension probability is set to $P=0.9$. } 
\end{figure}

\subsection{Baselines and hyper parameters}\label{tune hyper parameters}
\paragraph{FedAvg}
FedAvg is a commonly used synchronous aggregation algorithm proposed by \citet{FedAvg}. In each round of FedAvg, the server selects some clients randomly for local training and weighted averages the received local models, where the weights depend on the amount of local data. Limited by the number of participating clients, we choose all clients to train there local model in each global round. One update step at the server can be denoted as:
\begin{equation}
    x_{t+1} = \frac{1}{m}\sum_{i=1}^{m}\frac{|\xi_i|}{\sum_{i=1}^{m}|\xi_i|}x_{t,K}^i, 
\end{equation}
where the $|\xi_i|$ represent the amount of local data stored in client $i$. 

\paragraph{FedProx} FedProx\cite{FedProx} is an improved version of FedAvg that shares the same update scheme with the FedAvg. The difference lies in that all local loss functions in FedProx have been applied with a regularization term to improve the performance on Non-IID data, that is 
\begin{equation}
    \begin{aligned}
    h_i(x_{t,k}) &= f_i(x_{t,k}) + \frac{\mu}{2} \left\|x_{t,k}^i - x_{t} \right\|^2.
    \end{aligned}
\end{equation}
We tune the hyperparameters by grid searching, and the corresponding sweep range and the values we select are shown in Table \ref{hyperFedProx}.
\begin{table}[H]
  \caption{Sweep range and selected hyperparameter for FedProx}
  \label{hyperFedProx}
  \centering
  \begin{tabular}{cccc}
    \toprule
    Task    & Hyperparameter     & Sweep range  &  Selected\\
    \midrule
    \vspace{0.2cm}
    Synthetic-1-1 & $\mu$  & $\{0.001,0.01,0.1,0.5,1\}$     &    0.1 \\
    \vspace{0.2cm}
    FEMNIST     & $\mu$ & $\{0.001,0.01,0.1,0.5,1 \}$      &  1   \\
    ShakeSpeare text data     & $\mu$  & $\{0.001,0.01,0.1,0.5,1 \}$  &   0.01    \\
    \bottomrule
  \end{tabular}
\end{table}

\paragraph{FedAsync+Constant} FedAsync\cite{AsyncFed} is the first proposed pure-asynchronous federated aggregation algorithm, which is most related to our work. FedAsync enables the server to update global model by computing a weighted average of the current global model and the received local model from client $i$, which is given by Eq.(\ref{fedasyncAggregration}):  
\begin{equation}
    x_{t+1} = (1-\alpha) x_{t} + \alpha x_{t-\tau,K}^i\label{fedasyncAggregration},
\end{equation}
where $\alpha$ in FedAsync+Constant is a constant number. We chose the best value of $\alpha$ by grid search with the corresponding sweep range and the values shown in Table \ref{hyperConstant}.
\begin{table}[H]
  \caption{Sweep range and selected hyperparameter for FedAsync+Constant}
  \label{hyperConstant}
  \centering
  \begin{tabular}{cccc}
    \toprule
    Task    & Hyperparameter     & Sweep range  &  Selected\\
    \midrule
    \vspace{0.2cm}
    Synthetic-1-1 & $\alpha$  & $\{0.01,0.1,0.3,0.5,0.9 \}$     & 0.1 \\
    \vspace{0.2cm}
    FEMNIST     & $\alpha$ & $\{0.01,0.1,0.3,0.5,0.9 \}$      &   0.5 \\
    ShakeSpeare text data     & $\alpha$  & $\{0.01,0.1,0.3,0.5,0.9 \}$  &    0.1 \\
    \bottomrule
  \end{tabular}
\end{table}

\paragraph{FedAsync+Hinge} AsyncFed+Hinge is an adaptive case of FedAsync. In FedAsync+Hinge scheme, the server or clients record the time of downloading the global model as $\tau$, and the time of receiving the local update as $t$, from which the $\alpha_t$ can be adaptively set by the hinge function. The global model update step is given by  Eq.(\ref{hinge})
\begin{equation}
\begin{aligned}
     s_{a, b}(t-\tau)&= \begin{cases}1 & \text { if } t-\tau \leq b \\ \frac{1}{a(t-\tau-b)+1} & \text { otherwise }\end{cases} \\
     \alpha_t &= \alpha \cdot s_{a,b}(t-\tau) \\
     x_{t+1} &= (1-\alpha_t) x_{t} + \alpha_t x_{t-\tau,K}^i\label{hinge}
\end{aligned}
\end{equation}
We also tune the two hyperparameters $(a,b)$ in FedAsync+Hinge scheme by grid search, which are presented in Table \ref{hyperHinge}.
\begin{table}[H]
  \caption{Sweep range and selected hyperparameters for FedAsync+Hinge}
  \label{hyperHinge}
  \centering
  \begin{tabular}{cccc}
    \toprule
    Task    & Hyperparameters     & Sweep range  &  Selected\\
    \midrule
    \multirow{2}*{Synthetic-1-1} & $a$  & $\{0.5,1,5,10,15 \}$   & 5 \\
    \vspace{0.2cm}
                ~                   &  $b$   &   $\{0.5,1,5,10,15 \}$ &   5   \\ 
    \multirow{2}*{FEMNIST}    & $a$ & $\{0.5,1,5,10,15 \}$      & 0.5 \\
    \vspace{0.2cm}
    ~                   &  $b$   &   $\{0.5,1,5,10,15 \}$ &   0.5   \\ 
    \multirow{2}*{ShakeSpeare text data}  & $a$  & $\{0.5,1,5,10,15 \}$  & 15  \\
    ~                   &  $b$   &   $\{0.5,1,5,10,15 \}$ &    15  \\ 
    \bottomrule
  \end{tabular}
\end{table}

\paragraph{AsyncFedED} AsyncFedED is the proposed asynchronous federated learning algorithm with adaptive weights aggregation scheme in this paper and the procedures have been summarized in Algorithm 1. The update at the server and the number of local epochs are given by :
\begin{equation}
    \begin{aligned}
         x_{t+1} &= x_t + \eta_{g,i} \cdot \Delta_i(x_{t-\tau,K})  \\
         K_{i,n+1} &= K_{i,n} + \mit{E}[(\Bar{\gamma} - \gamma(i,\tau_n)) \cdot \kappa]\label{furtherSimplify}
    \end{aligned}
\end{equation}
Note the maximum of $\eta_{g,i}$ is  $\dfrac{\lambda}{\varepsilon}$. We tune hyperparameters, $\dfrac{\lambda}{\varepsilon}, \lambda, \Bar{\gamma}, \kappa$, by grid search,  the sweep ranges and the selected values of which are shown in Table \ref{hyperAsyncFedED}.
\begin{table}[H]
  \caption{Sweep range and selected hyperparameters for AsyncFedED}
  \label{hyperAsyncFedED}
  \centering
  \begin{tabular}{cccc}
    \toprule
    Task    & Hyperparameters     & Sweep range  &  Selected\\
    \midrule
    \multirow{4}*{Synthetic-1-1} & $\dfrac{\lambda}{\varepsilon}$  & $\{0.1,0.5,1,3\}$     &   1   \\
            ~      & $\lambda$ &  $\{1,3,5,10 \}$   &  5                                   \\
            ~      & $\Bar{\gamma}$ &    $\{0.1,0.5,1,3,5 \}$    &  3                           \\
            \vspace{0.3cm}
            ~      & $\kappa$ &      $\{0.01,0.05,0.1,0.5,1 \}$   &  1                                \\
    \multirow{4}*{FEMNIST} & $\dfrac{\lambda}{\varepsilon}$  & $\{0.1,0.5,1,3\}$      &    1   \\
            ~      & $\lambda$ &  $\{1,3,5,10 \}$          &    1                            \\
            ~      & $\Bar{\gamma}$ &    $\{0.1,0.5,1,3,5 \}$       &  3                           \\
            \vspace{0.3cm}
            ~      & $\kappa$ &      $\{0.01,0.05,0.1,0.5,1 \}$         &  0.05                             \\
    \multirow{4}*{ShakeSpeare text data} & $\dfrac{\lambda}{\varepsilon}$  & $\{0.1,0.5,1,3\}$     &    0.5 \\
            ~      & $\lambda$ &    $\{1,3,5,10 \}$           &  5                            \\
            ~      & $\Bar{\gamma}$ &    $\{0.1,0.5,1,3,5 \}$        &  3                       \\
            ~      & $\kappa$ &      $\{0.01,0.05,0.1,0.5,1 \}$          &  1                         \\
    \bottomrule
  \end{tabular}
\end{table}

\paragraph{Local optimizer}
Similar to the other federated learning frameworks, the local training at each client in AsyncFedED is independent from the server and other clients given the initial global weights. Therefore, each client can use any optimizer and local learning rate of its choice as long as it uploads the pseudo gradients to the server after each iteration of local training. To ensure the fairness of comparison, we set the same local optimizer for all baselines and AsyncFedED. More specifically, we chose the momentum optimizer and set the momentum as 0.5 for all the clients. The local learning rate was set as 0.01 for Synthetic-1-1 and FEMNIST, 1 for ShakeSpeare text data, and the decay rate as 0.995 for all tasks. The number of local epochs $K$ of all the baselines and the initial number local epochs for AsyncFedED were set as 10.

\section{Main notions}
To make this paper easy to read, we summarize main notations in Table \ref{notations}.
\begin{table}[H]
  \caption{Main notations}
  \label{notations}
  \centering
    \begin{tabular}{c p{0.8\textwidth}} 
    \toprule
    Name     &  Description      \\
    \midrule
    $[m]$   & The set of clients participating in AFL.\\
    $i$   & The index of clients, $i \in [m]$. \\
    $\xi_i, \xi_{i,k}$ & Local dataset at client $i$ and the training batch from $\xi_i$ used in the $k$th training epoch at client $i$, respectively.       \\
    $x_{t}$     & Global model weights at iteration $t$.    \\
    $\eta_{i,k}$    & Local learning rate for $k-th$ local epoch at client $i$.   \\
    $x_{t,K}^i$     & Local model parameters of client $i$ after $K$ epochs of local training starting from the global weights $x_t$.      \\
    $\tau$     &  The lag of the locally available global weights to the current ones, $\tau \leq t$.    \\
    $\Delta_i(x_{t-\tau,K})$     &  Pseudo gradients by client $i$ with $K$ epochs local training started from the global weights $x_{t-\tau}$.    \\
    $\gamma(i,\tau)  $   &  Staleness of $\Delta_i(x_{t-\tau,K})$ with regards to the current version of global weight $x_t$.\\
    $\eta_{g,i}$     &  Global learning rate adaptively set for client $i$.      \\
    $\lambda,\varepsilon  $   & Hyperparameters for adaptively setting $\eta_g$.       \\
    $K_{i,n}  $   & The number of local epoch for client $i$ in its $n-th$ local training.   \\
    $\Bar{\gamma}  $   & Hyperparameters for adaptively setting $K$.   \\
    $\kappa  $   &  Step length for adaptively setting $K$.      \\
    $\mit{E}[\cdot] $   &    Floor function that round each value to the closest smaller integer.    \\
    $\left\|\cdot\right\|$  & $\ell^2$-norm $\parallel \cdot \parallel_2$. \\
    $\mathbb{E}[\cdot]$   &  total expectation with respect to the randomness from the mini-batch and the clients randomly participated in, i.e., $\mathbb{E} = \mathbb{E}_{i \in [m]}\mathbb{E}_{\xi_{i,k}}$      \\
    $x^{*}$  &  Optimum global weights to minimal the global loss function and $\mathbb{E}(x_T) = x^{*}$ when $T \to \infty$      \\
    \bottomrule
  \end{tabular}
\end{table}

\end{document}